\documentclass[mnsc,nonblindrev]{informs3Mod} 

\OneAndAHalfSpacedXI 

 \usepackage{natbib}
\bibpunct[, ]{(}{)}{,}{a}{}{,}%
\def\bibfont{\small}%
\def\bibsep{\smallskipamount}%
\def\bibhang{24pt}%
\def\BIBand{and}%

\TheoremsNumberedThrough     
\ECRepeatTheorems

\EquationsNumberedThrough    

\MANUSCRIPTNO{}

\usepackage{multicol}
\usepackage{multirow}
\usepackage{threeparttable}	
\usepackage{adjustbox, makecell}
\usepackage{arydshln}
\usepackage{booktabs}

\usepackage{xcolor}
\usepackage{xr, xr-hyper}
\usepackage[colorlinks	=true,
			urlcolor    =magenta,
			citecolor	=blue,
			linkcolor	=blue,
			filecolor	=blue]{hyperref}

\usepackage[ruled,vlined,noend]{algorithm2e} 
\SetKwInput{KwInit}{Initialize}
\SetKwInput{KwIn}{Receive}
\SetKwInput{KwOut}{Return}
\SetAlFnt{\small       }

\SetAlCapFnt{\small       }
\SetAlCapNameFnt{\small       }
\usepackage{algorithmic}
\algsetup{linenosize=\small       }

\usepackage{amssymb,amstext}
\usepackage{scalerel}
\usepackage{upgreek}
\makeatletter
\newcommand{\raisemath}[1]{\mathpalette{\raisem@th{#1}}}
\newcommand{\raisem@th}[3]{\raisebox{#1}{$#2#3$}}
\makeatother
\usepackage{mathtools}

\newcommand{\sSpace}{\ensuremath{\mathcal{S}}}

\newcommand{\aSpaceS}{\ensuremath{{\mathcal{A}}_{s}}}
\newcommand{\polCost}{\ensuremath{\mathrm{PC}}}
\newcommand{\expt}{\ensuremath{\mathbb{E}}}
\DeclareMathOperator*{\diff}{d}
\DeclareMathOperator*{\prob}{Pr}
\newcommand{\R}{\mathbb{R}}
\newcommand{\saSpaceS}{\ensuremath{\sSpace\times\mathcal{A}_s}}

\newcommand{\sDim}{{d}}

\newcommand{\coefInf}{\ensuremath{\boldsymbol{\beta}}}
 
\newcommand{\randBasisSet}{\ensuremath{\mathcal{R}}}
\newcommand{\sNorm}[1]{{\lVert #1 \rVert}}

\newcommand{\lipConst}{\ensuremath{\mathrm{L}}}
\newcommand{\coefFELPVec}{\ensuremath{{\coefInf}^{\raisemath{1pt}{\scaleto{\mathrm{*}}{3.5pt}}}}}
\newcommand{\coefFELP}{\ensuremath{\beta_{0}^{\raisemath{2pt}{\scaleto{\mathrm{*}}{3.5pt}}}}}

\newcommand{\diamSState}{\ensuremath{D_s}} 
\newcommand{\tallNorm}[1]{{\raisemath{1pt}{\big\lVert} #1 \raisemath{1pt}{\big\rVert}}}

\newcommand{\visitFreq}[1]{\ensuremath{\raisemath{1.5pt}{\mu}_{\scaleto{\mathrm{#1}}{4pt}}}}
\usepackage{bbm}
\newcommand{\indicator}{\ensuremath{\mathbbm{1}}}

\newcommand{\VALP}{\ensuremath{V}(\coefALPVec)}
\newcommand{\coef}{\ensuremath{\beta}}
\newcommand{\coefALPVec}{\ensuremath{ {\beta}^{\raisemath{2pt}{\scaleto{\mathrm{FA}}{3.5pt}}}_{{\scaleto{{\mathrm{N}}}{5pt}}}}}
\newcommand{\coefALP}[1]{\ensuremath{{\beta}^{\raisemath{2pt}{\scaleto{\mathrm{FA}}{3.5pt}}}_{\scaleto{#1}{4pt}}}}

\newcommand{\coefALPVecN}[1]{\ensuremath{{\beta}^{\raisemath{2pt}{\scaleto{\mathrm{FA}}{3.5pt}}}_{{\scaleto{{\mathrm{#1}}}{4pt}}}}}
\newcommand{\coefSGVecK}[1]{\ensuremath{{\beta}^{\raisemath{2pt}{\scaleto{\mathrm{SG}}{3.5pt}}}_{{\scaleto{{\mathrm{#1}}}{4pt}}}}}

\newcommand{\simpleVecK}[1]{\ensuremath{{\beta}_{{\scaleto{{\mathrm{#1}}}{4pt}}}}}

\newcommand{\contractionFactor}{\ensuremath{{\Omega}}}

\newcommand{\onHandState}[1]{\ensuremath{s_{\scaleto{#1}{4.5pt}}}}
\newcommand{\pipeState}[1]{\ensuremath{u_{\scaleto{#1}{4.5pt}}}}
\newcommand{\programIndex}[1]{_{(\scaleto{\mathrm{#1}}{6pt})}}

\newcommand{\FALPprog}[1]{\ensuremath{\mathrm{\normalfont\texttt{FALP}}_{\scaleto{\mathrm{#1}}{4.5pt}}}}
\newcommand{\stateRelFALPprog}[2]{\ensuremath{\mathrm{\normalfont\texttt{FALP}}_{\scaleto{\mathrm{#1}}{5pt}}}[#2]}

\newcommand{\PGFALPprog}[1]{\ensuremath{\mathrm{\normalfont\texttt{FALP}}_{\scaleto{\mathrm{#1}}{5pt}}^{\raisemath{2pt}{\scaleto{\mathrm{PG}}{4pt}}}}}
\newcommand{\SGFALPprog}[1]{\ensuremath{\mathrm{\normalfont\texttt{FALP}}_{\scaleto{\mathrm{#1}}{5pt}}^{\raisemath{2pt}{\scaleto{\mathrm{SG}}{4pt}}}}}

\newcommand{\FGLPprog}[1]{\ensuremath{\mathrm{\normalfont\texttt{FALP}}_{\scaleto{\mathrm{#1}}{4pt}}^{\raisemath{2pt}{\scaleto{\mathrm{SG}}{4pt}}}}}

\newcommand{\FAconstr}{\ensuremath{h^{\scaleto{\mathrm{FA}}{4pt}}}}

\defcitealias{hernandez1996discrete}{HL}
\defcitealias{calafiore2006scenario}{CC}

\newcommand{\FALPBasesPO}{\text{ALP}$^{\raisemath{2pt}{\scaleto{\mathrm{DFM}}{4pt}}}$}

\newcommand{\constrViolLearn}{{constraint violation learning }}

\newcommand{\coefFELPVecBar}{\ensuremath{{\coefInfBar}^{\raisemath{0pt}{\scaleto{\mathrm{*}}{3.5pt}}}}}

\newcommand{\FELPFeasSlnIntercept}[1]{\ensuremath{{\beta}^{\raisemath{2pt}{\scaleto{\mathrm{FE}}{3.5pt}}}_{#1}}}      
\newcommand{\FELPFeasSlnVectorBar}[1]{\ensuremath{{{\coefInfBar}}^{\raisemath{1pt}{\scaleto{\mathrm{FE}}{3.5pt}}}_{#1}}}       
\newcommand{\FELPFeasSlnVector}[1]{\ensuremath{{{\coefInf}}^{\raisemath{1pt}{\scaleto{\mathrm{FE}}{3.5pt}}}_{#1}}}       

\newcommand{\vectorIndex}[1]{_{\scaleto{\mathrm{#1}}{4pt}}}

\newcommand{\optVproj}{\ensuremath{\coefInf^{\raisemath{1pt}{\mathrm{*,o}}}_{\mathrm{N}}}}
\newcommand{\optVprojBar}{\ensuremath{\coefInfBar^{\raisemath{1pt}{\mathrm{*,o}}}_{\mathrm{N}}}}
\newcommand{\optVprojPerpBar}{\ensuremath{\coefInfBar^{\raisemath{1pt}{\mathrm{*,\bot}}}_{\mathrm{N}}}}
\newcommand{\optVprojPerp}{\ensuremath{\coefInf^{\raisemath{1pt}{\mathrm{*,\bot}}}_{\mathrm{N}}}}

\newcommand{\coefVCVecBar}{\ensuremath{{\coefInfBar}^{\raisemath{0pt}{\scaleto{\mathrm{C}}{3.5pt}}}}}
\newcommand{\coefVCVec}{\ensuremath{{\coefInf}^{\raisemath{1pt}{\scaleto{\mathrm{C}}{3.5pt}}}}}
\newcommand{\coefVC}{\ensuremath{\beta_{0}^{\raisemath{2pt}{\scaleto{\mathrm{C}}{3.5pt}}}}}

\usepackage{verbatim}

\DeclareMathSymbol{\BetaB}{\mathalpha}{operators}{"42}
\newcommand{\Beta}{\boldsymbol{\BetaB}}

\newcommand{\coefInfBar}{\ensuremath{\Beta}}

\newcommand{\shiftFeas}{\ensuremath{\Gamma}}
\newcommand{\coefFeas}[1]{\ensuremath{{\beta}^\theta_{\scaleto{#1}{4pt}}}}
\newcommand{\coefFeasVec}{\ensuremath{{\beta}^{\theta}}}
\newcommand{\saDim}{{d_{\scaleto{(s,a)}{7pt}}}}
\newcommand{\saSpace}{\sSpace\times\mathcal{A}}
\newcommand{\diamSSaspace}{\ensuremath{D_{(s,a)}}}

\usepackage{xr, xr-hyper}
\begin{document}



\RUNAUTHOR{Pakiman et al.} 

\RUNTITLE{Self-guided Approximate Linear Programs}

\TITLE{Self-guided Approximate Linear Programs}

\ARTICLEAUTHORS{%
	\AUTHOR{Parshan Pakiman, Selvaprabu Nadarajah, Negar Soheili}
	\AFF{College of Business Administration, University of Illinois at Chicago, 601 South Morgan Street, Chicago, IL60607, USA \EMAIL{\{ppakim2@uic.edu,selvan@uic.edu,nazad@uic.edu\}}} 
	\AUTHOR{Qihang Lin}
	\AFF{Tippie College of Business, The University of Iowa, 21 East Market Street, Iowa City, IA 52242, USA\\ \EMAIL{qihang-lin@uiowa.edu}} 
} 

\ABSTRACT{%
	Approximate linear programs (ALPs) are well-known models based on value function approximations (VFAs) to obtain policies and lower bounds on the optimal policy cost of discounted-cost Markov decision processes (MDPs). Formulating an ALP requires (i) basis functions, the linear combination of which defines the VFA, and (ii) a state-relevance distribution, which determines the relative importance of different states in the ALP objective for the purpose of minimizing VFA error. Both these choices are typically heuristic: basis function selection relies on domain knowledge while the state-relevance distribution is specified using the frequency of states visited by a heuristic policy. We propose a self-guided sequence of ALPs that embeds random basis functions obtained via inexpensive sampling and uses the known VFA from the previous iteration to guide VFA computation in the current iteration. Self-guided ALPs mitigate the need for domain knowledge during basis function selection as well as the impact of the initial choice of the state-relevance distribution, thus significantly reducing the ALP implementation burden. We establish high probability error bounds on the VFAs from this sequence and show that a worst-case measure of policy performance is improved. We find that these favorable implementation and theoretical properties translate to encouraging numerical results on perishable inventory control and options pricing applications, where self-guided ALP policies improve upon policies from problem-specific methods. More broadly, our research takes a meaningful step toward application-agnostic policies and bounds for MDPs.\looseness=-1
}%


\HISTORY{December 2019 (initial version); October 2021 (this version).
}

\maketitle

\section{Introduction}\label{sec:intro}
Computing high-quality control policies in sequential decision making problems is an important task across several application domains. Markov decision processes (MDPs; \citealt{puterman1994MDP}) provide a powerful framework to find optimal policies
in such problems but are often intractable to solve exactly due to their large state and action spaces or the presence of high-dimensional expectations (see \S 1.2 and \S 4.1 of \citealp{powell2007ADP}). Therefore, a class of approximate dynamic programming (ADP) approaches instead approximate the value functions of MDPs and use the resulting approximations to obtain control policies in simulations (\citealt{bertsekas1996neuro}). Approximate linear programming  \citep{schweitzer1985ALP,farias2003ALP} is a math-programming based ADP approach for computing value function approximations (VFAs) that has been applied to a wide variety of domains, including operations research, reinforcement learning, and artificial intelligence \citep{adelman2003price,guestrin2003efficient,forsell2006approximate,desai2009smoothed,adelman2013dynamic,tong2013approximate,nadarajah2015relaxALP,mladenov2017approximate,balseiro2019multiagent,blado2019relaxation}.
It solves a so called approximate linear program (ALP) to obtain a VFA, from which a control policy can be computed. This VFA can also be used to obtain a lower bound on the optimal policy cost, which facilitates the computation of an optimality gap for the ALP policy as well as other heuristic policies. \looseness=-1


Formulating an ALP requires (i) basis functions, the linear combination of which defines the VFA over the MDP state space, and (ii) a state-relevance distribution, which determines the relative importance of different states in the ALP objective for the purpose of minimizing VFA error. It is well known that the choices of basis functions and the state-relevance distribution are challenging and are typically handled heuristically, the former using domain knowledge and the latter most commonly by considering the states visited by a heuristic policy (see \S5 in \citealt{Farias2006} and \S 3.2.2 in \citealt{sun2014quadratic}). Once an ALP is formulated, its solution provides the weights associated with basis functions defining a VFA but requires tackling a large-scale, potentially semi-infinite, linear program. Solving an ALP can be approached, for example, using techniques such as constraint generation, constraint sampling, and constraint-violation learning (see \citealp{lin2017ContViolLearning} for a recent overview of ALP solution techniques). If the ALP VFA gives rise to a policy and a lower bound with a small optimality gap, the near-optimal policy is used; otherwise, the choice of basis functions and state-relevance distribution will need to be modified. The initial choice and possible modification of ALP parameters is of fundamental importance to ensure the quality of the ALP VFA but their choice has received limited attention in the literature (\citealp{klabjan2007infinite}, \citealp{adelman2012GJR}, and \citealp{bhat2012NonParaALP}). The goal of this paper is to broaden the applicability of ALP by reducing the burden of making these choices.


Our starting point is to provide a new reformulation of a discounted-cost MDP as a large-scale linear program. This linear program has  infinitely many variables corresponding to a weighted integral of a continuum of basis functions, referred to as random basis functions (or random features in machine learning), and a large number of constraints (possibly infinite), one for each MDP state and action pair. Random Fourier basis functions defined using cosines are popular examples \citep{rahimi2008large}. A functional analogue of Monte Carlo sampling can be used to approximate the integral over random basis functions. The resulting model, dubbed feature-based approximate linear program (FALP), has variables corresponding to the VFA weights in a linear combination of randomly sampled basis functions. We establish high-probability bounds on the worst-case error between the FALP VFA and the MDP value function. In particular, this error bound converges at the dimension-free rate of one divided by the square root of the number of sampled random basis functions, analogous to the convergence rate of standard Monte Carlo sampling with respect to the number of samples. \looseness=-1 

While FALP does not rely on defining basis functions using domain knowledge, its formulation still requires choosing a state-relevance distribution. Misspecifying this distribution can lead to poor ALP policies (\citealt{farias2003ALP,sun2014quadratic}). The literature documents an iterative approach to guide this choice using ALP policy information (\citealp[Page 854]{farias2003ALP} and \citealp{Farias2006}). This approach first solves ALP formulated with fixed basis functions and a heuristic choice of the state-relevance distribution, then simulates the ALP policy to evaluate the probabilities of visiting states, then uses these probabilities as a state-relevance distribution in a subsequent solution of the ALP, and so on.  While intuitive, this approach lacks conceptual backing and also requires simulating the policy after every iteration, which can be expensive. We consider this iterative approach with FALP and refer to it as policy-guided FALP. \looseness=-1

We propose an alternative iterative approach that does not rely on policy simulation to update a state-relevance distribution. Instead, it leverages our ability to sample additional random basis functions inexpensively. To elaborate, we solve a sequence of FALP models with increasing numbers of random basis functions that include guiding constraints, which require the VFA being computed at a given iteration to be no smaller than the VFA available from the immediately preceding iteration. By dualizing the guiding constraints, we show that this sequence of models is equivalent to a series of FALP models with increasing numbers of random basis functions and adaptively updated state-relevance distributions. We thus label the model class as self-guided FALP because state-relevance distribution updates involve its own past VFA information. The sequence of VFAs associated with these models provides
monotonically increasing lower bounds and a monotonically non-increasing worst-case measure of policy performance. These properties are not guaranteed for policy-guided FALP. The “price” of the aforementioned desirable properties
is the larger number of constraints in self-guided FALP compared to FALP. We establish an error bound for self-guided FALP, highlight the effect of the guiding constraints on this bound, and discuss how solution approaches for FALP can be directly applied to solve self-guided FALP. \label{why-name-self-guided}

We validate the performance of the aforementioned ALP models on inventory control and options pricing applications, which give rise to discounted-cost MDPs with rather distinct properties.

The MDP for perishable inventory control has an infinite horizon, a non-convex cost structure, a continuous action space, and a continuous state space that is affected by decisions (i.e., controllable state space). MDPs that share similar properties also arise in lost-sales inventory control, healthcare screening, and dual sourcing, among other applications \citep{zipkin2008structure,steimle2017markov,hua2015structural}.
Policies for the perishable inventory control application with optimality gaps of not more than 12\% have been found by \citet{lin2017ContViolLearning} for a three-dimensional state space using an ALP model with problem-specific basis functions. FALP with a uniform state-relevance distribution essentially closes the optimality gaps on these known instances, that is, no iterative versions of FALP are needed. In contrast, for new instances that we create with five- and ten-dimensional state spaces, FALP performs poorly with optimality gaps ranging from 5.9\% to 27.7\% and  policy-guided FALP becomes unstable. Self-guided FALP instead obtains excellent policies in these high-dimensional instances with optimality gaps of less than 7.5\%. 

The options pricing application involves a finite-horizon, a non-convex reward, a finite action space, and a continuous state space that evolves in an exogenous manner (that is, it is not affected by decisions). This MDP structure is representative of many financial and real options problems (\citealt{Smith1998,haugh2004pricing,secomandi2010optimal,glasserman2013monte}), and ALP has been shown to perform poorly relative to least-squares Monte Carlo for real options pricing (\citealt{nadarajah2015relaxALP,nadarajah2017relationship}). For our experiments, we consider the Bermudan option instances from \cite{desai2012pathwise} with up to a sixteen-dimensional state space and add the least-squares Monte Carlo policy, which is near-optimal in these instances, to our set of ALP benchmarks. We find that the self-guided FALP policy value is higher than the least-squares Monte Carlo policy value by 2\% on average and by upto 4\%, a significant improvement for this application. In contrast, the average performance of an application-specific ALP model and FALP are worse than least-squares Monte Carlo by roughly $1\%$.\looseness=-1
Our results show that self-guided FALP promises to reduce the ALP implementation burden and improve the effectiveness of its policies in a broader class of applications. This approach, which does not directly exploit domain knowledge, may not always improve on application-specific methods. Regardless, it can still serve as a useful benchmark to assess the value of procedures that exploit application structures. To facilitate such benchmarking, we have made Python code implementing the approaches developed in this paper publicly available.  


\subsection{Novelty and Contributions}\label{subsec:Novelty and Contributions}

Research on ALPs predominantly assumes a fixed set of basis functions and a heuristic choice of the state-relevance distribution. Work relaxing these assumptions, as we do, is limited. \looseness=-1

\cite{klabjan2007infinite} is a seminal paper that develops a convergent algorithm to generate basis functions for semi-Markov decision processes. It requires the solution of a challenging nonlinear program. Building on this work, \citet{adelman2012GJR} considered an innovative algorithm for basis function generation in a generalized joint replenishment problem. Their algorithm leverages structure and numerical experience for this application. Our approach differs from this work because it uses low-cost sampling to generate basis functions, focuses on discounted-cost MDPs, and is application agnostic. \looseness=-1

\cite{bhat2012NonParaALP} side-stepped basis function selection when computing VFAs by applying the kernel trick (see, e.g., chapter 5 of \citealp{mohri2012foundations}) to replace the inner-products of such functions in the dual of a regularized ALP relaxation. Guarantees on the approximation quality of their VFAs depend on the kernel and an idealized sampling distribution that assumes knowledge of an optimal policy. Our approach instead works directly on the primal ALP formulation and samples over the parameters of a class of basis functions as opposed to state-action pairs. Moreover, the sampling distribution is readily available in our framework and the error bounds that we develop are not linked to the knowledge of an optimal policy. \looseness=-1

The papers above do not address the choice of the state-relevance distribution. Parametric forms for the state-relevance distribution that are close to the steady-state distribution of an optimal policy can be obtained for some queuing applications but not in general \citep{farias2003ALP}. The use of policies to choose this distribution in policy-guided FALP is based on an approach discussed in \citet[page 854]{farias2003ALP} and \citet{Farias2006}, a version of which is employed in \citet{sun2014quadratic}. Self-guided FALP, while iterative, is fundamentally different as it leverages the ability to cheaply sample new random basis functions and uses only past VFA information available from solving an ALP model to guide new VFAs; we demonstrate that this can be interpreted as modifying the state-relevance distribution. Along with the theoretical guarantees already discussed, one can view self-guided FALP as a conceptually sound mechanism for updating the state-relevance distribution. 

Overall, the exact representation of an MDP based on random bases, the FALP and self-guided FALP models, and their associated theoretical guarantees are novel. A useful property of our results is that they apply to MDPs with state spaces containing continuous and discrete elements. The implementation guidelines and our numerical study highlight how these developments can ease the use of ALP, while providing effective policies on two challenging applications. On the perishable inventory control problem, we close the optimality gaps of the prior application-specific policies on known instances and obtain near-optimal policies on much larger instances. For options pricing, self-guided ALP improves on least-squares Monte Carlo with problem-specific basis functions in terms of policy performance, which is encouraging. \looseness=-1


Our work builds on the seminal research on random bases by \citet{rahimi2008large,rahimi2008uniform} and \citealp{rahimi2009RKS}. There is extant literature applying this idea to data mining and machine learning applications (\citealp{lu2013faster}, \citealp{mcwilliams2013correlated}, \citealp{beevi2016detection}, and \citealt{wu2018scalable}) and to a value iteration algorithm by \cite{haskell2017empiricalDPwithRBF}. These papers embed random bases in what amounts to a regression setting, whereas we show that such bases can be effectively used in ALPs that have complicated constraints. We also added to this literature in terms of theory. Our approximation guarantees for FALP adapt the arguments in \citet{rahimi2008uniform} to an ALP setting and also strengthen the error bounds. Similar analysis of self-guided FALP, unfortunately, does not lead to insightful bounds. In this case, we develop error bounds based on functional projections, which is new to this literature, and potentially of independent interest. \looseness=-1

More broadly, our work adds to the rich literature on reinforcement learning that attempts to reduce the burden of feature engineering \citep{mnih2015human,silver2017mastering}. Here, neural networks and deep learning have received significant research attention as they facilitate the approximation of complex functions with limited domain knowledge 
\citep{fujimoto2018addressing,osband2019deep,franke2021sampleefficient}. They give rise to VFAs that depend nonlinearly on the parameters but involve the solution of non-convex optimization problems \citep{wang2020random}. Our use of random basis functions in ALP also mitigates domain knowledge but it retains linear programming structure and can thus be viewed as a complementary strategy. 

\subsection{Organization of Paper}\label{subsec:organization}
In \S\ref{sec:Exact Linear Programs}, we present the standard linear programming approach to solve MDPs and then introduce an alternative approach that employs random basis functions. In \S\ref{sec:Approximate Linear Programs with Random Basis Functions}, we discuss FALP. In \S\ref{sec:Greedy Policy Guided ALPs}, we present iterative FALP-based approaches: policy-guided FALP and self-guided FALP. In \S\ref{sec:extensions}, we present extensions to discrete-state MDPs and finite-horizon MDPs. The numerical studies on perishable inventory control and options pricing are in \S\ref{sec:Perishable Inventory Control} and \S\ref{sec:Option-Pricing}, respectively. We conclude in \S\ref{sec:Conclusions}. All proofs can be found in an electronic companion. Python code accompanying this paper can be found at \url{https://github.com/Self-guided-Approximate-Linear-Programs}.

\section{Exact Linear Programs}\label{sec:Exact Linear Programs}
In \S \ref{section:Background}, we provide background on infinite-horizon, discounted-cost MDPs and their known linear programming MDP reformulations. In \S \ref{sec:FELP}, we propose an alternative linear programming reformulation for MDPs based on random basis functions, which plays a central role in the approximations we consider in later sections.

\subsection{Background}\label{section:Background}

Consider a decision maker controlling a system over an infinite horizon. A policy $\pi:\sSpace\mapsto\aSpaceS$ assigns an action $a\in \aSpaceS$ to each state $s\in \sSpace$, where $\sSpace$ denotes the MDP state space and $\aSpaceS$ represents the feasible action space at state $s$. An action $a\in\aSpaceS$ taken at state $s \in \sSpace$ results in an immediate cost of $c(s,a)$ and the transition of the system to the next state according to the probability distribution $P(\cdot|s,a)$.  \looseness=-1

The decision maker's objective is to find a stationary and deterministic optimal policy $\pi$ that minimizes discounted expected costs. Starting from an initial state $s_0 = s \in\sSpace$, the discounted expected cost of a policy $\pi$ is 
\begin{equation*}
	\polCost(s,\pi) \coloneqq \expt \Bigg[ \sum_{t=0}^{\infty} \gamma^t c(s^{\pi}_t,\pi(s^{\pi}_t)) \ \bigg | \ s_0 = s\Bigg],
\end{equation*}	
where $\gamma \in [0,1)$ denotes the discount factor, expectation $\expt$ is with respect to the state-action probability distribution induced by the transition probability distribution $P(\cdot|s,a)$ and the policy $\pi$, and $s^{\pi}_t$ is the state reached at stage $t$ when following this policy. The quality of a given policy is evaluated with respect to a distribution $\chi(s)$ for the initial state. Specifically, we define the cost of policy $\pi$ as $\mathrm{PC}(\pi) \coloneqq  \mathbb{E}_{\chi}[\polCost(s,\pi)]$. 

The policy-cost minimization problem is
\begin{equation}\label{eq:minCostMDP}
	\underset{\pi:\sSpace\mapsto\aSpaceS}{\inf} \ 		\mathrm{PC}(\pi).
\end{equation}

\begin{assumption}\label{asm:MDP}
	An optimal policy $\pi^* \in \Pi$ that solves \eqref{eq:minCostMDP} exists and the MDP value function $V^*:\sSpace\mapsto\R$ satisfies $V^*(\cdot) \coloneqq \polCost(\cdotp,\pi^*)$. The state space $\sSpace$ is a continuous, compact real-valued set and the action spaces $\aSpaceS$ for all $s \in \sSpace$ either share this property or are finite. Moreover, the MDP value function $V^*(\cdot)$ is continuous.
\end{assumption}
The existence of $\pi^*$ in the literature is guaranteed under different requirements, mainly over the cost function $c(\cdot,\cdot)$ and state transition kernel $P(\cdot|s,a)$. Informally, one such set of conditions requires the lower semi-continuity of the immediate cost and the strong continuity of state transitions. We present them formally in \S\ref{subsec:SumAssump} and refer to Theorem 4.2.3 in \citet{hernandez1996discrete} for a more elaborate discussion. Continuous state spaces and value functions arise in applications such as lost-sales inventory control \citep{zipkin2008structure}, healthcare screening \citep{steimle2017markov}, dual sourcing \citep{hua2015structural}, robotics \citep{peters2003reinforcement,haarnoja2018learning}, and flight simulators \citep{mcgrew2010air,yang2019maneuver}. Our models and analysis in the remainder of this section and \S\S\ref{sec:Approximate Linear Programs with Random Basis Functions}--\ref{sec:Greedy Policy Guided ALPs} focus on MDPs satisfying Assumption \ref{asm:MDP}. We discuss in \S\ref{sec:extensions} how they apply to a broader class of MDPs, for instance, where the state space can have discrete components.\looseness=-1

The computation of the value function can be conceptually approached without knowing $\pi^*$ via the exact linear program (ELP; see, e.g., pages 131-143 in \citealp{hernandez1996discrete})
\begin{align} \setlength{\jot}{10pt}
	\max_{V':\sSpace\mapsto\R}   		\quad  & \expt_{\nu} \big[V'(s) \big ]\nonumber\\
	\text{s.t.}     \ \ \quad  & \hspace{18pt} V'(s)  \ - \ \gamma  \expt[V'(s^\prime) \  | \ s,a]    \ \le \  c(s,a), \quad \forall (s,a)\in \saSpaceS\label{constr:ELP},
\end{align} 
where $\nu$ is a {state-relevance distribution} that specifies the relative importance of each state in the state space. ELP is well defined because Assumption \ref{asm:MDP} ensures that the MDP value function $V^*$ solves the optimality equations $V^*(s) =  \min_{a\in\aSpaceS}\{c(s,a) + \gamma  \expt[V'(s^\prime) |  s,a] \}$ for every $s\in\sSpace$. Thus, $V^*$ is an optimal solution to ELP, which follows from its constraints holding as equalities at $V^*$. Since $V^*$ is continuous over a compact domain (Assumption \ref{asm:MDP}), it is bounded and the objective function of ELP, which is an expectation of $V^*$, is also bounded. However, ELP is  intractable to solve since it is a doubly infinite linear program. It has continua of decision variables and constraints, one for each state and state-action pair, respectively. \looseness=-1


\subsection{Feature-based Exact Linear Program}\label{sec:FELP}

To be able to approximate ELP, we present a reformulation of it below that relies on a class of random basis functions defined by a vector $\theta\coloneqq(\theta_0,\theta_1,\dots,\theta_\sDim) \in \Theta\subseteq\R^{\sDim+1}$ and an associated sampling density $\rho(\theta)$, where  integer $\sDim$ denotes the dimension of the state space $\sSpace$. Consider scalar mapping $\varphi(\cdot): \R \mapsto \R$. This mapping can be used to represent a random basis function over the state space $\sSpace$ as $\varphi\big(\theta^\top (1,s)\big)$ using the inner product $\theta^\top (1,s) := \theta_0+\sum_{i=1}^{\sDim}\theta_i s_i$. In other words, for a given $\theta$, we can define the basis function $\varphi(s;\theta) \equiv \varphi\big(\theta^\top (1,s)\big)$. These basis functions are referred to as random basis functions because $\theta$ is sampled using $\rho$. Table \ref{table:rand_basis} lists the components of three random basis functions: the mapping $\varphi(\cdot)$, the sampling density $\rho(\cdot)$ for the vector $\theta$, and the parameters defining this density. Fourier basis functions are defined using a cosine mapping with $\theta_0$ sampled from a uniform distribution with support involving the Archimedes constant $\pi$ and the remaining elements of $\theta$ sampled from a normal distribution with mean zero and standard deviation $\varrho$, which is a tunable scalar parameter. ReLU basis functions employ a mapping that is a maximum with respect to zero. It samples $\theta$ from a uniform distribution over a unit sphere with no tunable parameters. Stump basis functions use a signum mapping that evaluates to a $-1$, $0$, or $1$, depending on whether the input is negative, zero, or positive, respectively. The element $\theta_0$ is sampled from a uniform distribution with support over an interval that depends on a tunable scalar parameter $\varrho$. The remaining elements of $\theta$ are sampled from a uniform distribution defined on the discrete set $\{e^1,\dots,e^\sDim\}$, where $e^i$, $i \in \{1,2,\dots,d\}$ is a $d$-dimensional unit vector with 1 in the $i$-th coordinate and zero elsewhere. 

\begin{table}[t]
	\centering
	\setlength{\tabcolsep}{8pt}
	\renewcommand{\arraystretch}{1.3}
	\caption{Examples of universal random basis functions.}
	\adjustbox{width=\textwidth}{
		\begin{tabular}{lccc}
			\hline
			& $\varphi(\cdot)$  & $\rho(\theta)$ & Parameters \\
			\cline{2-4}
			Fourier     & $\cos(\cdot)$   
			& $\theta_0 \sim \mathrm{uniform}([-\uppi,\uppi]); \ \theta_i \sim \mathrm{normal}(0,\varrho)$, for $i\ge 1$  & $\varrho$ \\
			ReLU        & $\max(\cdot,0)$   
			& $\theta \sim \mathrm{uniform}(\text{$d$-dimensional unit sphere})$& None\\
			Stump       & $\mathrm{sign}(\cdot)$                     
			&{$\theta_0 \sim \mathrm{uniform}([-\varrho,\varrho])$; \ $(\theta_1,\dots,\theta_\sDim) \sim \mathrm{uniform}(\{e^1,\dots,e^\sDim\})$}&$\varrho$\\   
			\hline
	\end{tabular}
	}
	\label{table:rand_basis}
\end{table}

To approximate ELP, we consider random bases with known ``universal'' approximation power; that is, they can approximate continuous functions with arbitrary accuracy. Given a class of random bases, we define the following function over the state space $\sSpace$ using the pair $\coefInf\coloneqq(\beta_0,\coefInfBar)$ containing an intercept $\beta_0\in\R$ and an integrable weighting function $\coefInfBar : \Theta \mapsto \mathbb{R}$:
\begin{equation}\label{eq:integral-form}
V(s;\coefInf)\coloneqq \beta_0 +\int_\Theta \coefInfBar(\theta)\varphi(s;\theta)\diff \theta.
\end{equation}
A class of functions that can be covered by this construction is 
\[	\randBasisSet \coloneqq \left\{V: \sSpace \mapsto \mathbb{R} \ \Big |  \  \exists \coefInf=(\beta_0,\coefInfBar)  \ \mbox{ s.t. } \ V(s) = V(s;\coefInf), \ \forall s\in\sSpace, \ \text{ and } \ \sNorm{\coefInfBar/\rho}_{2,\rho} < \infty \right\}, \]
where the $(2,\rho)$-norm of ${{\coefInfBar(\cdot)}\big/{\rho(\cdot)}}:\Theta\mapsto\R$  is defined as
\[ \sNorm{\coefInfBar/\rho}_{2,\rho}\coloneqq \int_{\Theta} \left(\frac{\coefInfBar(\theta)}{\rho(\theta)}\right)^2 \rho(\diff \theta) = \int_{\Theta} \frac{(\coefInfBar(\theta))^2}{\rho(\theta)} \diff\theta. \]
When the random bases are universal, the class $\mathcal{R}$ contains a function that is arbitrarily close to any continuous function under an $\infty$-norm. Definition \ref{defn:randBasisFns} formalizes universality, a property satisfied by the examples in Table \ref{table:rand_basis}. For a continuous function $V:\sSpace\mapsto\R$, the $\infty$-norm is $\sNorm{V}_{\infty}\coloneqq\max_{s\in\sSpace}|V(s)|$. We also use the shorthand $V(\coefInf)\equiv V(\cdotp;\coefInf)$ for a function $V\in\randBasisSet$. 
\begin{definition}\label{defn:randBasisFns}
	\textit{A class of random basis functions $\varphi$ with sampling density $\rho$ is called universal if for any continuous function  $V:\sSpace\mapsto\R$ and $\varepsilon>0$, there exists  $\coefInf_{\varepsilon}\coloneqq(\beta_{0,\varepsilon},\coefInfBar_{\varepsilon})$ such that $V( \coefInf_{\varepsilon})\in\randBasisSet$ and $\sNorm{V- V(\coefInf_{\varepsilon})}_\infty<\varepsilon$.}
\end{definition}

Since the MDP value function $V^*$ is continuous (Assumption \ref{asm:MDP}), replacing it with the integral form \eqref{eq:integral-form} with universal basis functions should intuitively not result in any significant error. Performing this replacement and requiring the weighting function to have a finite norm as in the definition of $\mathcal{R}$ gives the following linear program:
\begin{equation}
	\begin{aligned}
		\sup_{\beta_0,\coefInfBar}  \quad  &  
		\beta_0 +\int_\Theta \coefInfBar(\theta)\expt_\nu[\varphi(s;\theta)]\diff \theta  && \nonumber\\
		\text{s.t.}  \quad 
		&    (1-\gamma)\beta_0 +   \int_\Theta \coefInfBar(\theta)\big(\varphi(s)  \ - \ \gamma  \expt[{\varphi({s^\prime})}  \ |  \ s,a] \big)\diff \theta     && \le c(s,a),  &&& \forall (s,a)\in \saSpaceS \\
		& \sNorm{\coefInfBar/\rho}_{2,\rho}  && <\infty. &&& \nonumber
	\end{aligned}
\end{equation}
Unlike ELP, which directly optimizes a value function, the above linear program optimizes the weights associated with the \emph{feature based} representation of the value function in the set $\randBasisSet$. Hence, we refer to it as the feature-based exact linear program (FELP). 

Proposition \ref{prop:V-optimal-to-FELP} states an easily verifiable relationship between an optimal FELP solution and $V^*$: \looseness=-1
\begin{proposition}\label{prop:V-optimal-to-FELP}
	If  $V^* \in \randBasisSet$, there is an optimal FELP solution $\coefFELPVec = (\coefFELP,\coefFELPVecBar)$ such that $V^*(s) =  \coefFELP +\int_\Theta \coefFELPVecBar(\theta)\varphi(s;\theta)\diff \theta$ for all $s\in\sSpace$.
\end{proposition}
When using universal random basis functions, the assumption $V^* \in \randBasisSet$ is mild. Specifically, if $V^*$ is continuous but not in $\mathcal{R}$, then an optimal FELP solution defines a function that is arbitrarily close to $V^*$ under an $(1,\nu)$-norm as we show formally in \S\ref{EC:opt-V-not-in-R}. Assumption \ref{asm:random basis function}, which holds for the rest of the paper, includes $V^* \in \randBasisSet$ and additional conditions needed for our theoretical analysis, all of which are standard in the random basis functions literature (see, e.g., \citealp[Theorem 3.2]{rahimi2008uniform}). 
\begin{assumption}\label{asm:random basis function}
	The MDP value function $V^*$ belongs to $\randBasisSet$. The class of random basis functions $\varphi$ is universal, and its sampling distribution $\rho$ has a finite second moment. Moreover, $\varphi$ has a Lipschitz constant $\lipConst>0$ and satisfies $\sNorm{\varphi}_\infty\le 1$ and  $\varphi(0) = 0$. 
\end{assumption}
This assumption is satisfied by Fourier and ReLU basis functions in Table \ref{table:rand_basis} but not by Stump basis functions as they are not continuous. While Assumption \ref{asm:random basis function} is needed for analysis, the algorithms we present in \S\S\ref{sec:Approximate Linear Programs with Random Basis Functions}--\ref{sec:Greedy Policy Guided ALPs} can be applied even when this assumption fails to hold. 

\section{Feature-based Approximate Linear Program}
\label{sec:Approximate Linear Programs with Random Basis Functions}

In \S \ref{sec:Model and Theory}, we introduce and analyze FALP, which is an approximation of FELP using random basis functions. In \S\ref{sec:FALP-Implementation}, we provide guidelines for the formulation and solution of FALP.

\subsection{Model and Theory}\label{sec:Model and Theory}
In the literature, an ALP is derived from ELP by substituting its decision variable $V'(s)$ by a linear combination of pre-specified basis functions. Our starting point is instead FELP. We replace the integral form \eqref{eq:integral-form} with a sampled VFA
\[V(s;\coef) \coloneqq \beta_{0}  + \sum_{i=1}^{N}\beta_i\varphi(s;\theta^i),\]
where $\theta^1, \theta^2, \ldots, \theta^N$ are iid samples of the basis function vector from  $\rho$ and $\coef$ is the finite weight vector $(\beta_0,\beta_1,\ldots,\beta_N)\in\R^{N+1}$. The weight $\beta_0$ represents an intercept as in FELP and the remaining elements of $\beta$ are weights associated with the random basis functions. In other words, $\beta_1, \beta_2, \ldots, \beta_N$ is the finite analogue of the weighting function $\coefInfBar$ in FELP and $V(s;\coef)$ can be viewed as an approximation constructed using a functional extension of Monte Carlo sampling applied to $V(s;\coefInf)$. The resulting ALP with $N$ random basis functions, denoted by \FALPprog{N}, is 
\begin{equation} 
\begin{aligned}    
\sup_{\coef}    \quad&   \beta_0 + \sum_{i=1}^{N}\beta_i\expt_{\nu} \big[\varphi(s;\theta^i) \big] && &&& \nonumber\\
\text{s.t.}    \ \quad&   (1-\gamma)\beta_0 + \sum_{i=1}^{N}\beta_i \left(\varphi(s;\theta^i)  - \gamma \expt \big[\varphi(s';\theta^i) \  | \ s,a\big]\right)   &&\le \ c(s,a), &&& (s,a) \in \saSpaceS.
\end{aligned}
\end{equation}
This model is a semi-infinite linear program with $N+1$ variables and an infinite number of constraints. We assume the existence of a solution to \FALPprog{N}. This is mild because we can always bound the absolute value of the elements of $\beta$ by a large constant to ensure the existence of a finite optimal solution without affecting our results. We show this formally in \S\ref{EC:FinitenessOfFALP}.
\begin{assumption}\label{finitenessOfFALPOpt}
A finite optimal solution to \FALPprog{N} exists. 
\end{assumption}
 
 Theorem \ref{prop:ALP} establishes key properties of \FALPprog{N} and relies on the constant
\[
\contractionFactor  \coloneqq 5(\diamSState+1) \lipConst  \sqrt{\mathbb{E}_\rho \big[\sNorm{\theta}_2^2\big]},
\]
where $\sNorm{\cdot}_2$ denotes the 2-norm, $\diamSState \coloneqq \max_{s\in\sSpace} ||s||_2$,  $\lipConst$ is the Lipschitz constant of random basis $\varphi(\cdot)$ defined in Assumption \ref{asm:random basis function}, and $\mathbb{E}_\rho$ is the expectation under the distribution $\rho$. Let $\coefALPVec\coloneqq (\coefALP{N,0},\ldots,\coefALP{N,N})$ represent an optimal solution to \FALPprog{N}. 


\begin{theorem}\label{prop:ALP} The following hold:
	\begin{itemize}
		\item[(i)] 
		For a given $N$, we have $V(s;\coefALPVec) \leq V^*(s)$ for all $s\in\sSpace$.
		\item[(ii)] 
		Suppose $\rho(\theta) \geq \underline{\rho}$ for all $\theta \in \Theta$. Given $\delta\in(0,1]$, we have that any finite optimal \FALPprog{N} solution $\coefALPVec$ satisfies \[{\tallNorm{V^* - \VALP}_{1,\nu} \ \leq \ \dfrac{2\tallNorm{\coefFELPVecBar/\rho}_{2,\rho}}{(1-\gamma)\underline{\rho}\sqrt{N}}\left(\contractionFactor+\sqrt{2\ln\left(\dfrac{1}{\delta}\right)}\right)
,}\] with a probability of at least $1-\delta$. 
	\end{itemize}
\end{theorem}
Part (i) of this theorem shows that \FALPprog{N} is well defined and provides a lower bound on the MDP value function $V^*$ at all states. The latter is a known result in approximate linear programming (see, e.g., \S2 in \citealp{farias2003ALP}). Part (ii) establishes a high probability $(1,\nu)$-norm error bound for this VFA. This bound decreases at the dimension-independent rate of $1/\sqrt{N}$ akin to Monte Carlo sampling, which is encouraging. The magnitude of the error increases only logarithmically as a more stringent probability guarantee is needed, that is, $\delta$ is decreased, and its growth with the dimension of the state space is captured in $\Omega$. As is the case with Monte Carlo sampling, this suggests that more random basis function samples are needed to approximate value functions over higher-dimensional state spaces. Indeed, the nature of the MDP value function $V^*$ also affects the error and this factor is signaled by the presence of the term $\tallNorm{\coefFELPVecBar}_{2,\rho}$ in the error bound. When the representation of $V^*(\cdot)= \coefFELP +\int_\Theta \coefFELPVecBar(\theta)\varphi(\cdotp;\theta)$ is not unique, one can select $(\coefFELP,\coefFELPVecBar)$ such that norm $\tallNorm{\coefFELPVecBar}_{2,\rho}$ is minimized and this minimum can be viewed as the approximation difficulty associated with $V^*$ when using a class of random basis functions. The condition in Theorem \ref{prop:ALP}(ii) of $\rho(\cdot) \geq \underline{\rho}$ is needed to avoid a situation where random basis functions with a certain set of $\theta$ values are needed to approximate the value function well but are not sampled because $\rho(\cdot)$ is zero in this set. This requirement is fairly mild. Sampling distributions with bounded support (e.g., uniform) clearly satisfy it. Since $N$ is finite, distributions with support over an unbounded set, such as the normal distribution, satisfy it with high probability because the sampled $\theta$ vectors highly likely come from a truncated version of the distribution, which has bounded support.\looseness=-1\label{lower-bound-on-rho}



The error bound in Theorem \ref{prop:ALP} extends to ALP the random basis function sampling results in \citet{rahimi2008uniform}, which proposes a functional form of Monte Carlo sampling in the regression setting and assumes knowledge of the function being approximated. First, we contend with the feasibility of VFA weights, which is possible because a given infeasible solution to ALP can be made feasible by appropriately scaling the intercept $\beta_{0}$. Second, a guarantee on the $(1,\nu)$-norm distance between $\VALP$ and $V^*$ is intuitively possible without the knowledge of $V^*$ because \FALPprog{N} is known (see Lemma 1 in \citealp{farias2003ALP}) to be equivalent to 
\begin{equation}\label{eqn:RegFormOfALP}
\begin{aligned}
\min_{\coef} & \quad \sNorm{V(\coef) - V^*}_{1,\nu}\\ 
\mbox{ s.t. } & \quad  \ V(s;\coef) -\gamma\expt\big[V(s^\prime;\coef) \ | \ s,a\big] \  \leq \ c(s,a), \qquad \forall (s,a)\in\saSpaceS.
\end{aligned}
\end{equation} 
Third, have a $(2,\rho)$-norm involving $\coefFELPVecBar$ in our error bound, which improves on an $\infty$-norm variant of this term in the original bound of \citet{rahimi2008uniform}, because we employ a solution construction in the proofs that differs from the one used in that paper. 

The utility of a VFA is that it can be used to obtain a policy. Given VFA weights $\beta\in\R^{N+1}$, we can define a so-called greedy policy $\pi_g(\coef)$ associated with $V(\beta)$ (see, e.g., \citealp{powell2007ADP}). The action $\pi_g(s; \beta)$ taken by this policy at state $s \in \sSpace$ solves 
\begin{equation}\label{eqn:GreedyOpt}\min_{a\in\aSpaceS} \Big \{ c(s,a) + \gamma\expt \big[V(s^\prime;\beta) \ | \ s,a \big] \Big \}.
\end{equation}  
The cost of the greedy (feasible) policy, which we denote by $\mathrm{PC}(\coef)$, is an upper bound on the optimal policy cost.

\subsection{Implementation Guidelines}\label{sec:FALP-Implementation} 

The literature contains three general strategies to solve \FALPprog{N}, which we overview. The most commonly used approaches are constraint generation and constraint sampling (\citealt{adelman2012GJR, farias2004constraintSampling}), which both rely on solving a relaxation of a \FALPprog{N}, but differ in how they construct the relaxation. Constraint generation works in an iterative fashion and starts with a subset of \FALPprog{N} constraints. Given an optimal solution of this relaxation, it identifies the most-violated ALP constraint, if any, adds this constraint, and repeats the procedure until no violated constraint is identified, at which point the incumbent solution is optimal to \FALPprog{N}. In principle, constraint generation can be used to obtain a near-optimal ALP solution and a lower bound on the optimal policy cost. The computational feasibility of this approach depends on the separation problem. Examples of its success in the literature rely on formulating this separation problem as a linear, convex, or mixed integer program \citep{adelman2004priceDirectedInventory,zhang2009NetRevMngmt,adelman2012GJR}. Constraint sampling instead constructs a random relaxation of \FALPprog{N} by sampling a finite number of its constraints. It is easy to implement but its performance depends on the constraint sampling distribution and it does not directly provide a lower bound on the optimal policy cost. A more recent approach for tackling \FALPprog{N} employs a saddle-point reformulation and applies a first order method to it. This approach is referred to as constraint violation learning \citep{lin2017ContViolLearning}.\looseness=-1

We outline in more detail a hybrid approach that uses constraint sampling to solve \FALPprog{N}, as it is the easiest to implement, and then relies on another approach to obtain a lower bound on the optimal policy for benchmarking. The key step is to replace the set of constraints in \FALPprog{N} with a subset obtained by sampling $K$ iid state-action pairs $\{({s}^k,{a}^k)\in\saSpaceS : k=1,2,\dots,K\}$ from a probability distribution $\psi$ over the state-action space $\saSpaceS$ \citep{calafiore2005uncertain}. The result is the following linear program with $N$ random basis functions and $K$ constraint samples:
\begin{align}    
\max_{\coef}    \quad&   \beta_0 + \sum_{i=1}^{N}\beta_i\expt_{\nu} \big[\varphi(s;\theta^i) \big] && &&&\label{sampleFALP1} \\
\text{s.t.}    \ \quad&   (1-\gamma)\beta_0 + \sum_{i=1}^{N}\beta_i \left(\varphi(s^k;\theta^i)  - \gamma \expt \big[\varphi(s';\theta^i) \  | \ s^k,a^k\big]\right)   &&\le \ c(s^k,a^k), &&& k=1,2,\dots,K.\nonumber
\end{align}
Proposition \ref{EC:prop:FALP-constr-sample} is an application of a key result in \cite{calafiore2006scenario} and shows that the linear program \eqref{sampleFALP1} for large enough $K$ provides a good randomized approximation of \FALPprog{N}. 
\begin{proposition}[Theorem 1 in \citealp{calafiore2006scenario}]\label{EC:prop:FALP-constr-sample}
        Given $\delta\in(0,1]$, if $\psi$ is supported over \saSpaceS, linear program \eqref{sampleFALP1} is bounded, and 
		\[K\ge  \left \lceil \frac{2}{\delta}\ln(\frac{1}{\delta})  + 2(N+1)   + \frac{2(N+1)}{\delta}\ln(\frac{2}{\delta}) \right \rceil,\]
		then for every optimal solution $\hat\beta$ to this program, the following inequality holds 
		\[
		    \psi\left(\left\{ (s,a)\in\saSpaceS \ : \  (1-\gamma)\hat\beta_0 + \sum_{i=1}^{N}\hat\beta_i \left(\varphi(s;\theta^i)  - \gamma \expt \big[\varphi(s';\theta^i) \  | \ s,a\big]\right)   \le  c(s,a) \right\}  \right) \ge 1-\delta,
		\]
		with a probability of at least $1-\delta$.
\end{proposition}
In particular, this proposition shows that as more samples are added the set of states where the FALP constraints are violated when measured using $\psi$ is at most $\delta$ and this holds with a probability of at least $1-\delta$. Therefore, if one solves the constraint-sampled version of FALP \eqref{sampleFALP1} with a large number of samples $K$, we expect the results in Theorem \ref{prop:ALP} to hold approximately. 

A sharper constraint sampling result specific to ALP can be found in \citet[Theorem 3.1]{farias2004constraintSampling} when $\psi$ is chosen using information from the optimal policy, which is unknown. During implementation, $\psi$ can be a uniform distribution or based on states visited by a baseline policy. Expectations in \eqref{sampleFALP1} are typically replaced by sample average approximations.  The number of constraint samples $K$ can be chosen so that the optimal objective function of \eqref{sampleFALP1} does not increase significantly as more samples are added. The optimal solution $\hat\beta$ to \eqref{sampleFALP1} defines a VFA $V(\hat\beta)$, which can be used to obtain a greedy policy. To obtain a lower bound for benchmarking, one could use $\hat\beta$ as a starting point in an approach to solve to \FALPprog{N} that does provide a bound, such as constraint generation or constraint violation learning (see \ref{ec:sec:A Lower Bound Estimator for Constraint-sampled ALPs}). Alternatively, one could use the VFA defined by $\hat\beta$ in the information relaxation and duality approach to generate lower bounds  (see \citealt{brown2021information} for details).  


The quality of the VFA obtained using the above procedure depends on how FALP is formulated, in particular, the number of basis function samples $N$, the choice of random basis functions, and the state relevance distribution $\nu$. We provide some guidance on these choices next.

Similar to standard Monte Carlo sampling, the value of $N$ depends on the computational budget. That is, one determines the largest $N$ for which the sampled version \eqref{sampleFALP1} of \FALPprog{N} can be tackled within a reasonable time limit (and possibly memory limit) using an off-the-shelf commercial solver. The ability of getting good VFAs with a small number of basis functions $N$ is thus an important consideration in choosing random basis functions. While multiple universal random basis functions guarantee the same theoretical convergence rate, their empirical rates may differ. A good starting point is to consider Fourier random basis functions (see Table \ref{table:rand_basis}), as they are known to provide better approximations as the continuous function becomes smoother (please see \S 2.1.1 of \citealt{canuto2012spectral} and \citealt{nersessian2019fourier} for recent examples). The non-smoothness of the MDP value function in several applications is localized, that is, even these value functions are smooth in most neighborhoods. Given a choice of random basis functions, the tunable parameters are a few and do not depend on the application. The random bases examples in Table \ref{table:rand_basis} have at most one such parameter. Thus, we recommend using cross-validation with the goal of minimizing the objective function of FALP for small values of $N$ to determine the parameter values of random basis functions. Fourier basis functions, which depend on a single bandwidth parameter, tuned using the aforementioned simple cross-validation strategy, worked well in both applications in our numerical experiments. The second decision factor could be computational. For example, in some applications, choosing an appropriate random basis function may facilitate the use of constraint generation. This is the case in the generalized joint replenishment studied in \citet{adelman2012GJR}. The separation problem in this application can be formulated as a mixed-integer program when employing random stump basis functions, which are piecewise constant.\looseness=-1


The state-relevance distribution $\nu$ plays an important role in linking the quality of the VFA to greedy policy performance \citep{farias2003ALP,desai2009smoothed,sun2014quadratic}. Proposition \ref{prop:worst case policy performance} formalizes this link using the state-visit frequency $\visitFreq{\chi}(\coef)$ of a greedy policy, which defines the following probability of visiting a subset of states $\sSpace_1\subseteq\sSpace$ (see, e.g., pages 132--133 in \citealt{hernandez1996discrete}): 
\begin{equation}\label{eqn:greedyPolicyVisitFrequency}
\visitFreq{\chi}(\sSpace_1;\coef) \coloneqq \chi(\sSpace_1) \ + \ \sum_{t=0}^{\infty}\gamma^{t+1} \expt\Big[ P\big(s^{\pi_g(\coef)}_{t+1}\in\sSpace_1  \  |  \ s_t,\pi_g(s_t;\coef)\big) \Big],
\end{equation}
where state $s^{\pi_g(\coef)}_{t+1}$ and transition probability distribution $P$ retain their definitions from \S\ref{sec:Exact Linear Programs}, and $\chi(\sSpace_1)$ is the probability of the initial state belonging to $\sSpace_1$. The expectation $\expt$ is taken with respect to control policy $\pi_g(\coef)$ and initial state distribution $\chi$ over initial state $s_0$. 

\begin{proposition}[Theorem 1 in \citealt{farias2003ALP}]\label{prop:worst case policy performance}
		For a VFA $V(\coef)$ such that $V(\coef) \le V^*$, we have \[\mathrm{PC}(\coef) - \mathrm{PC}(\pi^*) \ \le \ \frac{\sNorm{V(\coef) -V^*}_{1,  \mu_\chi(\coef)}}{1-\gamma}.\]
	\end{proposition}
	Proposition \ref{prop:worst case policy performance} shows that for a VFA $V(\coef)$ that lower bounds $V^*$ (e.g., the FALP VFA), the additional cost incurred by using the greedy policy $\pi_g(\coef)$ instead of the optimal policy $\pi^*$ is bounded above by the $(1,\visitFreq{\chi}(\coef))$-norm difference between the VFA $V(\coef)$ and the MDP value function $V^*$. This result motivates the search for good VFAs. 

If $\nu$ and $\mu_\chi(\coefALPVec)$ are identical, Proposition \ref{prop:worst case policy performance} and the reformulation \eqref{eqn:RegFormOfALP} imply that a FALP VFA with a small $(1,\nu)$-norm error also guarantees good greedy policy performance. However, one does not know $\mu_\chi(\coefALPVec)$ before solving FALP, which makes this choice challenging \citep{farias2003ALP}. Heuristics in the literature can be interpreted as approximating the expression \eqref{eqn:greedyPolicyVisitFrequency} for $\mu_\chi(\coefALPVec)$. The first approach sets $\nu$ equal to the initial state distribution $\chi$, which ignores the second term in \eqref{eqn:greedyPolicyVisitFrequency} that captures the effect of states visited by the policy in the future. This observation motivates the second strategy, in which a baseline policy $\pi$ is simulated to approximate the aforementioned second term. That is, $\pi$ replaces the greedy policy $\pi_g(\coef)$ in \eqref{eqn:greedyPolicyVisitFrequency}. The effectiveness of this approach depends on how close the states visited by the simulated heuristic policy overlap with those of good greedy policies. The third approach chooses $\nu$ to be a uniform distribution, which can be interpreted as acknowledging that we do not have any information about $\nu$. The effectiveness of these approaches will need to be tested numerically.

\section{{Guided Feature-based Approximate Linear Programs}}\label{sec:Greedy Policy Guided ALPs}

In this section, we discuss iterative strategies to mitigate the impact of the initial state-relevance distribution choice in FALP. In \S\ref{sec:state-rel-guide-FALP} and \S\ref{sec:Self-guided Approximate Linear Programs}, we describe such existing and new strategies, respectively. We take Assumption \ref{finitenessOfFALPOpt} to hold in these subsections and then discuss implementation and solution guidelines in  \S\ref{sec:implementation-guidelines-gided-FALP}.

\subsection{Policy-guided FALP}\label{sec:state-rel-guide-FALP}

We summarize in Algorithm \ref{alg:state-rel-guide-FALP} a procedure described in \citet{farias2003ALP} and \citet{Farias2006} that uses policies to guide the choice of the state-relevance distribution. To faciltate exposition, we make the dependence of \FALPprog{N} on $\nu$ explicit by writing \stateRelFALPprog{N}{$\nu$}. The procedure starts by solving \stateRelFALPprog{N}{$\nu^0$} based on an initial state-relevance distribution choice $\nu^0$ to obtain the VFA weights $\beta^0$. Then, it simulates the greedy policy $\pi_g(\beta^0)$ to obtain the state-visit distribution. This distribution is chosen as the new state-relevance distribution $\nu^1$. Iteration 1 starts by solving \stateRelFALPprog{N}{$\nu^1$} and so on. A total of $Q$ iterations are performed, after which the VFA weight vector $\beta^{Q-1}$ is returned. We refer to Algorithm \ref{alg:state-rel-guide-FALP} as policy-guided FALP. As Algorithm \ref{alg:state-rel-guide-FALP} iterates, one hopes that the state-relevance distribution overlaps more with states visited by good greedy policies but there is no gurantee that this will happen. \looseness=-1

    \setlength{\algomargin}{7pt}
    \begin{algorithm}[h!]
    	\DontPrintSemicolon
    	\SetAlgoLined
    	\KwIn{%
    		number of random basis functions $N$, random basis function $\varphi$ with sampling density $\rho$, initial state-relevance weight distribution $\nu^0$, and maximum number of iterations $Q$. 
    	}
    	\KwInit{%
    	formulate \stateRelFALPprog{N}{$\nu^0$} using $\nu^0$ and random basis functions in the class $\varphi$ based on $N$ iid $\theta$ vector samples from $\rho$.
    	}
    	        \For{$q= 0, 1,\dots, Q-1$}{\vspace{5pt}
    		     (ii)   Solve \stateRelFALPprog{N}{$\nu^q$} to obtain VFA weights $\beta^q$.\;
    		     (iii)  Simulate greedy policy $\pi_g(\beta^q)$ to estimate $\visitFreq{\chi}(\beta^q)$, and then set 
    		            $\nu_{q+1} \gets \visitFreq{\chi}(\beta^q).$\;
            	 }\vspace{6pt}
            	    	
    	\KwOut{%
    	VFA weights $\beta^{Q-1}$.
    	}
    	\caption{\normalfont{Policy-guided FALP}}
    	\label{alg:state-rel-guide-FALP}
    \end{algorithm}
%

\subsection{Self-guided Approximate Linear Programs}\label{sec:Self-guided Approximate Linear Programs}

   We present a new iterative scheme that leverages our ability to inexpensively sample new random basis functions to guide the state-relevance distribution. Specifically, this scheme gradually increases the number of basis functions in FALP by sampling new batches of random basis functions of size $B$ and adds guiding constraints to FALP that link the VFAs across consecutive iterations. For a given $N$, we refer to this modification of \FALPprog{N} as \FGLPprog{N}. We first present the modified model and explain its interpretation as a mechanism to update the state-relevance distribution, before formalizing an algorithm and providing theoretical support.

	
	Denoting by $\coefSGVecK{N-B}$ an optimal solution to \FGLPprog{N-B}, the model \FGLPprog{N} is
	\begin{align}    
		\max_{\coef}    \ \ &   \beta_0 + \sum_{i=1}^{N}\beta_i\expt_{\nu} \big[\varphi(s;\theta^{i}) \big] && \nonumber\\
		\text{s.t.}    \quad &   (1-\gamma)\beta_0 + \sum_{i=1}^{N}\beta_i \left(\varphi(s;\theta^{i})  - \gamma \expt \big[\varphi(s';\theta^{i}) | \ s,a\big]\right) \ \le  \ c(s,a), && \forall (s,a)\in\saSpaceS, \label{FALPConst1}\\
		& \beta_0 + \sum_{i=1}^{N}\beta_i\varphi(s;\theta^{i}) \ \ge \ V\big(s;\coefSGVecK{N-B}\big), && \forall s\in\sSpace.\label{FALPConst2}
	\end{align}
    The only difference between \FGLPprog{N} and \FALPprog{N} is that the former linear program includes additional constraints \eqref{FALPConst2} that require its VFA to be a state-wise upper bound on the VFA $V\big(s;\coefSGVecK{N-B}\big)$, which is computed in the previous iteration by solving \FGLPprog{N-B}. The VFA weights $\coefSGVecK{N-B}$ are feasible to \FGLPprog{N}. We assume $V\big(s;\coefSGVecK{N-B}\big) = -\infty$ for all $s \in \sSpace$ when $N = B$, which implies that the constraints \eqref{FALPConst2} are redundant in the first iteration. Dualizing these additional constraints provides insight into  \FGLPprog{N}. Let $y^*(s)\ge0$ denote the optimal dual value associated with the constraint \eqref{FALPConst2} at state $s \in \sSpace$. Define a state-relevance distribution $\nu'$ that evaluates at this state to
    \[\nu'(s) \coloneqq \frac{\nu(s) + y^*(s)}{1+\int_{\sSpace}y^*(s)\diff s}.\]
    If strong duality holds, then we have that an optimal solution of \FGLPprog{N} solves 
    \[\max_{\beta} \beta_0 + \sum_{i=1}^{N}\beta_i\expt_{\nu'} \big[\varphi(s;\theta^{i}) \big] \quad \mbox{s.t.} \quad \eqref{FALPConst1}.\]
The main takeway from this reformulation is that \FGLPprog{N} can be viewed as modifying the \FALPprog{N} state-relevance distribution using its own past VFA information, that is, the \FGLPprog{N-B} VFA. Thus, we refer to the iterative scheme involving \FGLPprog{N}, which is summarized in Algorithm \ref{alg:self-guide-FALP}, as self-guided FALP. For brevity, we do not discuss the technical conditions for strong duality here (see, e.g., \citealp[Theorem 2.3]{shapiro2009semi}, and \citealt[Theorem 4.1]{basu2017strong}) because the constraints will be sampled during implementation, in which case standard strong duality for finite linear programs will apply.  \looseness=-1

    \setlength{\algomargin}{7pt}
    \begin{algorithm}[h!]
    	\DontPrintSemicolon
    	\SetAlgoLined
    	\KwIn{%
    	sampling batch size $B$, random basis function $\varphi$ with sampling density $\rho$, initial state-relevance weight distribution $\nu^0$, and maximum number of iterations $Q$.
    	}
    	\KwInit{%
    	the number of random bases $N$ to $0$ and the set $\vartheta$ of sampled $\theta$ vectors to $\{\}$.
    	}
    	 \For{$q= 0, 1,\dots, Q-1$}{\vspace{5pt}
    	        (i) Draw $B$ iid samples $\{\theta^1,\ldots,\theta^B\}$ from $\rho(\theta)$, update $\vartheta \gets \vartheta \cup\{\theta^1,\ldots,\theta^B\}$, and set $N \gets N + B$.\;
    		    (ii) Compute coefficients $\coefSGVecK{N}$ by solving \FGLPprog{N} formulated using random basis functions with parameters in set $\vartheta$, the state-relevance distribution $\nu^0$, and the VFA $V(\coefSGVecK{N-B})$.\;
    	 }\vspace{6pt}
    	\KwOut{%
    	coefficients $\coefSGVecK{N}$.
    	}
    	\caption{\normalfont{Self-guided FALP}}
    	\label{alg:self-guide-FALP}
    \end{algorithm}
    
   The inputs to Algorithm \ref{alg:self-guide-FALP} are similar to Algorithm \ref{alg:state-rel-guide-FALP}, except for the batch size $B$, which replaces the apriori fixed number of basis functions $N$ across iterations. At each iteration, Algorithm \ref{alg:self-guide-FALP} (i) samples a batch of $\theta$ vectors of size $B$ and includes them in the current set $\vartheta$ of such vectors and increases the basis function count $N$ by $B$, and (ii) solves a revised \FGLPprog{N} model formulated with these additional random basis functions and the VFA computed in the previous iteration. After $Q$ iterations, it returns the VFA weights $\coefSGVecK{N}$, where $N = QB$.  
%

	Proposition \ref{prop:SG-ALP-basic} establishes a key property of the VFAs generated by Algorithm \ref{alg:self-guide-FALP}.
	\begin{proposition}\label{prop:SG-ALP-basic}
		For any integer $n\ge1$, it holds that
		\begin{equation}\label{FGLPVFAOrdering}
            V(s; \coefALPVecN{B}) \ = \ V(s; \coefSGVecK{B}) \ \le \ V(s; \coefSGVecK{2B}) \ \le \ \cdots \ \le \ V(s; \coefSGVecK{nB})  \ \le \ V^*(s), \quad \forall s\in\sSpace.
		\end{equation}
	\end{proposition} 
	The equality in \eqref{FGLPVFAOrdering} follows from our assumption that $V\big(\cdot;\coefSGVecK{0}\big) = -\infty$. The relationship $V(s; \coefSGVecK{\bar{n}B}) \leq V^*(s)$ holds for all $s \in \sSpace$ and $\bar n \in \{1,\ldots,n\}$ by Part (i) of Theorem \ref{prop:ALP} because $\coefSGVecK{\bar{n}B}$ is feasible to \FGLPprog{\bar{n}B}; thus, it is also feasible to \FALPprog{\bar{n}B}. The inequalities of the type $V(s; \coefSGVecK{N-B}) \ \le \ V(s; \coefSGVecK{N})$ are directly implied by the self-guiding constraints \eqref{FALPConst2}.

An important consequence of Proposition \ref{prop:SG-ALP-basic} is that Algorithm \ref{alg:self-guide-FALP} generates a sequence of VFAs that draws (weakly) closer to $V^*$ at all states. Therefore, two consecutive VFAs  with $N - B$ and $N$ random basis functions satisfy 
	\[\sNorm{V(\coefSGVecK{N}) -V^*}_{1,  \mu} \leq \sNorm{V(\coefSGVecK{N-B}) -V^*}_{1,  \mu},\]
	for any proper distribution $\mu$ defined over the state space and, in particular, when $\mu$ is the state-visit frequency $\mu_{\nu}(\coefSGVecK{N})$ associated with the greedy policy ${\pi_g(\coefSGVecK{N})}$.
	As a result, for any fixed iteration index $\bar{n}$ and its corresponding state-visit frequency $\mu_\chi(\coefSGVecK{\bar{n} B})$, it follows that the sequence of VFAs $V(\coefSGVecK{B}), V(\coefSGVecK{2B}), \ldots, V(\coefSGVecK{nB}),\ldots$ generated by Algorithm \ref{alg:self-guide-FALP} improves the worst-case performance bound of Proposition \ref{prop:worst case policy performance}, that is, $\sNorm{V(\coefSGVecK{n B}) -V^*}_{1, \mu_\chi(\coefSGVecK{\bar{n} B})}$ is non-increasing in $n$. These results, together with our Lagrangian reformulation of \FGLPprog{N}, show that self-guided FALP embeds a mechanism to adaptively update the state-relevance distribution such that a worst-case performance of their greedy policies is (weakly) improving. To understand this mechanism, recall the regression reformulation of ALP in \eqref{eqn:RegFormOfALP}, which shows that \FALPprog{N} considers a candidate set of VFAs that satisfy constraints \eqref{FALPConst1} and chooses its VFA as the one in this set that minimizes the $(1,\nu)$-norm with respect to $V^*$. The guiding constraints \eqref{FALPConst2} impose the additional condition  that the so computed VFA cannot worsen the $\infty$-norm distance to $V^*$ of the most recently computed VFA. Thus, from a VFA error minimization perspective, self-guiding FALP can be seen as using iteration to guard against $(1,\nu)$-norm improvements leading to a worsening of the $\infty$-norm distance to $V^*$. 

Studying the quality of the sequence of VFAs generated by Algorithm \ref{alg:self-guide-FALP} is challenging because consecutive VFAs in this sequence are coupled by the guiding constraints \eqref{FALPConst2}. Given the VFA $V\big(\coefSGVecK{N}\big)$ generated by solving \FGLPprog{N}, we bound the $\infty$-norm error of a VFA $V\big(\simpleVecK{N + H}\big)$ that is constructed with $H$ additional random basis functions, is feasible to constraints \eqref{FALPConst1}, and near feasible to \eqref{FALPConst2}. The techniques used to obtain a sampling bound for \FALPprog{N} in Theorem \ref{prop:ALP} (understandably) do not factor in the effect of $V\big(s;\coefSGVecK{N}\big)$ and, thus, do not provide a useful error bound of the type we require here (please see Online Supplement \S\ref{ec:sec:Analyzing an FALP-based Sampling Bound for FGLP} for details). Therefore, we develop a new projection-based analysis.
	

Consider the set of functions spanned by an intercept plus a linear combination of $N$ random basis functions in set $\Phi_N := \{\varphi(\cdot;\theta^1), \varphi(\cdot;\theta^2),\ldots, \varphi(\cdot;\theta^N)\}$:
    \[\mathcal{W}(\Phi_{N}) \coloneqq \bigg\{ V\in \randBasisSet \ \Big | \exists \ (\beta_0, \beta_1,\dots,\beta_N)\in \R^{N+1} \mbox{ s.t. } V(\cdot) = \beta_{0}  + \sum_{i=1}^{N}\beta_i\varphi(\cdot;\theta^i) \bigg\}.\]
    A strategy to account for the impact of $V(\coefSGVecK{N})$ on $H$ is to ask if $V^*$ is a part of the functional space $\mathcal{W}(\Phi_{N})$ containing $V(\coefSGVecK{N})$. If $V^* \in \mathcal{W}(\Phi_{N})$, then it would not be possible to improve the incumbent VFA $V(\coefSGVecK{N})$ via additional sampling. If $V^* \not \in \mathcal{W}(\Phi_{N})$, then $V^*$ intuitively has a (projected) component in the functional space $\mathcal{W}(\Phi_{N})$, as well as a nonzero (projected) component in the orthogonal complement of this space. We then bound the approximation error pertaining to this orthogonal component as $H$ increases, which allows us to bound the error of a VFA in $\mathcal{W}(\Phi_{N} \cup \Phi_H)$ that is feasible to \FGLPprog{N+H}. 

Formally, we decompose $V^*$ into a component that belongs to $\mathcal{W}(\Phi_{N})$ and a residual in an orthogonal complement space. Such decomposition is possible because the closure of $\randBasisSet$ is a Hilbert space (by Proposition 4.1 in \citealt{rahimi2008uniform}), where an orthogonal decomposition is well defined (by Theorem 5.24 in \citealt{folland1999real}). Because $V^*(\cdot) =  \coefFELP +\int_\Theta \coefFELPVecBar(\theta)\varphi(\cdotp;\theta)\diff \theta$ by Assumption \ref{asm:random basis function},  we can decompose it into $V(\optVproj)$ and $V(\optVprojPerp)$, that is, $V^* = V(\optVproj) + V(\optVprojPerp)$, where $V(\optVproj)$ and $V(\optVprojPerp)$ are projections of $V^*$ on to $\mathcal{W}(\Phi_{N})$ and its orthogonal complement (to be precise, the projections are performed on to the closures of these sets). Based on this construction, we have the below theorem. For given positive integers $N$ and $H$, define the constant 
    \[
            E\programIndex{N,H}\coloneqq \frac{2\tallNorm{\optVprojPerpBar/\rho}_{2,\rho}}{(1-\gamma)\underline{\rho}\sqrt{H}}\left( \Omega + 2\sqrt{2\ln\left(\dfrac{1}{\delta}\right)}\right).
    \]

    \begin{theorem}\label{thm:self-guided-VFA-rate}
        Suppose $\rho(\theta) \geq \underline{\rho}$ for all $\theta \in \Theta$. Given $\delta\in(0,1]$ and $N\ge1$, for any $H\ge 1$, there exists a function in $\mathcal{W}(\Phi_{N} \cup \Phi_H)$ with associated vector $\coef \in \R^{N+ H +1}$ such that with a probability of at least $1-\delta$, this vector (i) is feasible to constraints \eqref{FALPConst1}, (ii) is a ${E\programIndex{N,H}}$-feasible solution to constraints \eqref{FALPConst2}, and (iii) satisfies $\tallNorm{V^* - V(\coef)}_{\infty} \ \leq \ E\programIndex{N,H}$.
    \end{theorem}

Theorem \ref{thm:self-guided-VFA-rate} establishes that with a high level of probability the set $\mathcal{W}(\Phi_{N} \cup \Phi_H)$ will contain a VFA $V(\coef)$ satisfying the FALP constraints that simultaneously approaches $V^*$ at all states (i.e., $\infty$-norm) and becomes more feasible to the guiding constraints at the dimension-free rate of $1/\sqrt{H}$. This rate is analogous to the one associated with FALP in Theorem \ref{prop:ALP}, and so is the structure of the error bound. There are two distinct features of the bound in Theorem \ref{thm:self-guided-VFA-rate} worth noting. First, it contains the norm $\tallNorm{\optVprojPerpBar/\rho}_{2,\rho}$ in lieu of $\tallNorm{\coefFELPVecBar/\rho}_{2,\rho}$. It is easy to verify that $\tallNorm{\optVprojPerpBar/\rho}_{2,\rho} < \tallNorm{\coefFELPVecBar/\rho}_{2,\rho}$ if the projection of $V^*$ onto $\mathcal{W}(\Phi_{N})$ is nonzero. The difference between these norms signals the quality of the most recently computed VFA $V(\coefSGVecK{N})$.  This suggests that the number of additional samples $H$ needed to obtain a good approximation of $V^*$ decreases with $\tallNorm{\optVprojPerpBar/\rho}_{2,\rho}$, that is, when $V(\coefSGVecK{N})$ is itself closer to $V^*$. Second, Theorem \ref{thm:self-guided-VFA-rate} provides a rate at which the infeasibility of the guiding constraints decreases with $H$. This is because of the worst-case nature of the analysis. Specifically, it is possible for the reference VFA in the guiding constraints to be very close to $V^*$, in which case, satisfying the guiding constraints determines the worst-case distance of the new VFA from $V^*$. In practice, this is unlikely to happen, and the key insight from Theorem \ref{thm:self-guided-VFA-rate} is that good $\infty$-norm solutions will likely be a part of the \FGLPprog{N} feasible set as $H$ increases, which is corroborated by our numerical experience.

\subsection{Implementation Guidelines}  \label{sec:implementation-guidelines-gided-FALP}
We discuss the implementation guidelines for algorithms \ref{alg:state-rel-guide-FALP} and \ref{alg:self-guide-FALP}, focusing on parameter choices and solution issues that were not already discussed in \S\ref{sec:FALP-Implementation}. 

The additional parameters needed for the implementation of policy-guided FALP are the numbers of the basis functions and iterations $N$ and $Q$, respectively. As $N$ becomes larger, the time for a single iteration of Algorithm \ref{alg:state-rel-guide-FALP} increases, which includes solving \FALPprog{N} and simulating the greedy policy. This is because \FALPprog{N} will have more variables, so we need to evaluate expectations of a larger number of random basis functions during policy simulation. A sequential strategy is to first select $N$ such that the per iteration cost allows for choosing $Q$ such that a few iterations can be performed within an acceptable time limit. 

For self-guided FALP, we need to choose the batch size $B$ and the number of iterations $Q$. These choices become easier if we fix a target number of basis functions $N = QB$ following the logic discussed for FALP in \S\ref{sec:FALP-Implementation}. Then, smaller values of $B$ entail solving linear programs with fewer decision variables and doing so more often. In other words, the per iteration cost is lower with smaller $B$, but more iterations are needed and the improvement between iterations will likely be smaller. Therefore, the value of $B$ can be selected to balance improvement in the self-guided FALP objective function value and the per-iteration cost. Solving self-guided FALP requires handling the guiding constraints \eqref{FALPConst2}. We suggest replacing these constraints with a sampled subset, as done for FALP in \S\ref{sec:FALP-Implementation}. Under such replacement, analogues of Proposition \ref{prop:SG-ALP-basic} and the discussion following it hold over the sampled states. Similar to the implementation of FALP in \S\ref{sec:FALP-Implementation}, if we also replace constraints \eqref{FALPConst1} with $K$ sampled constraints, we expect the results in Theorem \ref{thm:self-guided-VFA-rate} to hold approximately as $K$ is sufficiently large.\looseness=-1


\looseness=-1

 Although we consider an iteration limit as the stopping criterion in algorithms \ref{alg:state-rel-guide-FALP} and \ref{alg:self-guide-FALP}, several alternatives are possible. For instance, the iteration limit can be replaced by a time limit, or both types of limits can be imposed together. Another strategy is to look at the improvement of consecutive policies and stop when these improvements are smaller than a certain threshold. If a lower bound on the optimal policy cost is available, these improvements can be converted to optimality gaps, and a termination gap can be set.

	\section{Extensions}\label{sec:extensions}
	
%
    Although we have assumed continuous state spaces and value functions thus far, the random basis function sampling approach underpinning our models can be readily extend to handle discounted cost MDPs with finite state spaces. A special structure that arises in important applications is a state space with a low dimensional discrete component and a high dimensional continuous component (e.g., financial and real options pricing). In this case, it is common to define a separate continuous VFA for each discrete state value, and our results directly apply. Next, we handle the more general case when such a strategy may not be computationally feasible. \looseness=-1

Consider the analgoue of the MDP in \S\ref{section:Background} with a discrete state space $\sSpace\coloneqq\{s^m\in\R^d:m\in\mathcal{M}\}$, where $\mathcal{M}$ is a finite index set and each state $s^m$ is a bounded real value. We denote by $V^*$ the MDP value function. Proposition \ref{prop:disc-MDP-error-rate} provides a bound on the $\infty$-norm error between the \FALPprog{N} VFA and $V^*$, which decreases at a rate of $1/\sqrt{N}$ as more random basis functions are sampled. Such a bound is possible because we can construct a continuous extension of $V^*$, as discussed next. Let $\sSpace^\mathrm{C}$ be the smallest continuous and compact set containing $\sSpace$. It is easy to verify that the following continuous function defined for each $s \in \sSpace^\mathrm{C}$ coincides with $V^*$ at all the discrete states: 
 \[
        V^\mathrm{C}(s)\coloneqq \sum_{m\in\mathcal{M}} V^*(s^m) \max\left\{0,1-\frac{\sNorm{s - s^m}_2}{\underline{s}} \right\},
    \]
    where $\underline{s}\coloneqq \min\big\{\sNorm{s^m - s^{m'}}_2: s^m,s^{m'}\in\sSpace, s^m\ne s^{m'}\big\}$ is a positive constant. We assume $V^\mathrm{C} \in \randBasisSet$, in which case, we have $V^\mathrm{C}(\cdot)=\coefVC+\int_\Theta \coefVCVecBar(\theta)\varphi(\cdotp;\theta)\diff \theta$ for some $\coefVCVec\coloneqq(\coefVC,\coefVCVecBar)$ (the results extend to the case when $V^\mathrm{C} \not\in \randBasisSet$, as explained in \S\ref{sec:Exact Linear Programs} and in \ref{EC:opt-V-not-in-R}). Compared with Theorem \ref{prop:ALP} in the continuous state space case, the weighting function $\coefFELPVecBar$ is replaced by $\coefVCVecBar$, and the constant $\contractionFactor$ is instead
    \[ \contractionFactor^\mathrm{C}  \coloneqq 5(\diamSState^\mathrm{C}+1) \lipConst  \sqrt{\mathbb{E}_\rho \big[\sNorm{\theta}_2^2\big]},\]
    where $\diamSState^\mathrm{C}\coloneqq\max_{s\in\sSpace^\mathrm{C}}\sNorm{s}_2$. Here, we will continue to use the notation related to \FALPprog{N} from \S\ref{sec:Model and Theory} and define $\tallNorm{V^* - \VALP}_{1,\nu}$ to denote the $(1,\nu)$-norm distance over the discrete state space, which is $\tallNorm{V^* - \VALP}_{1,\nu} = \sum_{m \in \mathcal{M}} \nu(s^m)|V^*(s^m) - V(s^m; \coefALPVec)|$. 
%
    
\begin{proposition}\label{prop:disc-MDP-error-rate}
    	Suppose Assumption \ref{asm:random basis function} with $V^*$ replaced by $V^\mathrm{C}$ and Assumption \ref{finitenessOfFALPOpt} hold, and in addition, $\rho(\theta) \geq \underline{\rho}$ for all $\theta \in \Theta$. Given $\delta\in(0,1]$, we have that any finite \FALPprog{N} optimal solution $\coefALPVec$ satisfies 
    	\[\tallNorm{V^* - V(\coefALPVec)}_{1,\nu} \le \dfrac{2\tallNorm{\coefVCVecBar/\rho}_{2,\rho}}{(1-\gamma)\underline{\rho}\sqrt{N}}\left( \Omega^C + \sqrt{2\ln\left(\dfrac{1}{\delta}\right)}\right)
,\]
    	with a probability of at least $1-\delta$.
    \end{proposition}
When the action space is finite for all states, we can drop Assumption \ref{finitenessOfFALPOpt}  and establish the existence of a finite optimal solution, although as discussed in \S\ref{sec:Model and Theory}, this assumption is already mild. We highlight that the construction of \FALPprog{N} does not change based on the structure of the state space since the sampling distribution $\rho(\cdot)$ does not depend on this structure. Therefore, the same procedures for generating basis functions apply in the discrete state space case. Using the arguments here, we can also handle state spaces with a mixture of discrete and continuous elements.

   Our results also extend to handle MDPs with a finite horizon $T < \infty$ by considering time to be in the state; that is, we can define the state as $(t,s)$. Because the options pricing application in \S\ref{sec:Option-Pricing} gives rise to a finite-horizon MDP, we formulate \FALPprog{N} next in the more familiar notation of such MDPs. Let the index set of stages in the horizon be $\mathcal{T} = \{0,1,\ldots,T\}$. The MDP value function at stage $t \in \mathcal{T} \setminus \{T\}$ is $V^*_t$, and we assume without a loss of generality that $V^*_T \equiv 0$. At stage $t \in \mathcal{T}$, the state space is $\mathcal{S}_t$, and the action space at this stage and state $s \in \mathcal{S}_t$ is $\mathcal{A}_t(s)$. Then, the finite horizon analogue of  \FALPprog{N} computes VFAs that approximate $V^*_t$ at each stage by sampling $\{\theta^1,\theta^2,\dots,\theta^N\}$: 
    \[ V^*_t \approx V(\beta_t) = \beta_{t,0} + \sum_{i=1}^{N} \beta_{t,i} \varphi(\cdot;\theta^i),\]
    where $\beta_t\coloneqq(\beta_{t,0},\beta_{t,1},\dots,\beta_{t,N})$ are the stage $t$ VFA weights. Because the sampling distribution $\rho(\cdot)$ does not depend on the stages or state space, the set of random basis functions can be the same across stages, which also provides the flexibility to use the same basis function weights across stages if needed. Assuming that the state-relevance distribution $\nu$ is defined over the stage $0$ state space $\mathcal{S}_0$ (it could easily be defined over the state spaces at all stages), \FALPprog{N} in the finite horizon setting is
    \resizebox{\linewidth}{!}{\begin{minipage}{\linewidth} \begin{align*}
    \max_{\coef}   \ \ &   \beta_{0,0} + \sum_{i=1}^{N}\beta_{0,i}\expt_{\nu} \big[\varphi(s;\theta^i) \big] && &&& \nonumber\\
    \text{s.t.}    \ \ &   (\beta_{t,0}-\gamma\beta_{t+1,0}) + \sum_{i=1}^{N} \left(\beta_{t,i} \varphi(s;\theta^i)  - \gamma \beta_{t+1,i} \expt_t \big[\varphi(s';\theta^i) \  | \ s,a\big]\right)   &&\le \ c_t(s,a), &&& (t,s,a) \in \mathcal{T}\setminus \{T\} \times \mathcal{S}_t \times \mathcal{A}_t(s),\\
    \end{align*}\end{minipage}}
where $c_t(s,a)$ and $\mathbb{E}_t$, respectively, are the stage $t$ cost function and expectation under the state transition function from stage $t$ to $t+1$. We omit the terminal condition for brevity. Theoretical guarantees that are analogous to the infinite horizon case for FALP and self-guided FALP can be derived in the finite horizon setting as well.

	\section{Perishable Inventory Control}
	\label{sec:Perishable Inventory Control}
	We perform a numerical study on the perishable inventory control problem considered in \citet[henceforth abbreviated LNS]{lin2017ContViolLearning}. We discuss the infinite-horizon discounted cost MDP formulation of the problem and instances in \S\ref{sec:PIC-MDP}, the experimental setup in \S\ref{subsec:PICcompSetup}, and numerical findings in \S\ref{subsec:PICResults}. \looseness=-1
	
	\subsection{MDP Formulation and Instances}\label{sec:PIC-MDP}
	Managing the inventory of a perishable commodity is a fundamental and challenging problem in operations management (\citealp{karaesmen2011PerishInv}, \citealp{chen2014pricingStrategiesForPerishableProducts}, \citealp{sun2014quadratic}, and LNS). We study a variant of this problem with partial backlogging and lead time from \S7.3 in LNS. 
	
	Consider a perishable commodity with
	$l\ge0$ and $J\ge0$ periods of life time and ordering lead time, respectively. Ordering decisions are made over an infinite planning horizon. At each decision epoch, the state vector is
	$
	s = (\onHandState{0},\onHandState{1},\dots,\onHandState{l-1},\pipeState{1},\pipeState{2},\dots,\pipeState{J-1})
	$ of size $l + J - 1$. The state element $\pipeState{i}$ for $i = 1,2,\dots,J-1$ is the previously ordered quantity that will be received $i$ periods from now.  If $s_0 \geq 0$, $\onHandState{i}$  for $i = 0,1,\dots,l-1$ is the amount of available commodity with $i$ periods of life remaining. If $s_0 < 0$, the values of these state elements are notional quantities to compute the total on hand inventory, which is $\onHandState{0} + \sum_{i = 1}^{l-1}\onHandState{i}$. Inventories $\onHandState{i}$ and  $\pipeState{j}$  take values in the interval $[0,\bar{a}]$ for all $i=1,\dots,l-1$ and $j=1,2,\dots,J-1$, respectively, where $\bar{a}\ge 0$ denotes the maximum ordering level. If $\onHandState{0} \in [-\sum_{i = 1}^{l-1}\onHandState{i}, \bar{a}]$, then the onhand inventory is non-negative. Instead, if $\onHandState{0} < -\sum_{i = 1}^{l-1}\onHandState{i}$, then the on-hand inventory  $\onHandState{0} + \sum_{i = 1}^{l-1}\onHandState{i}$ is negative and represents the amount of backlogged orders.


	The demand for the commodity is governed by a random variable. In each period, we assume that the demand is realized before order arrival and is satisfied in a first in-first out manner. Given a demand realization $D$, taking an ordering decision (i.e., action) $a$ from a state $s$ results in the system transitioning to a new state \looseness=-1
	\[
	s^\prime \coloneqq 
	\bigg(\max\bigg\{\onHandState{1} - (D-\onHandState{0})_+ ,\ \underline{s}-\sum_{i=2}^{l-1}\onHandState{i}\bigg\}, \onHandState{2},\dots,\onHandState{l-1},\pipeState{1}\pipeState{2},\dots, \pipeState{J-1},a \bigg),
	\]
	where $(\cdot)_+ := \max\{\cdot,0\}$ and $\underline{s}\le 0$ is a maximum limit on the amount of backlogged orders, beyond which we treat unsatisfied orders as lost sales. The updating logic in the first element of $s^\prime$ ensures that the backlogging limit is enforced. This can be understood as follows: If there was no backlogging limit, then the on-hand inventory after demand realization and before order arrival would be $\onHandState{1} - (D-\onHandState{0})_+ + \sum_{i=2}^{l-1}\onHandState{i}$; instead, in the presence of the maximum backlog limit $\underline{s}$, this total on-hand inventory of $\onHandState{1} - (D-\onHandState{0})_+ + \sum_{i=2}^{l-1}\onHandState{i}$ is greater than or equal to $\underline{s}$ if and only if $\onHandState{1} - (D-\onHandState{0})_+ \geq \underline{s} -\sum_{i=2}^{l-1}\onHandState{i}$. The remaining elements of $s^\prime$ are shifted elements of $s$, with the last element accounting for the latest order $a$.   \looseness=-1

	The immediate cost associated with a transition from a state-action pair $(s,a)$ is 
	\begin{equation*}\label{eq:PIC-cost-func}
		\resizebox{\hsize}{!}{%
			$
			c(s,a) \coloneqq 
			\gamma^J c_o a + \expt_{D}\Bigg[c_h\bigg[ \sum\limits_{i=1}^{l-1}\onHandState{i} - (D-\onHandState{0})_+\bigg]_+ 
			+ c_d(\onHandState{0}-D)_+ + c_b\bigg[D- \sum\limits_{i=0}^{l-1} \onHandState{i}\bigg]_+ +
			c_l\bigg[\underline{s} +D -\sum\limits_{i=0}^{l-1} \onHandState{i}\bigg]_+\Bigg],
			$    
		}
	\end{equation*}
	where expectation $\expt_{D}$ is given with respect to the demand distribution. The per-unit ordering cost $c_o\ge 0$ is discounted by $\gamma^J$ because we assume payments for orders are made only upon receipt. The holding cost $c_h\ge0$ penalizes leftover inventory $\big( \sum_{i=1}^{l-1}\onHandState{i} - (D-\onHandState{0})_+\big)_+$, while the per-unit disposal and backlogging costs $c_d\ge0$ and $c_b\ge0$ factor in, respectively, the costs associated with disposing $(\onHandState{0}-D)_+$ units and backlogging $\big(D- \sum_{i=0}^{l-1} \onHandState{i}\big)_+$ units. Finally, each unit of lost sales $\big(\underline{s} +D -\sum_{i=0}^{l-1} \onHandState{i}\big)_+$ is charged $c_l\ge0$. \looseness=-1
 
        We consider 24 perishable inventory control instances -- twelve from LNS with $l=J=2$ (three-dimensional state space) and twelve new higher-dimensional instances. Six of the new instances have $l=2$ and $J=4$ (five-dimensional state space), and the remaining six instances have $l=5$ and $J=6$ (ten-dimensional state space). Similar to LNS, across all instances, we fix the demand distribution to a truncated normal distribution with a mean of $5$ and support in the range $[0,10]$. We require the maximum limit on the amount of backlogged orders to equal the maximum ordering level, that is, $\underline s = -\bar{a}$. We vary the cost function parameters, the discount factor $\gamma$, the maximum ordering level $\bar{a}$, and the demand standard deviation $\sigma$. Their specific values are shown in tables \ref{table:FALP_vs_LNS}--\ref{table:PIC-10D}. 
    
    \subsection{Computational Setup}\label{subsec:PICcompSetup}
    We formulate \FALPprog{N} using the guidelines in \S\ref{sec:FALP-Implementation}. We use Fourier basis functions, with its bandwidth parameter $\varrho$ chosen via cross validation over the candidate set $\{10^{i}: i=5,4,\dots,-5\}$. For $\nu$, we considered both the initial MDP state of  $s_0=(5,5,5)$ (i.e., a degenerate initial distribution $\chi$) and a uniform distribution over the hyper-cube $[\underline s,\bar{a}]\times[0,\bar{a}]^{d-1}$. The latter choice leads to substantially better policies, so we report the results only for this choice. We use constraint sampling to solve \FALPprog{N} and choose $K=200000$ state-action pairs sampled from a uniform distribution over the hyper-cube $[\underline s,\bar{a}]\times[0,\bar{a}]^{d}$. The number of basis functions $N$ was set to $150$, $300$, and $600$ for the three-, five-, and ten-dimensional instances, respectively. We approximate expectations in \FALPprog{N} using sample average approximations constructed using $2000$ iid samples. 
    
    We formulate policy-guided FALP and self-guided FALP using the guidelines in \S\ref{sec:implementation-guidelines-gided-FALP}. For the former model, we set $Q$ equals 5, and for the latter one, we set $B$ equal to $25$, $50$, and $100$ on the three-, five-, and ten-dimensional instances, respectively. We keep the samples of the basis functions and constraints in these models the same as \FALPprog{N} for each instance. In particular, the guiding constraint in self-guided FALP are added for the states that appear in the state-action pair samples used to construct \FALPprog{N}. We use the notation \PGFALPprog{N,Q} and \SGFALPprog{N, B} when reporting the results for policy-guided FALP and self-guided FALP, respectively.  
    
    We use the Gurobi commercial solver to solve linear programs. We simulate greedy policies using $500$ sample paths to estimate their value. Similar to LNS, we replace the action space $[0,\bar{a}]$ by $\bar{a}$ equally spaced points and find the best action by using enumeration. We estimate the lower bounds using an approximate version of the \constrViolLearn approach that only performs dual updates involving Markov Chain Monte Carlo samples, for which we use the Metropolis–Hastings algorithm with $500$ independent Markov chain trajectories with a length of $1000$ and a burn-in size of $200$. Please see \ref{ec:sec:A Lower Bound Estimator for Constraint-sampled ALPs} for details. The maximum standard errors of all estimates were less than $1.3\%$.  In addition, to understand the variation caused by solving sampled models, for each instance and method, we repeat the solution of the models and simulations to estimate bounds ten times, that is, we perform ten trials.\looseness=-1

 \begin{table}[h]
		\centering
		\setlength{\tabcolsep}{20pt}
		\renewcommand{\arraystretch}{1.05}
		\caption{Comparison of FALP and LNS optimality gap percentages computed using the FALP lower bound on the three-dimensional perishable inventory control instances ($\sigma = 2$ and $c_l = 100$).}
		\adjustbox{width=\textwidth}{ 
			\begin{tabular}{rrrrrccr@{\hspace{1.5em}}cc}
				\hline\\[-14pt]
				& & & & & 
				\multicolumn{2}{c}{$\gamma=.95$} & & \multicolumn{2}{c}{$\gamma=.99$}\\
				\cline{6-7} \cline{9-10} \\[-12pt]
				\multirow{1}{*}{$c_h$} & \multirow{1}{*}{$c_d$} & \multirow{1}{*}{$c_b$} & \multirow{1}{*}{$\bar a$} & &
				\multicolumn{1}{r}{ LNS} &
				\multicolumn{1}{r}{{\FALPprog{150}}} & &
				\multicolumn{1}{r}{ LNS} &
				\multicolumn{1}{r}{{\FALPprog{150}}} \\
				\cline{6-7} \cline{9-10} \\[-8pt]
				2 &       5 &      10 &         10 &  &   0.0       & 0.1       &   & 2.3       & 0.0   \\
				\vspace{2pt}
				2 &       5 &      10 &         50 &  &   6.5       & 2.4       &   & 1.2       & 2.2   \\ 
				5 &      10 &      8  &         10 &  &   3.6       & 0.0       &   & 3.7       & 0.0   \\
				\vspace{2pt}
				5 &      10 &      8  &         50 &  &   13.0      & 2.2       &   & 12.7      & 3.3   \\
				2 &      10 &      10 &         10 &  &   0.8       & 0.0       &   & 1.8       & 0.1   \\
				2 &      10 &      10 &         30 &  &   5.1       & 1.6       &   & 2.3       & 2.1   \\
				\hline 
				\multicolumn{4}{r}{Average}        &  &   4.8       & 1.0       &   & 4.0       & 1.2   \\
				\hline 
		\end{tabular}}
		\label{table:FALP_vs_LNS}
	\end{table}

	\subsection{Results}\label{subsec:PICResults}

	Table \ref{table:FALP_vs_LNS} reports the optimality gaps of \FALPprog{150} and the LNS method, which solves an ALP model that embeds a fix set of 19 application-specific basis functions that involve hinges (i.e., $(\cdot)_+$) to mirror the structure of the cost function presented in \S\ref{sec:Perishable Inventory Control}. Optimality gaps are computed using the \FALPprog{150} lower bound. For each instance, the reported values are averaged over the 10 trials. For a discount factor of $0.95$, the LNS and \FALPprog{150} optimality gap ranges are $0\%$--$13\%$ and $0\%$--$2.4\%$, respectively. The behavior is similar for the larger discount factor of $0.99$. The variation of the \FALPprog{150} optimality gaps across 10 trials is less than 1\% across instances; that is, the \FALPprog{150} policy cost is robust to the resampling of random bases and constraints. The significant improvement of \FALPprog{150} over the LNS method highlights the value of using sampled random basis functions, which are not designed based on application structure. The policy and lower bound of the former model are near-optimal and close the gap on almost all the three-dimensional instances. In addition, these results show that more advanced iterative methods to guide state-relevance choice, such as policy-guided FALP and self-guided FALP, are not needed on these instances. In fact, we confirmed that the performance of the policy-guided FALP and self-guided FALP policies are comparable to the ones from FALP on these instances.\looseness=-1

	\begin{table}[h!]
	\centering
	\setlength{\tabcolsep}{27pt}
	\renewcommand{\arraystretch}{1.1}
	\caption{Comparison of FALP, policy-guided FALP, and self-guided FALP optimality gap percentages computed using the FALP lower bound on the five-dimensional perishable inventory control instances ($\gamma = 0.95$ and $c_l = 1000$).}
	\begin{large}
		\adjustbox{width=\textwidth}{
			\begin{tabular}{ccccc@{\hspace{6em}}rrrr}
				\hline\\[-15pt]
				\multicolumn{1}{c}{$c_h$}       &       \multicolumn{1}{c}{$c_d$}       & 
				\multicolumn{1}{c}{$c_b$}       &       \multicolumn{1}{c}{$\sigma$}    & &
				\multicolumn{1}{c}{\begin{tabular}[r]{@{}c@{}} $\texttt{FALP}_{\scaleto{\mathrm{300}}{6pt}}$ \end{tabular}} & 
				\multicolumn{1}{c}{\begin{tabular}[r]{@{}c@{}} $\texttt{FALP}^{\raisemath{2pt}{\mathrm{PG}}}_{\scaleto{\mathrm{300,5}}{7pt}}$ \end{tabular}}    & 
				\multicolumn{1}{c}{\begin{tabular}[r]{@{}c@{}} $\texttt{FALP}^{\raisemath{2pt}{\mathrm{SG}}}_{\scaleto{\mathrm{300,50}}{7pt}}$ \end{tabular}}   \\ 
				\cline{6-8} \\[-12pt]
				 1 &     8 &      2 &    5 &   &  16.3              & 9.9     & 8.7     \\ 
				 1 &     8 &      2 &    2 &   &  18.7              & 12.4    & 9.1     \\
				 1 &     2 &      8 &    5 &   &  13.6              & 21.3    & 7.3     \\   
				 1 &     2 &      8 &    2 &   &  10.6              & 6.2     & 4.2     \\   
				 2 &     8 &      5 &    5 &   &  13.2              & 7.6     & 9.6     \\   
				 2 &     8 &      5 &    2 &   &  13.6    	        & 10.0    & 7.1     \\   
				\hline
				\multicolumn{4}{r}{Average}&   & 14.3            & 11.2    & 7.6     \\
				\hline
			\end{tabular}
	}\end{large}
	\label{table:PIC-5D}
\end{table}

    Table \ref{table:PIC-5D} compares the (average) optimality gaps of the \FALPprog{300}, \PGFALPprog{300,5}, and \SGFALPprog{300, 50} policies on the five-dimensional instances. The optimality-gap ranges for these respective policies are 10.6\%--18.7\%, 6.2\%--21.3\%, and 4.2\%--9.6\%. Across the 10 trials in each instance, the optimality gaps vary by at most 1\% for \FALPprog{300} and \SGFALPprog{300, 50} and by at most 10\% for \PGFALPprog{300,5}. Policy-guided FALP exhibits somewhat unstable behavior as witnessed by the larger optimality gap range and performance variance across trials. In contrast to the three-dimensional instances, we see here that \FALPprog{300} without state-relevance distribution (guiding) updates, can lead to highly suboptimal policies. Despite the weak policies, the \FALPprog{300} lower bound used to compute the optimality gap is excellent as witnessed by the \PGFALPprog{300,5} results. This observation is consistent with the discussion in \S\ref{sec:FALP-Implementation} and \S\ref{sec:Self-guided Approximate Linear Programs}, suggesting that ALP VFAs providing good lower bounds may not provide good policies because of a poor state-relevance distribution choice. The policy-guided FALP policy improves on the FALP policy on almost all instances, except for one instance, where it has a substantially worse optimality gap. Self-guided FALP leads to consistent and large improvements over FALP across all the instances, as well as a significant benefit over policy-guided FALP, again with the exception of one instance. This underscores the value of the guiding mechanism underpinning self-guided FALP.

   %
    

      	\begin{table}[h!]
	\centering
	\setlength{\tabcolsep}{27pt}
	\renewcommand{\arraystretch}{1.1}
	\caption{Comparison of FALP and self-guided FALP optimality gap percentages computed using the FALP lower bound on the ten-dimensional perishable inventory control instances ($\gamma = 0.95$ and $c_l = 1000$).}
	\begin{large}
		\adjustbox{width=\textwidth}{
			\begin{tabular}{ccccc@{\hspace{6em}}rrr}
				\hline\\[-15pt]
				\multicolumn{1}{c}{$c_h$}       &       \multicolumn{1}{c}{$c_d$}       & 
				\multicolumn{1}{c}{$c_b$}       &       \multicolumn{1}{c}{$\sigma$}    &  &
				\multicolumn{1}{r}{\begin{tabular}[r]{@{}c@{}}  $\texttt{FALP}_{\scaleto{\mathrm{600}}{6pt}}$  \end{tabular}} & 
				\multicolumn{1}{r}{\begin{tabular}[r]{@{}c@{}}   $\texttt{FALP}_{\scaleto{\mathrm{1000}}{6pt}}$ \end{tabular}} & 
				\multicolumn{1}{r}{\begin{tabular}[r]{@{}c@{}}   $\texttt{FALP}^{\raisemath{2pt}{\mathrm{SG}}}_{\scaleto{\mathrm{600,100}}{7pt}}$ \end{tabular}} \\
				\cline{6-8} \\[-15pt]
				 1 &     8 &      2 &    5 &  &    11.9       &  26.4        &  7.3      \\
				 1 &     8 &      2 &    2 &  &    5.9        &  10.4        &  4.7      \\
				 1 &     2 &      8 &    5 &  &    10.9       &  23.8        &  7.4      \\
				 1 &     2 &      8 &    2 &  &    7.4        &  10.6        &  5.9      \\
				 2 &     8 &      5 &    5 &  &    12.4       &  27.7        &  7.3      \\
				 2 &     8 &      5 &    2 &  &    9.2        &  13.2        &  7.2      \\   
				\hline
				\multicolumn{4}{r}{Average}&  &     9.6             &      18.6          &  6.6       \\
                \hline
			\end{tabular}
	}\end{large}
	\label{table:PIC-10D}
    \end{table}

        \begin{figure}[h!]
		\centering
		\caption{Variation of \SGFALPprog{600, 100} optimality gaps on two representative ten-dimensional perishable inventory control  instances with $(c_h, c_d, c_b, \sigma)$ equal to $(1,2,8,2)$ and $(1,2,8,5)$ in the left and right panels, respectively.}
		\includegraphics[width=\linewidth]{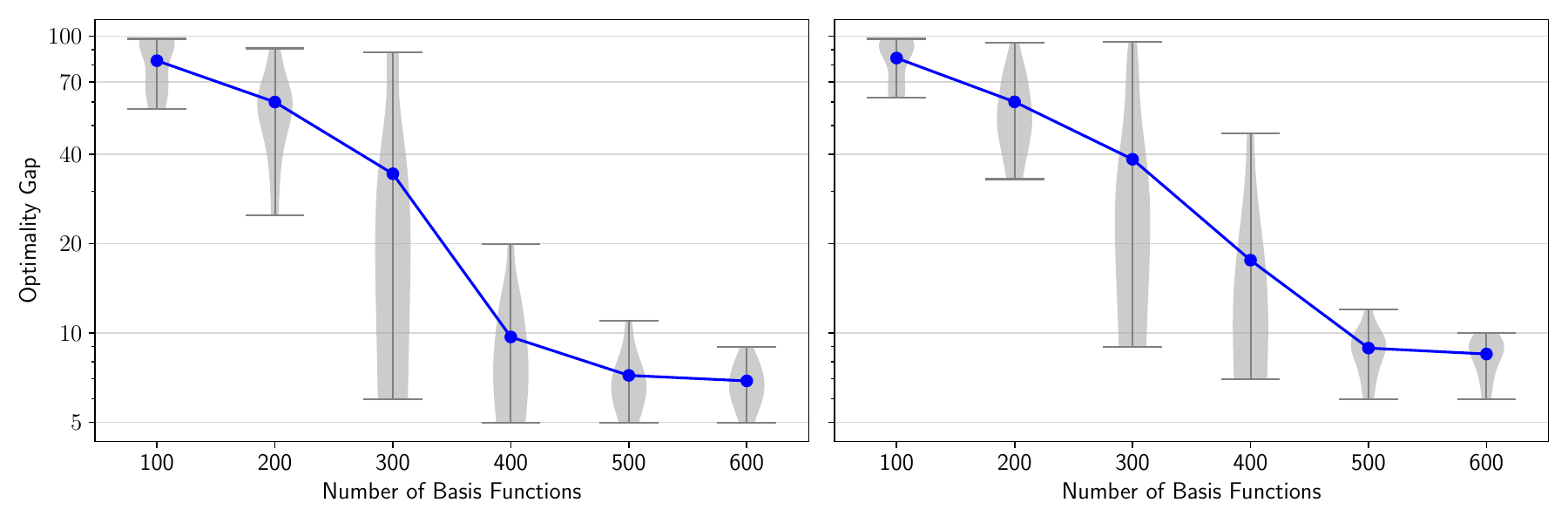}
		\label{fig:FGLP_convergence_PIC}
	\end{figure}
    
    Table \ref{table:PIC-5D} reports the (average) optimality gaps of the \FALPprog{600} and \SGFALPprog{600, 100} policies on the ten-dimensional instances computed using the \FALPprog{600} lower bound. We were unable to obtain reliable results with \PGFALPprog{600,5} because it results in very erratic behavior as we iterated the steps in Algorithm \ref{alg:state-rel-guide-FALP}. This observation is consistent with the behavior reported by \citet{Farias2006} on a Tetris application when using policy guided state-relevance distribution updates. In particular, they observe improvement in policy performance for the first few iterations and then an unexplainable dropoff. We also added \FALPprog{1000} as a benchmark to see if additional random basis functions in FALP help. The optimality gap ranges for \FALPprog{600}, \FALPprog{1000}, and \SGFALPprog{600, 100} are 5.9\%--12.4\%, 10.4\%--27.7\%, and 4.7\%--7.3\%, respectively. The optimality gap variation across trials is at most 2.3\%. Once again, policies based on adhoc state-relevance distribution choices can perform poorly, even if they have a sufficient number of basis functions, as shown by the substantially worse  performance of \FALPprog{1000} compared with \FALPprog{600}. The optimality gaps of \SGFALPprog{600, 100} show that the lower bounds from \FALPprog{600} are very good. The mechanism in self-guided FALP to guide the state-relevance distribution adds significant value relative to the benchmarks and leads to excellent policies. It improves on the  \FALPprog{600} policy by an average of 3\% and by as much as 5\%. To gain some insight into how the self-guiding mechanism in \SGFALPprog{600, 100} helps, we display the average optimality gap of the \SGFALPprog{600, 100} policy on two representative ten-dimensional instances as basis functions are iteratively added in batches of 100, as well as, a violin plot for each iteration showing the variation of the optimality gaps. At the beginning (100 basis functions), the policies are bad across all trials. After a few iterations (200 and 300 basis functions), the variation of optimality gaps increases, and we find very good policies in some trials and poor policies in others. In subsequent iterations, the variance of policy performance quickly decreases, and at 600 basis functions, the worst case optimality gap across trials is less than 10\%. Our results on the ten-dimensional instances when compared to those on the three- and five-dimensional instances suggest that the importance of the state-relevance distribution becomes more critical for higher dimensional problems. Intuitively, this would happen if near-optimal policies visit smaller and smaller regions of the state space as its dimension becomes larger, in which case, there is a greater need for having a good state-relevance distribution that aligns the ALP VFA error minimization objective with policy performance. The guiding mechanism in \SGFALPprog{600, 100} helps with such alignment.  
    
    Finally, as expected, the average run times increase as we employ more basis functions and move to higher dimensional instances. On the three-dimensional instances, these run times for \FALPprog{600} and the LNS method were six and three minutes, respectively. For instances with a five-dimensional state space, \FALPprog{300}, \PGFALPprog{300,5}, and \SGFALPprog{300, 50} take on average 12, 108, and 32 minutes, respectively. The average run times of \FALPprog{600}, \FALPprog{1000}, and \SGFALPprog{600, 100} were 23, 86, and 54 minutes, respectively, on the ten-dimensional instances. Thus, the computational times of self-guided FALP to obtain the policy improvements discussed earlier are encouraging.

\section{Bermudan Options Pricing}\label{sec:Option-Pricing}
    We perform a numerical study on the Bermudan option pricing problem in \citet[henceforth abbreviated DFM]{desai2012pathwise}. In \S\ref{subsec:Option-Pricing-MDP}, we present the finite-horizon discounted MDP formulation. In \S \ref{sec:Option-Pricing-Setup}, we describe our computational setup. In \S\ref{sec:Option-Pricing-results}, we discuss the results and findings.

    \subsection{MDP Formulation}\label{subsec:Option-Pricing-MDP}
    
    We consider the pricing of a knocked-out Bermudan call option with $J$ assets over a finite time horizon. The components of the finite-horizon MDP formulation will be presented following the notation in \S\ref{sec:extensions}, but with one exception: instead of a cost function $c_t(s_t,a_t)$, we use a reward function $r_t(s_t,a_t)$. The option has $T$ exercise opportunities over $Y$ years; that is, exercise is possible at times $\{\tau,2\tau,\dots,T\tau\}$, where $\tau\coloneqq Y/T$. The asset prices at stage $t \in \mathcal{T} = \{0,1,\dots,T\}$ are $p_t\coloneqq(p_{t,1},p_{t,2},\dots,p_{t,J})$, where $p_{t,j}$ is the price of the $j$-th asset at this time. Prices evolve according to a multi-asset geometric Brownian motion. The option becomes worthless any time the maximum of the  $J$ asset prices exceeds a pre-specified barrier price $p^{\mathrm{B}}$. We use the binary variable $y_t \in \{0,1\}$ to indicate if the option is knocked out at time $t$. It takes a value of one in this case and is zero otherwise. The transition equations governing $y_t$ are $y_0=\indicator\{\max_j p_{0,j}\ge p^\mathrm{B}\}$ and $y_t = \max\big\{y_{t-1}, \indicator\{\max_j p_{t,j}\ge p^\mathrm{B}\}\big\}$ for $t > 0$, where \indicator\{a\} equals one if $a$ is true and zero otherwise.
    At time $t$, the MDP state is given by the vector $s_t = (p_{t,1},p_{t,2},\dots,p_{t,J},y_t)$ that belongs to the state space $\sSpace = [0,p^\mathrm{B}]^J\times\{0,1\}$. The MDP action $a_t$ is binary, with values of one and zero corresponding to ``stop'' and ``continue,'' respectively. Stopping at stage $t$ yields the reward $r_t(s_t,0) = \gamma^tg(s_t)$, where the discount factor $\gamma= \exp(-r\tau)$, $r$ is the risk-free interest rate, and the payoff function $g(\cdot):\R^{J+1}\mapsto\R$ with respect to a pre-specified strike price $p^\mathrm{S}$ is 
    \[
        g(s_t)\coloneqq \max\left\{\left(\max_{j=1,2,\dots,J} p_{t,j} \ - \  p^\mathrm{S}\right),0 \right\} (1-y_t).
    \]
  A continue decision at state $s_t$ has zero reward, that is, $r_t(s_t,1) = 0$. The objective is to find an exercise policy that maximizes the discounted expected reward.
  
   Our experiments use nine instances from DFM, for which $Y$, $T$, $p^\mathrm{S}$, $p^\mathrm{B}$, and $r$ are 3, 54, 100, 170, and 5\%, respectively. The geometric Brownian motion driving the prices has zero correlation and volatilities equal to $20\%$. All assets share the same initial price $p^{\mathrm{I}}>0$, that is, $p_{0,1}=p_{0,2}=\dots=p_{0,J}=p^{\mathrm{I}}$. This price is varied between $90, 100$, and $110$, and the number of assets $J$ takes on the values $4, 8$, and $16$. Although the asset prices can take values greater than the barrier price $p^\mathrm{B}$, they need not be included in the state space because the option becomes worthless at all such prices. Thus, the range of each price relevant to the MDP belongs to the interval $[0,p^\mathrm{B}]$.

    \subsection{Computational Setup and Benchmarks} \label{sec:Option-Pricing-Setup}
    
    Our setup of models mirrors \S\ref{sec:Perishable Inventory Control}. We formulate the finite-horizon version of \FALPprog{N} given in \S\ref{sec:extensions} using $N=500$ Fourier random bases, with its bandwidth parameter $\varrho$ chosen via cross-validation over the candidate set $\{10^5,10^4,\dots,10^{-5}\}$. The strategy of using a policy to obtain a state-relevance distribution in \S\ref{sec:FALP-Implementation} is simplified because the exercise decisions do not affect prices. Therefore, the price-portion of the state evolves according to the geometric Brownian motion model, regardless of the policy used. Motivated by this property, we use a lognormal state-relevance distribution of prices. We find that \FALPprog{500} performs much better with this choice than a uniform distribution. We do not consider policy-guided FALP given its unstable behavior. For self-guided FALP, we choose $B$ equal to $100$; that is, we consider \SGFALPprog{500,100}. We sample the constraints of both \FALPprog{500} and \SGFALPprog{500,100} by generating $3000$ trajectories of prices from the geometric Brownian motion model. We approximate the expected values by sampling 500 transitions from this model. 
    
    We consider two application-specific benchmarks. The first is a least squares Monte Carlo (LSM), which is popular for financial and real option valuation (\citealt{ValuationCarriere1996,longstaff2001valuing,glasserman2004simulation}, and see \citealt{nadarajah2021real} for a recent review) and provides very good policies on the instances we consider. LSM approximates the optimal continuation function $C_t(s_t) \coloneqq \expt[V_{t+1}^*(p_{t+1})y_{t+1}   |  p_t]$ with the boundary condition $C_T(s_T) \equiv 0$ using a backward recursive scheme that uses a regression. To construct the continuation function approximation, we use the same application-specific $J+2$ basis functions considered in DFM, which are $\phi_1(s_t) = 1-y_t$, $\phi_2(s_t)= g(s_t)$, and $\phi_j(s_t) = (1-y_t)p_{t,j}$ for $j=1,2,\dots,J$. We use 100,000 sample paths to estimate the weights of these basis functions at each time $t$. Our second benchmark is an ALP with the same $J+2$ basis functions as LSM. We denote this model by \FALPBasesPO. We construct the constraints of this model using the same price trajectories and transitions used in the construction of \FALPprog{500}. \looseness=-1 

    We simulate 20,000 price trajectories to evaluate the value of greedy policies, which provide a lower bound on the optimal policy value (because we are maximizing reward). The maximum standard error of these estimates is 0.4\%. We benchmark the performance of all policies using the essentially optimal upper bound for these instances from the pathwise optimization approach of DFM. For each instance and method, we perform ten trials. 
    
    
\begin{table}[h!]
		\centering
		\caption{Comparison of optimality gaps on the Bermudan options pricing application.}
		\setlength{\tabcolsep}{25pt}
		\adjustbox{width=\textwidth}{
        \begin{tabular}{rrrcccc}
        \hline\\[-9pt]
        \multicolumn{1}{c}{$J$} & 
        \multicolumn{1}{c}{$p^{\mathrm{init}}$} & &
        \multicolumn{1}{c}{LSM} & 
        \multicolumn{1}{c}{\FALPBasesPO} & 
        \multicolumn{1}{c}{\FALPprog{500}} & 
        \multicolumn{1}{c}{\SGFALPprog{500, 100}} \\
        \cline{4-7}\\[-9pt]
        4 &   90    &  & 6.8     & 5.1         & 2.8             & 2.8         \\
        4 &  100    &  & 6.1     & 5.9         & 3.7             & 3.6         \\ 
            \vspace{6pt}
        4 &  110    &  & 5.0     & 6.2         & 5.7             & 3.5         \\
        8 &  90     &  & 5.9     & 5.5         & 3.4             & 3.3         \\  
        8 &  100    &  & 4.3     & 5.8         & 8.5             & 1.8         \\
            \vspace{6pt}     	
        8 &  110    &  & 3.0     & 5.5         & 16.0            & 1.4         \\  
       16 &  90     &  & 3.9     & 5.4         & 2.3             & 2.3         \\ 
       16 &  100    &  & 2.6     & 5.0         & 1.8             & 1.6         \\
       16 &  110    &  & 1.9     & 4.4         & 1.7             & 1.2         \\   
        \hline \\[-9pt]
        \multicolumn{2}{r}{Average}  
                    &  & 4.4	 & 5.4     	   & 5.1 	         & 2.4 \\
        \hline
		\end{tabular}
		}
		\label{table:LSM_FALP_FGLP_Opt_Price}
	\end{table}

    \subsection{Results}\label{sec:Option-Pricing-results}
        Table \ref{table:LSM_FALP_FGLP_Opt_Price} reports the optimality gaps of LSM, \FALPBasesPO, \FALPprog{500}, and \SGFALPprog{500,100} on the nine DFM instances averaged across 10 trials. The respective optimality gap ranges for each method are 1.9\%--6.8\%, 4.4\%--6.2\%, 1.7\%--16\%, and 1.4\%--3.6\%, which shows that the \SGFALPprog{500,100} policy is near optimal and improves on the remaining benchmarks. The performance of the \FALPprog{N} policy is within 1\% of the one from \SGFALPprog{500,100} on six of the nine instances but 2.2\%, 6.7\%, and 14.6\% worse on the remaining instances. Once again, we see significant value in updating the state-relevance distribution using the logic in \SGFALPprog{500,100}. There is no clear ordering between the policies of \FALPBasesPO \ and LSM -- the average optimality gap of the LSM method across all the instances is 1\% smaller than \FALPBasesPO.  The \SGFALPprog{500,100} policy is significantly better than the LSM policy, with improvements of less than 2\% on 5 instances and greater than 2\% on the remaining four. The largest such improvement is 4\%. \looseness=-1
        
        The policy improvements obtained using  \SGFALPprog{500, 100} over LSM are comparable to or larger than those reported in DFM with a pathwise optimization policy. The superior self-guided FALP policies come at a computational cost. The average runtime of LSM, \FALPBasesPO, \FALPprog{500} and \SGFALPprog{500,100} across trials and instances are, respectively, 2.42, 5.1, 99.5, and 117.9 minutes. There is thus an additional, albeit manageable, computational overhead to obtain the improved \SGFALPprog{500,100} policies.
        
        A broader takeaway from these experiments is that an application-agnostic ALP model with random basis functions and a guided state-relevance distribution can provide near-optimal policies for a challenging option pricing problem, also improving on application-specific benchmarks.

\section{Conclusions}\label{sec:Conclusions}

We revisit the approximate linear programming approach for computing value function approximations (VFAs) for Markov decision processes (MDPs). We focus on the key choices needed to formulate an approximate linear program (ALP) that affects the quality of the VFA and its associated policy. The first is the selection of the basis functions defining the ALP VFA, and the second is the choice of a state-relevance distribution in the ALP objective. These choices are typically made in an ad-hoc manner based on domain knowledge, which limits the applicability of ALP. We embed VFAs based on cheaply sampled random basis functions in ALP, hence sidestepping the need for ad-hoc basis function engineering. We refer to this model as feature-based ALP (FALP). We also propose an iterative scheme to guide the state-relevance distribution in FALP using its past VFA information, which leverages the ability to add new random basis functions in an inexpensive manner. We develop error bounds for the VFAs from these models and also show that self-guided FALP has desirable theoretical properties not shared by an existing iterative scheme for updating the state-relevance distribution. We test FALP and self-guided FALP on challenging perishable inventory control and option pricing applications. Self-guided FALP outperforms FALP and application-specific benchmarks. Our findings showcase the potential for our procedure to (i) significantly reduce the implementation burden of using ALP and (ii) provide an application-agnostic policy and lower bound for MDPs that can be used to benchmark other methods.

Our research suggests several interesting directions for future work, of which we state two. The first is to study the possibility and value of a guided sampling mechanism for ALP where the new samples of random basis functions leverage information from past VFAs. Approaches for the data-dependent sampling of random basis functions in machine learning (see, e.g, \citealp{sinha2016learning,shahrampour2018data}) can query the function being approximated, which is the unknown MDP value function in our setting. It is unclear how to develop inexpensive and approximate queries of the MDP value function that still provide useful information, which would be needed to obtain an effective and efficient sampling approach. The second is to investigate the value of random basis functions in other approximate dynamic programming methods and compare against neural networks and deep learning that also attempt to mitigate tuning but lead to nonlinearly parametrized VFAs, which are typically harder to train. \looseness=-1

%
%
%

	\bibliographystyle{informs2014}

\ECSwitch

\ECHead{Electronic Companion to Self-guided Approximate Linear Programs}

All proofs are in \S \ref{ec:sec:Proofs}. In \S\ref{sec:RelaxAssumps}, we discuss how assumptions used in \S\ref{sec:FELP} and \S\ref{sec:Model and Theory} can be relaxed. In \S\ref{ec:sec:Analyzing an FALP-based Sampling Bound for FGLP}, we discuss why applying the analysis of FALP does not provide an insightful bound for self-guided FALP. In \S\ref{ec:sec:A Lower Bound Estimator for Constraint-sampled ALPs}, we outline a heuristic version of the exact constraint violation learning approach in \cite{lin2017ContViolLearning}, which we use to obtain lower bounds in \S\ref{sec:Perishable Inventory Control}.

\section{Proofs}\label{ec:sec:Proofs}
We define a constant $\Gamma := (1+\gamma)/(1-\gamma)$ which we will use in various proofs. We also use the notation $\indicator\{a\}$ to show the indicator function that is 1 when $a$ is true and 0 otherwise.

\subsection{Additional Details of Assumption \ref{asm:MDP}}\label{subsec:SumAssump}
Assumptions~\ref{asm:MDP} and~\ref{asm:random basis function} will hold for all proofs in the electronic companions. In particular, Assumption~\ref{asm:MDP} ensures the existence of an an optimal policy solving program \eqref{eq:minCostMDP}. There are known conditions in the literature that guarantee such existence. We summarize some of these conditions below. 
\begin{assumption} \label{ec:asm:MDP-Kernel-cost}
	It holds that (i) The MDP cost function is bounded over $\saSpaceS$ and function $c(s,\cdot):\aSpaceS\mapsto\R$ is lower semicontinuous for all $s\in\sSpace$. (ii) For every bounded and measurable function $V:\sSpace\mapsto\R$, the mapping $(s,a) \mapsto \int_{\sSpace} V(s^\prime)P({\diff}s^\prime|s,a)$ is bounded and continuous over $\saSpaceS$.
	(iii) There exists a finite-cost policy $\pi$ such that $\polCost(s,\pi)<\infty$ for all $s\in\sSpace$.
\end{assumption}

Assumption \ref{ec:asm:MDP-Kernel-cost} is adopted from assumptions 4.2.1 and 4.2.2 in \citealt[henceforth abbreviated as \citetalias{hernandez1996discrete}]{hernandez1996discrete}. Specifically, in Part (a) of Assumption 4.2.1 in \citetalias{hernandez1996discrete}, the cost
function $c(s,\cdot)$ is assumed to be lower semi-continuous, non-negative, and inf-compact (defined in Condition 3.3.3 in \citetalias{hernandez1996discrete}) whereas, in our setting, non-negativity is replaced by boundedness and
the inf-compactness is guaranteed by the virtue of $c(s,\cdot)$ being lower semi-continuous and its domain $\aSpaceS$ being either a continuous compact real valued set or a finite set (please see Assumption \ref{asm:MDP}). Part (b) of Assumption 4.2.1 and Assumption 4.2.2 in \citetalias{hernandez1996discrete} are equivalent to parts (ii) and (iii) of Assumption \ref{ec:asm:MDP-Kernel-cost}, respectively. It is noteworthy that the condition specified in part (iii) of Assumption \ref{ec:asm:MDP-Kernel-cost} is the definition of the strong continuity of the MDP stochastic kernel $P$ (see Condition 3.3.3 in \citetalias{hernandez1996discrete}). Under the aforementioned technical conditions, Part (b) of Theorem 4.2.3 in \citetalias{hernandez1996discrete} guarantees the
existence of a deterministic and stationary policy  that is ``$\gamma$-discount optimal". In other words, $\pi^*\in\Pi$ solves \eqref{eq:minCostMDP} in our setting.

\subsection{Proofs of Statements in \S \ref{sec:Exact Linear Programs}}

\proof{\normalfont\textbf{Proof of Proposition \ref{prop:V-optimal-to-FELP}.}} Since $V^*\in\randBasisSet$, there exists $(\coefFELP, \coefFELPVecBar)$ such that $V^*(s) = \coefFELP +\int_\Theta \coefFELPVecBar(\theta)\varphi(s;\theta)\diff \theta$ for all $s\in\sSpace$. We show that $(\coefFELP, \coefFELPVecBar)$ is our desirable solution. This solution is feasible to FELP since $V^*$  satisfies the constraints in~\eqref{constr:ELP}. It is also optimal because $V^*$ satisfies the optimality equations $V^*(s) =  \min_{a\in\aSpaceS}\{c(s,a) + \gamma  \expt[V'(s^\prime) |  s,a] \}$ for every $s\in\sSpace$ which indicates that all the constraints of~\eqref{constr:ELP} hold as equality.
\hfill \Halmos

\subsection{Proofs of Statements in \S \ref{sec:Approximate Linear Programs with Random Basis Functions}}
To prove Theorem~\ref{prop:ALP}, we require the following lemmas and propositions. 
\begin{lemma}\label{ec:lem:optV-properties} 
	Any continuous function $V:\sSpace\mapsto \R$ that is feasible to constraints \eqref{constr:ELP} satisfies $V(s) \le V^*(s)$ for all $s\in\sSpace$.
\end{lemma}
\proof{Proof.}
The proof follows from Part (b) of Lemma 4.2.7 in \citetalias{hernandez1996discrete}, which requires four assumptions to hold. We now show that these assumptions are true in our setting. (i) Since $V$ is continuous, it is measurable; (ii) the Bellman operator $\mathrm{T}V(s) \coloneqq \min_{a\in\aSpaceS}\{c(s,a) + \gamma\expt[V(s^\prime)|s,a]\}$ is well defined for every continuous function $V$, i.e. the minimum over $\aSpaceS$ is attained since $\aSpaceS$ is either a real-valued continuous compact set or a finite set from Assumption \ref{asm:MDP}, $c(\cdot,\cdot)$ is bounded, and the expectation $\expt[V(s^\prime)|s,a]=\int_{\sSpace} V( s^\prime) P(\diff s^\prime | s,a)$ is finite by  Assumption \ref{ec:asm:MDP-Kernel-cost}; (iii) since $V$ is feasible to constraints \eqref{constr:ELP}, we have 
\[V(s) \ \le  \  \min_{a\in\aSpaceS}\left\{ c(s,a) + \gamma  \expt[V(s^\prime) \vert s,a]\right\} \ = \ \mathrm{T}V(s), \qquad \forall s\in \sSpace;\]
(iv) finally, the continuity of $V$ and the compactness of $\sSpace$ imply $\max_{s\in\sSpace}{|V(s)|} <\infty$ and thus \[
\lim_{n\rightarrow\infty}\gamma^n \expt\Bigg[\sum_{t=0}^{n} V(s^\pi_t)  \Big\vert s_0 = s\Bigg] \  \le  \ \max_{s\in\sSpace}{|V(s)|}  \lim_{n\rightarrow\infty}(n+1)\gamma^n  \ = \ 0, \qquad \forall s\in\sSpace, \pi \in\Pi,
\]
where expectation $\expt$ and the notation $s_t^\pi$ retain their definitions from  \S \ref{sec:Exact Linear Programs}. These indicate that the function $V$ fulfills the four assumptions of Part (b) of Lemma 4.2.7 in \citetalias{hernandez1996discrete} and hence $V(s)\le V^*(s)$ for all $s\in\sSpace$.
\hfill\Halmos
\endproof

\begin{proposition}\label{prop:rahimi-recht}
	Suppose $\rho(\theta) \ge \underline{\rho}$, for all $\theta\in\Theta$ and Assumption \ref{asm:random basis function} holds. Consider $\delta\in (0,1]$ and a function $V(s;\coefInf) = \beta_0 + \int_\Theta \coefInfBar(\theta)\varphi(s;\theta)\diff \theta$ with $\sNorm{\coefInfBar/\rho}_{2,\rho}<\infty$. Given $N$ iid samples $\{\theta^{i}: i=1,2,\dots,N\}$ from $\rho$, there exist finite coefficients $\bar\beta_i,i=0,1,2,\ldots,N,$ such that 
	\begin{equation}\label{eq:inftynormbound}
		\left\lVert V(\coefInf) \ - \ \left(\bar\beta_0 +\sum_{i=1}^N \bar\beta_i \varphi(\cdot;\theta^i) \right) \right\rVert_{\infty} \le 
		\frac{\tallNorm{\coefInfBar/\rho}_{2,\rho}}{\underline{\rho}\sqrt{N}}\left( \Omega + \sqrt{2\ln\left(\dfrac{1}{\delta}\right)}\right)
	\end{equation}
	with a probability of at least $1-\delta$. 
\end{proposition} 

\proof{Proof.} The proof of this proposition follows similar steps to the proof of Theorem 3.2 in \cite{rahimi2008uniform}. In particular, given a constant $r>0$ and $N$ iid samples $\vartheta\coloneqq(\theta^1,\theta^2,\dots,\theta^N)$, we first define random variable $\bar{V}_\vartheta(s)\coloneqq \beta_0 + \frac{1}{N}\sum_{i=1}^{N}V_{i,\vartheta}(s)$ where $V_{i,\vartheta}(s) \coloneqq \beta^r_i \varphi(s;\theta^i) $ and
$\beta^r_i \coloneqq \frac{1}{\rho(\theta^i)} \int_{\Theta} \coefInfBar(\theta)\indicator\left\{\theta: \sNorm{\theta - \theta^i}_2 \le r\right\} \diff\theta$. Let 
\[ g(\vartheta) \coloneqq \left\lVert V(\coefInf) - \bar{V}_\vartheta \right\rVert_\infty.\]
We provide an upper bound on $g(\vartheta)$ that is decreasing in $N$ and holds with high probability. To do so, we take the following steps: 

Step (i): We first prove
\begin{equation}\label{eq:boundonEg}\expt\left[g(\vartheta)\right]\leq L\left(1+\diamSState\right)\tallNorm{\coefInfBar/\rho}_{2,\rho}\left[r + \dfrac{4}{\underline{\rho}}\sqrt{\dfrac{\mathbb{E}_\rho \left[\sNorm{\theta}_2^2\right]}{N}}\right].\end{equation}

Step (ii): We then use the McDiarmid’s inequality to show the inequality
\begin{equation}\label{eq:boundongandEg}g(\vartheta) \leq \expt\left[g(\vartheta)\right] + \dfrac{\tallNorm{\coefInfBar/\rho}_{2,\rho}}{\underline{\rho}}\sqrt{\dfrac{2}{N}\ln\left(\dfrac{1}{\delta}\right)},\end{equation} 
holds with a probability of at least $1-\delta$. 

The inequality~\eqref{eq:inftynormbound} then follows from combining \eqref{eq:boundonEg} and \eqref{eq:boundongandEg}, using the definitions of $g(\cdot)$ and $\Omega$, and setting $\bar\beta_0 = \beta_0$ and $\bar\beta_i = \frac{1}{N}\beta_i^r, i$ for $ r \coloneqq \sqrt{\mathbb{E}_\rho \left[\sNorm{\theta}_2^2\right]}  / (\underline{\rho}\sqrt{N}) $.

\underline{Proof of Step (i):} The inequality~\eqref{eq:boundonEg} can be easily derived from the following two inequalities:
\begin{equation}\label{eq:1st}
	\expt\left[\left\|V(\coefInf) -\expt_\rho\left[\bar{V}_\vartheta\right]\right\|_\infty\right] \leq \lipConst r (1+\diamSState) \tallNorm{\coefInfBar/\rho}_{2,\rho}.
\end{equation}

and 
\begin{equation}\label{eq:2nd}
	\expt\left[\left\|\bar{V}_\vartheta -\expt_\rho\left[\bar{V}_\vartheta\right]\right\|_\infty\right] \leq \frac{4L}{\underline{\rho}\sqrt{N}}\tallNorm{\coefInfBar/\rho}_{2,\rho}(1+\diamSState)\sqrt{\mathbb{E}_\rho \left[\sNorm{\theta}_2^2\right]}
\end{equation}
In particular, using these two inequalities we get
\begin{align}
	\expt\left[ g(\vartheta)\right] & =  \expt\left[\left\|V(\coefInf) - \bar{V}_\vartheta \right\|_\infty\right] \nonumber \\
	& =  \expt\left[\left\|V(\coefInf) -\expt_\rho\left[\bar{V}_\vartheta\right] + \expt_\rho\left[\bar{V}_\vartheta\right] - \bar{V}_\vartheta\right\|_\infty\right]\nonumber \\
	& \le \expt\left[\left\|V(\coefInf) -\expt_\rho\left[\bar{V}_\vartheta\right]\right\|_\infty\right] +  \expt\left[\left\|\bar{V}_\vartheta -\expt_\rho\left[\bar{V}_\vartheta\right]\right\|_\infty\right] \nonumber \\
	& \le \lipConst r (1+\diamSState) \tallNorm{\coefInfBar/\rho}_{2,\rho} + \frac{4L}{\underline{\rho}\sqrt{N}}\tallNorm{\coefInfBar/\rho}_{2,\rho}(1+\diamSState)\sqrt{\mathbb{E}_\rho \left[\sNorm{\theta}_2^2\right]} \nonumber \\
	& = L\left(1+\diamSState\right)\tallNorm{\coefInfBar/\rho}_{2,\rho}\left[r + \dfrac{4}{\underline{\rho}}\sqrt{\dfrac{\mathbb{E}_\rho \left[\sNorm{\theta}_2^2\right]}{N}}\right] \label{eq:upp-bound-on-inf-norm}
\end{align}
We next prove \eqref{eq:1st} and \eqref{eq:2nd}. 

First notice that since $\theta^i,i =1,\ldots,N,$ are iid samples, we have $\expt_{\rho}\left[\bar V_\vartheta\right] = \beta_0 +  \expt_{\rho}\left[V_{1,\vartheta}\right]$. In addition, since $\coefInfBar:\Theta\mapsto\R$ is $(2,\rho)$-integrable function and thus measurable, it can be written by its positive and negative parts as follows: $ \coefInfBar =  \coefInfBar_+ -  \coefInfBar_-$ where $\coefInfBar_+ \coloneqq \max(0,\coefInfBar)$ and $ \coefInfBar_- \coloneqq \max(0,-\coefInfBar)$. It is also known that both positive and negative parts of a measurable function are measurable. Hence, for every $s\in\sSpace$ we can write
\begin{align}
	\expt_{\rho}\left[\bar V_\vartheta(s)\right] & = \beta_0 + \expt_{\rho}\left[V_{1,\vartheta}(s)\right] \nonumber\\
	& = \beta_0 +  \int_{\Theta} \rho(\theta^1)\left[ \frac{\varphi(s;\theta^1)}{\rho(\theta^1)} \int_{\Theta} \coefInfBar(\theta)\indicator\{\theta: \lVert \theta - \theta^1 \rVert_2\le r\}\mathrm{d}\theta \right]\mathrm{d}\theta^1 \nonumber\\
	& = \beta_0 +  \int_{\Theta} \left( \coefInfBar_+(\theta) -  \coefInfBar_-(\theta)\right) \left[ \int_{\Theta} \varphi(s;\theta^1) \indicator\{\theta: \left\lVert \theta - \theta^1 \right\rVert_2 \le r\}\mathrm{d}\theta^1 \right]\mathrm{d}\theta \nonumber\\
	& \le \beta_0 +  \int_{\Theta} \coefInfBar_+(\theta) \left[ \int_{\Theta} \left(\varphi(s;\theta) + \lipConst\lVert (1,s)\rVert_{2}\lVert\theta^1 - \theta\rVert_{2}\right) \indicator\{\theta: \lVert \theta - \theta^1 \rVert_2 \le r\}\mathrm{d}\theta^1 \right]\mathrm{d}\theta \nonumber\\
	& \qquad- \int_{\Theta} \coefInfBar_-(\theta) \left[ \int_{\Theta} \left(\varphi(s;\theta) - \lipConst\left\lVert (1,s)\right\rVert_{2}\left\lVert\theta^1 - \theta\right\rVert_{2}\right) \indicator\{\theta: \lVert \theta - \theta^1 \rVert_2 \le r\}\mathrm{d}\theta^1 \right]\mathrm{d}\theta \nonumber\\
	& \leq \beta_0 + \int_{\Theta} (\coefInfBar_+(\theta) - \coefInfBar_-(\theta))\varphi(s;\theta)\mathrm{d}\theta  \ + \ \lipConst\left\lVert (1,s)\right\rVert_{2}r \int_{\Theta} \left[\coefInfBar_+(\theta) + \coefInfBar_-(\theta)\right]\mathrm{d}\theta \nonumber\\
	& \le V(s;\coefInf) +  \lipConst r\left\lVert (1,s)\right\rVert_{2} \int_{\Theta} \sqrt{\left(\frac{\coefInf(\theta)}{\rho(\theta)}\right)^2} \  \rho(\mathrm{d}\theta)\nonumber\\
	& \le V(s;\coefInf) +  \lipConst r (1+\diamSState) \tallNorm{\coefInfBar/\rho}_{2,\rho},\label{eq:boundonexptrho}
\end{align}
where the second equality follows from the definition of $V_{1,\vartheta}(s)$ and $\expt_{\rho}[V_{1,\vartheta}(s)]$;  the third equality from the Fubini's theorem on the exchange of integrals and using $\coefInfBar =  \coefInfBar_+ -  \coefInfBar_-$; the first inequality from the Lipschitz continuity of $\varphi$ (by Assumption \ref{asm:random basis function}), cauchy schwarz inequality, and the fact that both functions $\coefInfBar_+$ and $\coefInfBar_-$ are non-negative; the second inequality from the fact that the indicator function is less than one and $\theta$ is considered in a ball of radius $r$; the third inequality from the definition of $V(\beta)$ and the Jensen's inequality $\expt[\sqrt{\cdot \ }]\le \sqrt{\expt[\cdot]}$; and the last inequality form the definitions of $\diamSState$ and $\tallNorm{\coefInfBar/\rho}_{2,\rho}$. Recalling that \eqref{eq:boundonexptrho} holds for every $s\in\sSpace$, taking expectation from its both sides, and rearranging the terms, we obtain \eqref{eq:1st}.

To prove \eqref{eq:2nd}, we consider a sequence of Rademacher random variables $(\epsilon_1,\dots,\epsilon_N)$, where each $\epsilon_i$ is a uniform sample from $\{-1,1\}$. It is easy to see the function $\beta^r_i\varphi(\cdot)$ is $\frac{L}{\underline{\rho}}\tallNorm{\coefInfBar/\rho}_{2,\rho}$-Lipschitz  and $\beta^r_i\varphi(0) = 0$. This follows from the fact that the function $\varphi$ is $\lipConst$-Lipschitz continuous (by  Assumption \ref{asm:random basis function}) and 
\begin{align}\sup_{{\theta^i}} |\beta^r_i(\theta^i)| &=  \sup_{\theta^i} \left\{ \frac{1}{\rho(\theta^i)}  \int_{\Theta} \left\vert\coefInfBar(\theta) \right\vert \indicator\left\{\theta: \left\lVert \theta - \theta^i\right\rVert_2 \le r\right\}\mathrm{d}\theta \right\}\nonumber\\
	& \le \frac{1}{\underline{\rho}} \int_{\Theta} \sqrt{\left(\frac{\coefInfBar(\theta)}{\rho(\theta)}\right)^2} \rho(\mathrm{d}\theta)\nonumber\\ 
	& = \frac{1}{\underline{\rho}}{\lVert  \coefInfBar/\rho\rVert}_{2,\rho}\label{eq:boundbetasup},
\end{align}
where the first equality holds by the definition of $\beta_i^r$; the first inequality by our assumption that $\rho(\cdot)$ is bounded below by $\underline{\rho}$, and the fact that the indicator function is less than one.

Using Theorem 12(4) of \cite{bartlett2002rademacher}, Cauchy-Schwartz inequality, and Jensen’s inequality, we get
\begin{align*}
	\expt_\rho\left[\left\|\bar{V}_\vartheta -\expt_\rho\left[\bar{V}_\vartheta\right]\right\|_\infty\right] &=
	\expt_\rho\left[ \sup_{s} \left\vert \bar{V}_\vartheta -\expt_\rho\left[\bar{V}_\vartheta\right] \right\vert\right] \\
	&\le \frac{2}{N} \expt_{\rho,\epsilon}\left[  \sup_{s} \left| \sum_{i=1}^{N}  \epsilon_i \beta^r_i \varphi(s;\theta^i)  \right| \right]\\
	&\le \frac{4L}{\underline{\rho}N}\tallNorm{\coefInfBar/\rho}_{2,\rho}\expt_{\rho,\epsilon}\left[  \sup_{s} \left| \sum_{i=1}^{N}  \epsilon_i (1,s)^\top \theta^i  \right| \right]\\
	&\le \frac{4L}{\underline{\rho}N}\tallNorm{\coefInfBar/\rho}_{2,\rho}(1+\diamSState)\expt_{\rho,\epsilon} \left\lVert \sum_{i=1}^{N}  \epsilon_i  \theta^i  \right\rVert_2\\
	&\le \frac{4L}{\underline{\rho}\sqrt{N}}\tallNorm{\coefInfBar/\rho}_{2,\rho}(1+\diamSState)\sqrt{\mathbb{E}_\rho \left[\sNorm{\theta}_2^2\right]}.
\end{align*}
Note that the above inequalities follow similar steps as in inequalities (21) - (24) in \cite{rahimi2008uniform}.

\underline{Proof of Step (ii):} Observe that $g$ is stable under any perturbation of its arguments. In particular, for an arbitrary $\ell\in\{1,2,\dots,N\}$, let $\hat{{\vartheta}}\coloneqq(\theta^1,\theta^2,\dots,\hat\theta^\ell,\dots,\theta^N)$ be the same as $\vartheta$ except its $\ell$-th component i.e. $\hat\theta^i = \theta^i$, for all $i\neq \ell$ and $\hat\theta^\ell \neq \theta^\ell$.  We then have 
\begin{align}
	\left\vert g(\vartheta)  - g(\hat\vartheta)\right\vert
	& = \left\vert \left\lVert V(\coefInf) - \beta_0 - \frac{1}{N}\sum_{i\ne \ell}V_{i,\vartheta}(s) - \frac{1}{N} V_{\ell,\vartheta}(s) \right\rVert_\infty - \left\lVert V(\coefInf) -  \beta_0 - \frac{1}{N}\sum_{i\ne \ell}^{N}V_{i,\hat\vartheta}(s) -  \frac{1}{N} V_{l,{\hat\vartheta}}(s) \right\rVert_\infty\right\vert\nonumber\\
	& \le \frac{1}{N} {\left\lVert V_{\ell,{\vartheta}}(s) \ - \  V_{\ell,{\hat\vartheta}}(s)  \right\rVert}_\infty\nonumber\\
	& = \frac{1}{N}  \left\lVert \beta^r_\ell(\theta^\ell) \varphi(s;\theta^\ell) \ - \  \beta^r_\ell(\hat \theta^\ell) \varphi(s;\hat\theta^\ell)  \right\rVert_\infty\nonumber\\
	& \le \frac{2}{N} \sup_{\theta^\ell}|\beta^r_\ell(\theta^\ell)|\nonumber\\
	& \le \frac{2}{N\underline{\rho}}\tallNorm{\coefInfBar/\rho}_{2,\rho}, \label{eq:boundonperturbation}
\end{align}
where the first equality follows from the definition of $g(\cdot)$; the first inequality from the triangle inequality; the second equality from the definition of $V_{l,\vartheta}(s)$, the second inequality from $\|\varphi\|_\infty \le 1$ (by Assumption \ref{asm:random basis function}), and the last inequality from \eqref{eq:boundbetasup}. 

Given $\varepsilon>0$ and~\eqref{eq:boundonperturbation}, McDiarmid’s concentration inequality guarantees that
\begin{align*}
	&\mathrm{Pr}\left[ g(\vartheta) - \expt\left[g({\vartheta})\right] \ge  \varepsilon \right]   \ \le \ \exp\left(\dfrac{-N\underline{\rho}^2\varepsilon^2}{2\tallNorm{\coefInfBar/\rho}_{2,\rho}^2}\right),
\end{align*}
where $\prob(\cdot)$ denotes the probability over the samples $\vartheta=(\theta^{1},\dots,\theta^N)$.
This inequality indicates that
\[
g\left(\vartheta\right) \le \expt\left[g\left(\vartheta\right)\right] + \dfrac{1}{\underline{\rho}}\tallNorm{\coefInfBar/\rho}_{2,\rho}\sqrt{\dfrac{2}{N}\ln\left(\dfrac{1}{\delta}\right)},
\]
with a probability of at least $1-\delta$.
\hfill \Halmos	
\endproof


\begin{definition}\label{ec:def:high-prob-feas}
	Let $r \coloneqq \sqrt{2\ln\left(\frac{1}{\delta}\right)}  / (L(1+\diamSState)\sqrt{N})$. Given an optimal solution  $\coefFELPVec = (\coefFELP,\coefFELPVecBar)$ to FELP, for $N$ iid samples $\left\{\theta^i, i = 1,2,\ldots,N\right\}$ from $\rho$, we define $\coefFeasVec \in\R^{N+1}$ as follows:
	\begin{equation*}
		\coefFeas{i}  \coloneqq\begin{cases}
			\coefFELP  &\quad \text{for} \quad i=0; \\[6pt]
			\dfrac{1}{N\rho(\theta^i)} \int_{\Theta} \coefFELPVecBar(\theta)\indicator\left\{\theta: \sNorm{\theta - \theta^i}_2 \le r\right\} \diff\theta &\quad \text{for} \quad i=1,2,\dots,N, 
		\end{cases} 
	\end{equation*}	
	and $V(\coefFeasVec) = \coefFeas{0} + \sum_{i = 1}^N \coefFeas{i} \varphi(\cdot; \theta^i)$.
	
\end{definition}

\begin{lemma}\label{ec:lem:high-prob-feas-soln}
	Suppose $\rho(\theta) \ge \underline{\rho}$, for all $\theta\in\Theta$ and Assumption \ref{asm:random basis function} holds. Given  $\varepsilon>0$ and $\delta\in(0,1]$, let $(\coefFELP,\coefFELPVecBar)$ denote an optimal solution to FELP with value function $V^*$ and $\coefFeasVec$ be the corresponding vector defined in Definition~\ref{ec:def:high-prob-feas}. Define \begin{equation}\label{ec:eq:N_epsilon}
		N_\varepsilon \coloneqq\left\lceil \dfrac{\tallNorm{\coefFELPVecBar/\rho}_{2,\rho}^2}{\underline{\rho}^2\varepsilon^2}\left(\contractionFactor+\sqrt{2\ln\left(\dfrac{1}{\delta}\right)}\right)^2 \right \rceil.
	\end{equation}
	\begin{itemize}
		\item[(i)] If $N \ge N_\varepsilon $, with a probability of at least $1-\delta$, it holds that $\tallNorm{V^* - V(\coefFeasVec)}_{\infty} \le\varepsilon$ . 
		\item[(ii)] If $N \ge N_\varepsilon$, with a probability of at least $1-\delta$, the vector $(\coefFeas{0} - \Gamma\varepsilon,\coefFeas{1},\dots,\coefFeas{N})$ is feasible to \FALPprog{N} and 
		\[
		\big\lVert{V^* -\big(V(\coefFeasVec)- \Gamma\varepsilon\big)\big\rVert}_\infty \le \frac{2\varepsilon}{(1-\gamma)}.
		\]
	\end{itemize}
	
\end{lemma}

\proof{Proof.}

\underline{Part (i).}
First notice that the vector $\coefFeasVec$ defined in the Definition \ref{ec:def:high-prob-feas} is the same vector of coefficients $(\bar\beta_0, \bar\beta_1, \ldots, \bar\beta_N)$ defined in Proposition~\ref{prop:rahimi-recht} corresponding to $V(\coefFELPVec) = \coefFELP + \int_\Theta \coefFELPVecBar(\theta)\varphi(s;\theta)\diff \theta$. Following similar steps as in the proof of this proposition, we guarantee that with a probability of at least $1-\delta$
\begin{align*}
	\tallNorm{V^*-V(\coefFeasVec)}_\infty &\le \frac{\tallNorm{\coefFELPVecBar/\rho}_{2,\rho}}{\underline{\rho}\sqrt{N}}\left( \Omega + \sqrt{2\ln\left(\dfrac{1}{\delta}\right)}\right).\label{eqn1:ec:lem:high-prob-feas-soln}
\end{align*}
For $N\ge N_\varepsilon$, this inequality indicates that $\tallNorm{V^*-V(\coefFeasVec)}_\infty \le \varepsilon$ holds
with a probability of at least $1-\delta$.\\[0.5em]
\underline{Part (ii).} If $N\ge N_\varepsilon$, the vector $(\coefFeas{0} - \Gamma\varepsilon,\coefFeas{1},\dots,\coefFeas{N})$ is feasible to \FALPprog{N} with a probability of at least $1-\delta$ since 
\begin{equation}\label{ec:eq:non-positive-F}
	\begin{aligned}
		&(1-\gamma)\big(\coefFeas{0}- \shiftFeas\varepsilon \big)+\sum_{i=1}^{N}\coefFeas{i} \big(\varphi(s;\theta_i) -  \gamma  \expt\big[\varphi({s^\prime};\theta_i) \big\vert s,a\big]\big) \\
		&\hspace{1cm}= V(s;\coefFeasVec) -  { \varepsilon} -  \gamma  \expt\big [V(s^\prime;\coefFeasVec)   + {\varepsilon}  \big \vert s,a\big]\\
		&\hspace{1cm}\le V^*(s) -  \gamma  \expt[V^*(s^\prime) \vert s,a] \\
		&\hspace{1cm}=(1-\gamma)\coefFELP +   \int_\Theta \coefFELPVecBar(\theta)\big(\varphi(s)  \ - \ \gamma  \expt[{\varphi({s^\prime})}  \ |  \ s,a] \big)\diff \theta \\
		&\hspace{1cm}\le c(s,a),
	\end{aligned}
\end{equation}
where the first equality comes from the definitions of $V(\coefFeasVec)$ and $\shiftFeas$; the first inequality holds because  $\lvert{V^*(s)-V(s;\coefFeasVec)}\rvert\le\lVert V^*-V(\coefFeasVec)\rVert_{\infty}\le { \varepsilon}$ for all $s\in\sSpace$ with a probability of at least $1-\delta$ by Part (i) of this lemma; the second equality results from using the definition of $V^*$; and the second inequality holds because $(\coefFELP,\coefFELPVecBar)$ is an optimal (hence feasible) solution of FELP. 

Moreover, if $N\ge N_\varepsilon$, by Part (i) of this lemma and the definition of $\shiftFeas$, we get
\[\tallNorm{V^* -\left(V(\coefFeasVec)-\shiftFeas\varepsilon\right)}_\infty \le \tallNorm{V^* -V(\coefFeasVec)}_\infty + \shiftFeas\varepsilon  \leq \varepsilon + \shiftFeas\varepsilon = \dfrac{2 \varepsilon}{(1-\gamma)}\]
with a probability of at least $1-\delta$.
\hfill \Halmos	
\endproof


\subsubsection*{Proof of Theorem \ref{prop:ALP}.}
\underline{Part (i).} The function  $V(\cdot;\coefALPVec)$ is continuous due to the continuity of the class of basis functions $\varphi$ (by Assumption \ref{asm:random basis function}), and is feasible to constraints \eqref{constr:ELP} due to the feasibility of $\coefALPVec$ to \FALPprog{N}. Hence, Lemma \ref{ec:lem:optV-properties} guarantees $V(s;\coefALPVec) \le V^*(s)$ for all  $s\in \sSpace$.\\[0.1em]

\underline{Part (ii).} Consider $\varepsilon>0$. Given $\coefFeasVec = (\coefFeas{0},\coefFeas{1},\ldots,\coefFeas{N})$ and $N_\varepsilon$ respectively defined in Definition~\ref{ec:def:high-prob-feas} and Lemma \ref{ec:lem:high-prob-feas-soln}, part (ii) of Lemma \ref{ec:lem:high-prob-feas-soln} ensures that when $N\ge N_\varepsilon$, the vector
$ \left(\coefFeas{0}- \shiftFeas\varepsilon,\coefFeas{1},\ldots,\coefFeas{N}\right)$ is  a feasible solution to  \FALPprog{N} with a probability of at least  $1-\delta$ and hence
\begin{equation*}
	\left\lVert V^* - V(\coefALPVecN{N}) \right\rVert_{1,\nu}   
	\le  \left\lVert V^*  -  \left(V(\coefFeasVec)-\shiftFeas\varepsilon\right) \right\rVert_{1,\nu}  \le
	\left\lVert V^*  -  \left(V(\coefFeasVec)-\shiftFeas\varepsilon\right) \right\rVert_\infty   \le \dfrac{2\varepsilon}{1-\gamma},
\end{equation*}
where we used the optimality of $\coefALPVecN{N}$ to obtain the first inequality, the relationship between $(1,\nu)$- and $\infty$-norms to obtain the second inequality, and part (ii) of Lemma \ref{ec:lem:high-prob-feas-soln} for the last one. Since $N\ge N_\varepsilon$, the proof is complete if we choose 
\[
\varepsilon \le \dfrac{\tallNorm{\coefFELPVecBar/\rho}_{2,\rho}}{\underline{\rho}\sqrt{N}}\left(\contractionFactor+\sqrt{2\ln\left(\dfrac{1}{\delta}\right)}\right).
\]

\subsubsection*{Proof of Proposition \ref{EC:prop:FALP-constr-sample}.}
The proof follows from the Corollary 1 and Theorem 1 in \citealt[abbreviated by \citetalias{calafiore2006scenario}]{calafiore2006scenario}, applied to the program \eqref{sampleFALP1}, which is a random relaxation of \FALPprog{N}.  Under Assumptions 1 and 2 in \citetalias{calafiore2006scenario}, Corollary 1 and Theorem 1 guarantee that with probability of at least $1-\delta$, the optimal solution $\hat\beta$ of problem \eqref{sampleFALP1} satisfies:
\[\psi\left(\left\{ (s,a)\in\saSpaceS \ : \  \FAconstr( \hat\beta; s,a) \right\}  \right) \ge 1-\delta,\]
where given $\beta=(\beta_0,\beta_1,\dots,\beta_N)\in\R^{N+1}$, the function $\FAconstr: \R^{N+1}\times \saSpaceS\mapsto \R$ is defined as follows:
\[\FAconstr( \beta; s,a)\coloneqq (1-\gamma)\beta_0 + \sum_{i=1}^{N}\beta_i \left(\varphi(s;\theta^i)  - \gamma \expt \big[\varphi(s';\theta^i) \  | \ s,a\big]\right)  - c(s,a). \]

We only need to show that Assumptions 1 and 2 of \citetalias{calafiore2006scenario} hold in our setting. First notice that we use the notations $\FAconstr$, $\coef$, $\R^{N+1}$, $N+1$, $(s,a)$, and $\saSpaceS$ in this paper instead of $f$, $\theta$, $\Theta$, $n_\theta$, $\delta$, and $\Delta$, respectively, in \citetalias{calafiore2006scenario}. Assumption 1 in \citetalias{calafiore2006scenario} requires the function $\FAconstr(\beta; \cdot,\cdot)$ to be convex in $\beta$ and continuous. This clearly holds in our paper since $\FAconstr(\beta; \cdot,\cdot)$ is linear in $\beta$ and we assume $\varphi(\cdot)$ is a Lipschitz continuous function.
We use a relaxation of Assumption 2 in \citetalias{calafiore2006scenario} as stated in their Appendix A. In particular, we only show that the program \eqref{sampleFALP1} is feasible and forgo the uniqueness assumption of the optimal solution to  \FALPprog{N}.  Define $\underline{c}\coloneqq\min_{s,a}c(s,a)/(1-\gamma)$ which is well-defined since $c(\cdot,\cdot)$ is bounded by Assumption \ref{subsec:SumAssump}. It is straightforward to verify that $(\underline{c},0,\dots,0)\in\R^{N+1}$ is feasible to \FALPprog{N} and hence feasible to  program \eqref{sampleFALP1} for all samples $\{(s^k,a^k)\in\mathcal{S}\times\mathcal{A}: k=1,2,\dots,K \}$.
\hfill \Halmos
\endproof

\subsection{Proofs of Statements in \S \ref{sec:Greedy Policy Guided ALPs}}

\subsubsection*{Proof of Proposition \ref{prop:SG-ALP-basic}.}
Any VFA in the set $\{V(\cdot;\coefSGVecK{\bar{n}B}): \bar{n}=1,2,\dots,n\}$ is a continuous function because of the Lipschitz continuity of $\varphi$ in Assumption \ref{asm:random basis function}. Moreover, each function $V(\cdot;\coefSGVecK{\bar{n}B})$ is feasible to constraints \eqref{constr:ELP} since the vector $\coefSGVecK{\bar{n}B}$ is feasible to the constraints \eqref{FALPConst1} of \FGLPprog{\bar{n}B}. As a result, Lemma \ref{ec:lem:optV-properties} guarantees $V(s;\coefSGVecK{\bar{n}B})\le V^*(s)$ for all  $\bar{n}=1,2,\dots,n$ and $s\in\sSpace$. In addition, self-guiding constraints \eqref{FALPConst2} in \FGLPprog{(\bar{n}+1)B} imply $V(\cdot;\coefSGVecK{\bar{n}B}) \le V(\cdot;\coefSGVecK{(\bar{n}+1)B})$ for $\bar{n}=1,2,\ldots,n-1$.
\hfill \Halmos
\endproof

We require the following definition and Propositions~\ref{EC:prop-Hilbert-space},~\ref{EC:prop-closure}, and~\ref{prop:EC-FGLP-init-rate} to prove Theorem \ref{thm:self-guided-VFA-rate}.\looseness = -1
\begin{definition}\label{def:EC-bspace}
	Given $N$ iid samples $\{\theta^{i}: i=1,2,\dots,N\}$ from $\rho$, we define
	\[\mathcal{B}_N\coloneqq\left\{\coefInfBar:\Theta\mapsto \R \ \  \Big\vert \ \  \exists (\beta_1,\dots,\beta_N)\in \R^{N}, \ \  \sum_{i=1}^N \beta_i^2<\infty,  \ \ \coefInfBar(\theta) = \sum_{i=1}^{N} \beta_i \indicator\{\theta=\theta^i\},  \right\}.\]
	Moreover, let $\bar{\mathcal{B}}_N$ and $\bar{\mathcal{B}}^\bot_N$ denote the closure of $\mathcal{B}_N$ and the perpendicular complement of $\bar{\mathcal{B}}_N$, respectively. In addition, suppose $\mathcal{B} \coloneqq \{\coefInfBar:\Theta\mapsto \R \ : \ \sNorm{\coefInfBar/\rho}_{2,\rho} <\infty \}$ denotes the space of all $(2,\rho)$-integrable functions  equipped with the following inner product
	\[\langle \coefInfBar,\coefInfBar^{\prime}\rangle_{\mathcal{B}}   \coloneqq \int_\Theta \frac{\coefInfBar(\theta) \ \coefInfBar^{\prime}(\theta)}{\rho(\theta)} \diff\theta, \quad \mbox{for}  \ \coefInfBar,\coefInfBar^{\prime}\in\mathcal{B}.\]
	
\end{definition}

\begin{proposition}\label{EC:prop-Hilbert-space} It follows that
	\begin{itemize}
		\item[(i)] The space  $\mathcal{B}$ defined in Definition~\ref{def:EC-bspace} equipped with inner product $\langle \cdotp ,\cdotp\rangle_\mathcal{B}$  is a Hilbert space.
		\item[(ii)] The set $\bar{\mathcal{B}}_N$ is a closed subset of $\mathcal{B}$ under addition and scalar multiplication.
		\item[(iii)] Let $(\coefFELP, \coefFELPVecBar)$ be the optimal solution associated to $V^*$. There exist $\optVprojBar \in \bar{\mathcal{B}}_N$ and $\optVprojPerpBar \in \bar{\mathcal{B}}^\bot_N$ such that  $\coefFELPVecBar = \optVprojBar + \optVprojPerpBar$ and  $\sNorm{\coefFELPVecBar/\rho}_{2,\rho} = \sNorm{\optVprojBar/\rho}_{2,\rho} + \sNorm{\optVprojPerpBar/\rho}_{2,\rho}$.
	\end{itemize}
\end{proposition}

\proof{Proof.}
\underline{Part (i):} The space $\mathcal{B}$ is a Hilbert space by Example 4.5 in \cite{rudin1987real}. 

\underline{Part (ii):} The set $\bar{\mathcal{B}}_N$ is a closed subset of $\mathcal{B}$ since for every $\coefInfBar\in\bar{\mathcal{B}}_N$ with $\coefInfBar(\theta) = \sum_{i=1}^{N} \beta_i \indicator\{\theta=\theta^i\}$, we have $\sNorm{\coefInfBar/\rho}_{2,\rho} \le   \sum_i\beta^2_i/\underline{\rho}<\infty$. In addition, $\bar{\mathcal{B}}_N$ is closed under addition since for every $\coefInfBar,\coefInfBar'\in\bar{\mathcal{B}}_N$, we have $\coefInfBar+\coefInfBar'\in\bar{\mathcal{B}}_N$. It is also closed under scalar multiplication because for every $\coefInfBar\in\bar{\mathcal{B}}_N$ and $\alpha\in\R$, we have $\alpha\coefInfBar\in\bar{\mathcal{B}}_N$. 

\underline{Part (iii):} Since  $\coefFELPVecBar\in\mathcal{B}$, using parts (i) and (ii) and the orthogonal projection theorem of Hilbert spaces (Theorem 5.23 in \citealt{folland1999real}), there exist functions $\optVprojBar \in \bar{\mathcal{B}}_N$ and $\optVprojPerpBar \in \bar{\mathcal{B}}^\bot_N$ such that $\coefFELPVecBar = \optVprojBar + \optVprojPerpBar$  and $\sNorm{\coefFELPVecBar/\rho}_{2,\rho} = \sNorm{\optVprojBar/\rho}_{2,\rho} + \sNorm{\optVprojPerpBar/\rho}_{2,\rho}$. 
\hfill \Halmos	
\endproof

\begin{proposition}\label{EC:prop-closure}
	Consider $\zeta>0$ and $N$ iid samples $\{\theta^{i}: i=1,2,\dots,N\}$ from $\rho$. Let $(\coefFELP, \coefFELPVecBar)$ denote an optimal solution to FELP with $\coefFELPVecBar = \optVprojBar + \optVprojPerpBar$ for some $\optVprojBar \in \bar{\mathcal{B}}_N$ and $\optVprojPerpBar \in \bar{\mathcal{B}}^\bot_N$ (see Proposition~\ref{EC:prop-Hilbert-space}). Define $\optVprojPerp\coloneqq(0,\optVprojPerpBar)$. There exists a coefficient function $\coefInfBar^\zeta_N\in \mathcal{B}_N$ such that for $\coefInf^\zeta_N\coloneqq(\coefFELP,\coefInfBar^\zeta_N)$, we get
	\begin{equation}\label{Ec:eq:zetabound}
		\tallNorm{V^* -  \left(V(\coefInf^\zeta_N)+ V(\optVprojPerp)\right)}_\infty  \le  \zeta.
	\end{equation}
	Moreover, $V(\coefInf^\zeta_N)$ can be represented  as $V(\cdotp;\coefInf^\zeta_N) = \coefFELP   + \sum_{i=1}^{N} \beta^\zeta_i\varphi(\cdotp;\theta^i)$ for some coefficients $\beta^\zeta_i\in\R, i= 1,2, \ldots, N$.
\end{proposition}
\proof{Proof.} 
Given $\zeta >0$, since $\optVprojBar \in\bar{\mathcal{B}}_{N}$ and $\bar{\mathcal{B}}_{N}$ is the closure of $\mathcal{B}_{N}$, there  exists a function $\coefInfBar^\zeta_N \in {\mathcal{B}}_{N}$ such that $\|(\optVprojBar - \coefInfBar^\zeta_N)/\rho\|_{2,\rho} \le \zeta^2$.  Therefore, for all $s\in\sSpace$, we have
\begin{align}
	\left\vert V^*(s) -  \left(V(s;\coefInf^\zeta_N)+ V(s;\optVprojPerp)\right) \right\vert^2 
	&= \left( \int_\Theta \frac{1}{\rho(\theta)}\left[\coefFELPVecBar(\theta) - \big(\coefInfBar^\zeta_N(\theta) + \optVprojPerpBar(\theta)\big)\right]\varphi(s;\theta) \rho(\diff \theta) \right)^2  \nonumber\\
	& \le   \int_\Theta \frac{1}{\rho(\theta)^2}\left[\optVprojBar(\theta) + \optVprojPerpBar(\theta) - \big(\coefInfBar^\zeta_N(\theta) + \optVprojPerpBar(\theta)\big)\right]^2  \rho(\diff \theta)  \nonumber\\
	& =  \tallNorm{(\optVprojBar - \coefInfBar^\zeta_N)/\rho}_{2,\rho}\nonumber\\
	&\le \zeta^2,\label{ec:eq:zeta2}
\end{align}
where the first equality follows from the definitions of $V^*(s)$ and $V(s; \cdot)$ evaluated at $\coefInf^\zeta_N$ and $\optVprojPerp$; the first inequality from the Jensen's inequality $(\expt[\cdot ])^2\le {\expt[\cdotp^2]}$ and $\|\phi\|_\infty\le 1$ from Assumption \ref{asm:random basis function}; and the second equality from the definition of the $(2,\rho)$-norm. Since the expression \eqref{ec:eq:zeta2} holds for all $s\in\sSpace$, we have
\[
\left\|V^* -  \left(V(\coefInf^\zeta_N)+ V(\optVprojPerp)\right)\right\|_\infty = \sup_{s\in\sSpace} \left\vert V^*(s) -  \left(V(s;\coefInf^\zeta_N)+ V(s;\optVprojPerp)\right) \right\vert \leq \zeta.
\]

Finally, since $\coefInf^\zeta_N\coloneqq(\coefFELP,\coefInfBar^\zeta_N)$ with $\coefInfBar^\zeta_N \in {\mathcal{B}}_{N}$, the VFA $V(\coefInf^\zeta_N)$ can be represented  as $V(\cdotp;\coefInf^\zeta_N) = \coefFELP   + \sum_{i=1}^{N} \beta^\zeta_i\varphi(\cdotp;\theta^i)$ for some coefficients $\beta^\zeta_i\in\R, i= 1,2, \ldots, N$.
\hfill \Halmos
\endproof

\begin{proposition}\label{prop:EC-FGLP-init-rate}
	Suppose $\rho(\theta) \ge \underline{\rho}$ for all $\theta\in\Theta$. Consider $\zeta>0$, $\delta\in(0, 1]$, and $N$ iid samples $\{\theta^{i}: i=1,2,\dots,N\}$ from $\rho$. Let $(\coefFELP, \coefFELPVecBar)$ denote an optimal solution to FELP and $(\beta^\zeta_1,\dots,\beta^\zeta_N)$ and $\optVprojPerp\coloneqq(0,\optVprojPerpBar)$ be the coefficients described in Proposition~\ref{EC:prop-closure}. For every $H\ge 1$ iid samples $\{\theta^{i}: i=N+1,N+2,\dots,N+H\}$, there exist $(\beta^\bot_0,\beta^\bot_{N+1},\beta^\bot_{N+2},\dots,\beta^\bot_{N+H})\in\R^{H}$
	such that the vector
	$\tilde\beta \coloneqq (\coefFELP + \beta^\bot_0, \beta^\zeta_1 ,\dots,\beta^\zeta_N,\beta^\bot_{N+1},\beta^\bot_{N+2},\dots,\beta^\bot_{N+H})\in\R^{N+H+1}$
	satisfies\looseness = -1
	\[ \left\lVert V^* \ - \ V(\tilde\beta) \right\rVert_{\infty} \le 
	\zeta + 	    \frac{\tallNorm{\optVprojPerpBar/\rho}_{2,\rho}}{\underline{\rho}\sqrt{H}}\left( \Omega + \sqrt{2\ln\left(\dfrac{1}{\delta}\right)}\right),
	\]
	with a  probability of at least $1-\delta$.
\end{proposition}

\proof{Proof.}
Since $\optVprojPerpBar\in\mathcal{B}$, it is easy to see that $V(\optVprojPerp)\in\randBasisSet$. Then, Proposition \ref{prop:rahimi-recht} applied to the function $V(\optVprojPerp)$ and $H$ samples $\{\theta^{i}: i=N+1,N+2,\dots,N+H\}$ guarantees that there are $H$ coefficients $(\beta^\bot_0,\beta^\bot_{N+1},\beta^\bot_{N+2},\dots,\beta^\bot_{N+H})\in\R^{H+1}$, such that
\begin{equation} \label{Ec:eq:ineq1}
	\left\lVert V(\optVprojPerp) \ - \ \left(\beta^\bot_0 + \sum_{i=N+1}^{N+H} \beta^\bot_i \varphi(\cdot;\theta^i) \right) \right\rVert_{\infty} \le 
	\frac{\tallNorm{\optVprojPerpBar/\rho}_{2,\rho}}{\underline{\rho}\sqrt{H}}\left( \Omega + \sqrt{2\ln\left(\dfrac{1}{\delta}\right)}\right),
\end{equation}
with a probability of at least $1-\delta$. Using Proposition \ref{EC:prop-closure} and the triangle inequality, with the same probability, we obtain 
\begin{align*}
	\left\lVert V^* \ - \ V(\tilde\beta) \right\rVert_{\infty} 
	& \le \left\|V^* -  \left(V(\coefInf^\zeta_N)+ V(\optVprojPerp)\right)\right\|_\infty + 
	\left\|\left(V(\coefInf^\zeta_N)+ V(\optVprojPerp)\right) - V(\tilde\beta)\right\|_\infty \\
	& \le \zeta + 
	\left\|{\left(V(\coefInf^\zeta_N)+ V(\optVprojPerp)\right) - \left(\coefFELP + \sum_{i=1}^{N} \beta^\zeta_i \varphi(\cdotp;\theta^i) + \beta^\bot_0+ \sum_{i=N+1}^{N+H} \beta^\bot_i \varphi(\cdotp;\theta^i) \right)}\right\|_\infty \\
	&\le \zeta +  \left\lVert V(\coefInf^\zeta_N) - \coefFELP - \sum_{i=1}^{N} \beta^\zeta_i \varphi(\cdotp;\theta^i) \right\rVert_{\infty} + 
	\left\lVert  V(\optVprojPerp) - \beta^\bot_0 - \sum_{i=N+1}^{N+H} \beta^\bot_i \varphi(\cdotp;\theta^i) \right\rVert_{\infty}\\
	& \le \zeta + 	    \frac{\tallNorm{\optVprojPerpBar/\rho}_{2,\rho}}{\underline{\rho}\sqrt{H}}\left( \Omega + \sqrt{2\ln\left(\dfrac{1}{\delta}\right)}\right),
\end{align*}
where we used \eqref{Ec:eq:zetabound} and definition of $V(\tilde\beta)$ to obtain the second inequality; the triangle inequality for the third inequality; and  $V(\coefInf^\zeta_N) = \coefFELP + \sum_{i=1}^{N} \beta^\zeta_i \varphi(\cdotp;\theta^i)$ and \eqref{Ec:eq:ineq1} for the last one. 
\hfill \Halmos
\endproof

\proof{\normalfont\textbf{Proof of Theorem \ref{thm:self-guided-VFA-rate}.}}
Consider $\zeta\coloneqq \tallNorm{\optVprojPerpBar/\rho}_{2,\rho}\sqrt{2\ln\left({1}/{\delta}\right)}\big/{\underline{\rho}\sqrt{H}}.$ Let $(\coefFELP, \coefFELPVecBar)$  with $\coefFELPVecBar = \optVprojBar + \optVprojPerpBar$ and $\tilde\beta = (\coefFELP + \beta^\bot_0, \beta^\zeta_1,\dots,\beta^\zeta_N,\beta^\bot_{N+1},\beta^\bot_{N+2},\dots,\beta^\bot_{N+H})$ respectively denote an optimal solution to FELP and the coefficient vector described in Proposition~\ref{prop:EC-FGLP-init-rate}. Define $E'\programIndex{N,H} := \frac{1}{1+\shiftFeas}E\programIndex{N,H} - \zeta$
and 
$\hat \beta \coloneqq (\coefFELP + \beta^\bot_0 -  (E'\programIndex{N,H}+ \zeta)\Gamma, \beta^\zeta_1,\dots,\beta^\zeta_N,\beta^\bot_{N+1},\beta^\bot_{N+2},\dots,\beta^\bot_{N+H})\in\R^{N+H+1}.$
We claim that $\hat \beta$ is the desired vector in Theorem \ref{thm:self-guided-VFA-rate}. Note that all elements of $\hat\beta$ are finite and hence the function $V(\hat \beta) = \hat\beta_0 + \sum_{i=1}^{N+H}\hat{\beta}_{i}$ belongs to the set $\mathcal{W}(\Phi_N\cup\Phi_H)$.

\underline{Part (i).} Proposition \ref{prop:EC-FGLP-init-rate} indicates that $\left\|V^* - V(\tilde\beta)\right\|_{\infty} \le \zeta + E'\programIndex{N,H}$ and hence
\begin{equation}\label{eq:EC-FGLP-error-1}
	V(s;\tilde\beta) - \zeta- E'\programIndex{N,H} \le V^*(s) \quad \text{and} \quad V(s;\tilde\beta) +\zeta+ E'\programIndex{N,H} \ge V^*(s), \qquad \forall s\in\sSpace,
\end{equation}
with a probability of at least $1-\delta$.
Let $\beta^\zeta_N := (\beta^*_0,\beta_1^\zeta, \ldots,\beta_N^\zeta)\in\R^{N+1}$ and $\coef^\bot_{N+H} \coloneqq (\beta^\bot_0, \beta^\bot_{N+1},\dots,\beta^\bot_{N+H})\in\R^{H+1}$. With the same probability it holds that
\begin{align*} 
	&(1-\gamma)\hat{\beta}_0+\sum_{i=1}^{N+H}\hat{\beta}_{i} \left(\varphi(s;\theta_i) -  \gamma  \expt\left[\varphi({s^\prime};\theta_i) \big\vert s,a\right]\right) \\
	&  = (1-\gamma)\left(\coefFELP + \beta^\bot_0 -\left(E'\programIndex{N,H}+ \zeta\right)\shiftFeas\right) + \sum_{i=1}^{N}\beta^\zeta_i \left(\varphi(s;\theta_i) -  \gamma  \expt\left[\varphi({s^\prime};\theta_i) \big\vert s,a\right] \right)\\ 
	& \quad + \sum_{i=N+1}^{N+H}\beta^\bot_i \left(\varphi(s;\theta_i) -  \gamma  \expt\left[\varphi({s^\prime};\theta_i) \big\vert s,a\right]\right)\\
	&  = \left[V(s;\coef^\zeta_N) + V(s;\coef^\bot_{N+H}) - (E'\programIndex{N,H}+ \zeta)\shiftFeas\right]  - \gamma\left[\expt\left[V(s';\coef^\zeta_N) + V(s';\coef^\bot_{N+H}) - (E'\programIndex{N,H}+ \zeta)\shiftFeas \big\vert s,a\right]\right]\\
	&   =  V(s;\tilde\beta) - (E'\programIndex{N,H}+ \zeta)\shiftFeas\ - \  \gamma\left[\expt\left[V(s';\tilde\beta) - (E'\programIndex{N,H}+ \zeta)\shiftFeas \big\vert s,a\right]\right]\\
	&   \le V^*(s) -\gamma \expt\big[V^*(s') \big\vert s,a\big]\\
	&  = c(s,a),
	%
\end{align*}
where the first three equalities follows from the definitions of $\tilde \beta$, $\hat \beta$,
$\coef^\zeta_N$, and $\coef^\bot_{N+H}$ and the inequality from \eqref{eq:EC-FGLP-error-1}. The last equality holds since $V^*$ is an optimal solution to ELP.

The above inequality ensures that $\hat\beta$ is feasible to constraints \eqref{FALPConst1} of \FGLPprog{N+H} with a probability of at least $1-\delta$.

\underline{Part (ii).} For any $s\in\saSpaceS$, with a probability of at least $1-\delta$ it holds that
\begin{align*} 
	V(s;\coefSGVecK{N}) 
	&\le V^*(s) \\
	&\le V(s;\tilde\beta)  +\zeta+ E'\programIndex{N,H}\\
	& = V(s;\tilde\beta) -  \left(E'\programIndex{N,H}+ \zeta\right)\shiftFeas + \left(E'\programIndex{N,H}+ \zeta\right)\left(1+\shiftFeas\right)\\
	& = V(s;\hat\beta) +E\programIndex{N,H},
\end{align*}
where the first inequality follows from Propositin~\ref{prop:SG-ALP-basic}; 
the second from \eqref{eq:EC-FGLP-error-1}; and the last equality from the definition $E\programIndex{N,H}$ and the fact that $V(\hat\beta) = V(\tilde\beta) - (E'\programIndex{N,H}+ \zeta)\shiftFeas$.  This shows that $\hat\beta$ is $E\programIndex{N,H}$-feasible solution to constraints \eqref{FALPConst2} with a probability of at least $1-\delta$.

\underline{Part (iii).} Using Proposition~\ref{prop:EC-FGLP-init-rate} and the triangle inequality, with a probability of at least $1-\delta$, we get
\begin{align*}
	\left\|V^* - V(\hat\beta)\right\|_{\infty} &\le \left\|V^* - V(\tilde\beta)\right\|_{\infty}  + (E'\programIndex{N,H}+ \zeta)\shiftFeas \\
	& \le E'\programIndex{N,H}+ \zeta + \left(E'\programIndex{N,H}+ \zeta\right)\shiftFeas = \left(E'\programIndex{N,H}+ \zeta\right)\left(1+\shiftFeas\right) = E\programIndex{N,H}.
\end{align*}


\hfill \Halmos
\endproof

\subsection{Proofs of Statements in \S\ref{sec:extensions}}

\proof{\normalfont\textbf{Proof of Proposition \ref{prop:disc-MDP-error-rate}.}}
Applying Proposition \ref{prop:rahimi-recht} to $V^\mathrm{C}(\cdot)=\coefVC+\int_\Theta \coefVCVecBar(\theta)\varphi(\cdotp;\theta)\diff \theta$ with $\sNorm{\coefVCVecBar/\rho}_{2,\rho}<\infty$ and replacing $\Omega$ with $\Omega^C$, we get that
for $N$ iid samples $\{\theta^{i}: i=1,2,\dots,N\}$ from $\rho$, there exist coefficients
$\bar\beta\coloneqq(\bar\beta_0,\bar\beta_1,\dots,\bar\beta_N)$ such that
\begin{equation}\label{EC-eq:ENbound}
	\sup_{s\in\sSpace^\mathrm{C}} \left|{V^\mathrm{C}(s) - V(s;\bar\beta) }\right| = \left\|{V^\mathrm{C} - V(\bar\beta) }\right\|_\infty \le E_N\coloneqq\frac{\tallNorm{\coefVCVecBar/\rho}_{2,\rho}}{\underline{\rho}\sqrt{N}}\left( \Omega^C + \sqrt{2\ln\left(\dfrac{1}{\delta}\right)}\right),
\end{equation}
with a probability of at least $1-\delta$. Using the definition of $V^C$ (see \S\ref{sec:extensions}), it is straightforward to see that $V^\mathrm{C}(s^m) = V^*(s^m)$ for all $s^m\in\sSpace$. Hence, the inequality~\eqref{EC-eq:ENbound} indicates that with a probability of at least $1-\delta$, 
\begin{equation}\label{EC-eq:supandEN}
	\sup_{s^m\in\sSpace} \left|V^*(s^m) - V(s^m;\bar\beta)\right| = \sup_{s^m\in\mathcal{\sSpace}} \left|V^\mathrm{C}(s^m) - V(s^m;\bar\beta)\right|  \le \sup_{s\in\sSpace^\mathrm{C}} \left|{V^\mathrm{C}(s) - V(s;\bar\beta) }\right| \le E_N,
\end{equation}
where we used the fact that $\sSpace\subseteq \sSpace^\mathrm{C}$ to obtain the first inequality. 

In addition, since $V^*(s^m)$ satisfies FALP  constraints, i.e.
$ V^*(s^m) - \gamma\expt\left[V^*(s') | s^m, a \right] \le c(s^m,a)$, for all $(s^m,a)\in\saSpaceS$, following similar steps as in \eqref{ec:eq:non-positive-F}, the inequality~\eqref{EC-eq:supandEN} indicates that the solution $\hat \beta\coloneqq(\bar \beta_0 - \Gamma E_N ,\bar\beta_1,\dots,\bar\beta_N)$ is feasible to \FALPprog{N} with a probability of at least $1-\delta$. 	Hence, we have
\begin{align*}
	\left\|V^* - V(\coefALPVecN{N})\right\|_{1,\nu} &\le \left\|V^* - V(\hat\beta)\right\|_{1,\nu}\\	
	&= \sNorm{V^* - V(\bar\beta)}_{1,\nu} +\Gamma E_N \\
	& = \sum_{m\in\mathcal{M}} \nu(s^m)\left\vert V^*(s^m) - V(s^m, \bar \beta)\right\vert +\Gamma E_N \\
	& \le (1+\Gamma)E_N \\
	&= \dfrac{2\tallNorm{\coefVCVecBar/\rho}_{2,\rho}}{(1-\gamma)\underline{\rho}\sqrt{N}}\left( \Omega^C + \sqrt{2\ln\left(\dfrac{1}{\delta}\right)}\right),
\end{align*}
where the first inequality follows from the feasibility of $\hat\beta$ and optimality of $\coefALPVecN{N}$ to \FALPprog{N}; the first equality from the definition of $\hat\beta$; the second equality from the $(1,\nu)$-norm definition; the second inequality from \eqref{EC-eq:supandEN}; and the last equality from the definition of $E_N$.

\hfill \Halmos
\endproof

\section{Relaxing Assumptions } \label{sec:RelaxAssumps}
In \S\ref{EC:opt-V-not-in-R} and \S\ref{EC:FinitenessOfFALP},  we discuss how our theory carries over when assumptions $V^*\in\randBasisSet$ and \ref{finitenessOfFALPOpt} do not hold, respectively.
\subsection{Relaxing Assumption of $V^*\in\randBasisSet$}  \label{EC:opt-V-not-in-R}
In this section, we show there exists a feasible solution to FELP such that its VFA is arbitrarily close to $V^*$ under an infinity norm, which then implies that an optimal solution to FELP is arbitrarily close to $V^*$ with respect to a $(1,\nu)$-norm. 
\begin{proposition}\label{prop:ELP-FELP-gap}
	Assume $V^*\notin \randBasisSet$. Given $\varepsilon>0$, there exists a feasible solution, $\FELPFeasSlnVector{\varepsilon} = (\FELPFeasSlnIntercept{0,\varepsilon},\FELPFeasSlnVectorBar{\varepsilon})$ to FELP such that 
	\[\tallNorm{V^* - V(\FELPFeasSlnVector{\varepsilon})}_{\infty} \le \frac{2\varepsilon}{1-\gamma}.\]
\end{proposition}
\proof{Proof.}
Since the optimal value function $V^*$ is continuous (by Assumption \ref{asm:MDP}) and the class of random basis function $\varphi$ is universal (by Assumption \ref{asm:random basis function}),  there is $\hat V\in\randBasisSet$ such that $\sNorm{V^* - \hat V}_\infty \le \varepsilon$.
Since $\hat V$ belongs to $\randBasisSet$, it can be written as $\hat V(s,\hat\coefInf) = \hat\beta_0 +\int_\Theta \hat\coefInfBar(\theta)\varphi(s;\theta)\diff \theta$ for some $\hat\coefInf = (\hat\beta_{0},\hat\coefInfBar)$ with $\sNorm{\hat\coefInfBar/\rho}_{2,\rho} <  \infty$. Recall that $\shiftFeas = {(1+\gamma)}/{(1-\gamma)}$. We now show that  $\FELPFeasSlnVector{\varepsilon} = (\FELPFeasSlnIntercept{0,\varepsilon},\FELPFeasSlnVectorBar{\varepsilon}) := \big( \hat\beta_{0}-\shiftFeas\varepsilon, \hat \coefInfBar\big)$ is the desired feasible FELP solution. This is because $\sNorm{\FELPFeasSlnVectorBar{\varepsilon}/\rho}_{2,\rho} = \sNorm{\hat\coefInfBar/\rho}_{2,\rho} < \infty$ and for any $(s,a)\in\saSpaceS$, we have
\[\begin{aligned}
	&(1-\gamma)\FELPFeasSlnIntercept{0,\varepsilon} +   \int_\Theta \FELPFeasSlnVectorBar{\varepsilon}(\theta)\big(\varphi(s)  \ - \ \gamma  \expt[{\varphi({s^\prime})}  \ |  \ s,a] \big)\diff \theta\\ 
	&\hspace{3cm} \ = \ (1-\gamma)\big(\hat\beta_{0}-\shiftFeas\varepsilon\big)+\int_\Theta \hat \coefInfBar(\theta)\big(\varphi(s)  \ - \ \gamma  \expt[{\varphi({s^\prime})}  \ |  \ s,a] \big)\diff \theta  \nonumber \\
	&\hspace{3cm} \ = \   -(1+\gamma)\varepsilon + \hat V(s) -\gamma  \expt[ \hat V(s^\prime) \vert s,a]  \nonumber \\
	&\hspace{3cm} \ \le	\	-(1+\gamma)\varepsilon+ V^*(s)+\varepsilon   -  \gamma  \expt[V^*(s^\prime) -\varepsilon \vert s,a] \nonumber \\
	&\hspace{3cm} \  = \ V^*(s) - \gamma  \expt[V^*(s^\prime)  \vert s,a] \nonumber \\
	&\hspace{3cm}  \ \le  \  c(s,a), \nonumber
\end{aligned}\]
where the first inequality is valid since $\sNorm{V^*- \hat V}_\infty\le\varepsilon$, which ensures $\hat V(s) \le V^*(s)+\varepsilon$ and $-\hat V(s)\le -V^*(s)+\varepsilon$  for all $s\in\sSpace$.  Thus, $\FELPFeasSlnVector{\varepsilon}$ is feasible to FELP. In addition, the VFA $V(\FELPFeasSlnVector{\varepsilon}) =  \hat V(\hat\beta)  -\shiftFeas\varepsilon$ belongs to $\randBasisSet$ and $\sNorm{V^*- V\left(\FELPFeasSlnVector{\varepsilon}\right)}_\infty\le\sNorm{V^*- \hat V}_\infty +\shiftFeas\varepsilon\le \varepsilon+ \shiftFeas\varepsilon= {2\varepsilon}/{(1-\gamma)}$, which completes the proof.
\hfill \Halmos	
\endproof

\subsection{Relaxing Assumption \ref{finitenessOfFALPOpt}}\label{EC:FinitenessOfFALP}

For a given $\alpha > 0$, define vector $\hat{\beta}\in\R^{N+1}$  as an optimal solution to the following program:
\begin{align}
	\max_{\coef}    \quad&   \beta_0 + \sum_{i=1}^{N}\beta_i\expt_{\nu} \big[\varphi(s;\theta^i) \big] && &&& \nonumber\\
	\text{s.t.}    \ \quad&   (1-\gamma)\beta_0 + \sum_{i=1}^{N}\beta_i \left(\varphi(s;\theta^i)  - \gamma \expt \big[\varphi(s';\theta^i) \  | \ s,a\big]\right)   &&\le  c(s,a), &&& (s,a) \in \saSpaceS   \label{restricted}\\[4pt]
	&  |\beta_i|\le \alpha,&& &&& \forall i=1,2,\dots,N.\nonumber
\end{align}

Although there are explicit bounds on $\beta_1, \beta_2, \ldots,\beta_N$, the constraints of the problem also imply 
\[\beta_0 \le \max_{\{(\beta_1,\ldots,\beta_N): |\beta_i| \le \alpha\}}\ \max_{(s,a) \in \saSpaceS}\left\{ \frac{1}{1-\gamma}\left[c(s,a) - \sum_{i=1}^{N}\beta_i \left(\varphi(s;\theta^i)  - \gamma \expt \big[\varphi(s';\theta^i) \  | \ s,a\big]\right)\right]\right\},\]
where the right hand side is upper bounded by a constant because the state and action spaces are compact, and the cost function evaluations are finite because $V^*$ is bounded, which follows from it being a continuous function defined over a compact set. If the objective function of \eqref{restricted} were a $\sup$ it is easy to see that it can be replaced by a $\max$.  

Proposition \ref{prop:BoundedFALP} develops an error bound for the VFA associated with \eqref{restricted}. 

{\begin{proposition}\label{prop:BoundedFALP} 
		Suppose $\rho(\theta) \geq \underline{\rho}$ for all $\theta \in \Theta$. Given $\delta\in(0,1]$, we have that any optimal solution $\hat{\beta}\in\R^{N+1}$  to linear program \eqref{restricted} with $\alpha\ge {{\lVert  \coefFELPVecBar/\rho\rVert}_{2,\rho}}/(N\underline{\rho})$ satisfies
		\[{\tallNorm{V^* - V(\hat\beta)}_{1,\nu} \ \leq \dfrac{\tallNorm{\coefFELPVecBar/\rho}_{2,\rho}}{\underline{\rho}\sqrt{N}}\left(\contractionFactor+\sqrt{2\ln\left(\dfrac{1}{\delta}\right)}\right),}\] 
		with a probability of at least $1-\delta$. 
	\end{proposition}
	\proof{proof.}
	(i) Any feasible solution  $\beta$ to \eqref{restricted} satisfies $V(s;\beta) \le V^*(s)$ for all  $s\in \sSpace$ by Lemma \ref{ec:lem:optV-properties} since $V(\cdotp;\beta)$ is continuous by Assumption \ref{asm:random basis function}. From this it follows that $\mathbb{E}_{\nu}[V(\beta)] \le \mathbb{E}_{\nu}[V^*]$. By Assumption \ref{asm:MDP}, $V^*$ is a continuous function over a compact domain and is thus bounded by a finite constant, which implies that the optimal objective function value of \eqref{restricted} is also bounded above by this constant. Therefore, \FALPprog{N} has a finite optimal objective function value. 
	
	Let $\beta^*$ be an optimal solution to \eqref{restricted}. Then $\beta^*_1, \beta^*_2, \ldots, \beta^*_N$ are finite because of the bounding constraints. The next proposition develops a VFA error rate for this program.

	(ii) Consider $\varepsilon>0$. Given $\coefFeasVec = (\coefFeas{0},\coefFeas{1},\ldots,\coefFeas{N})$ and $N_\varepsilon$ respectively defined in Definition~\ref{ec:def:high-prob-feas} and Lemma \ref{ec:lem:high-prob-feas-soln}, part (ii) of Lemma \ref{ec:lem:high-prob-feas-soln} ensures that when $N\ge N_\varepsilon$, the vector
	$ \left(\coefFeas{0}- \shiftFeas\varepsilon,\coefFeas{1},\ldots,\coefFeas{N}\right)$ is  a feasible solution to  \FALPprog{N} with a probability of at least  $1-\delta$. From the definition of each element $\coefFeas{i}$, we have that 
	\[
	|\coefFeas{i}|\ \le \ \frac{{\lVert  \coefFELPVecBar/\rho\rVert}_{2,\rho}}{N\underline{\rho}}.
	\]
	Hence, vector
	$ \left(\coefFeas{0}- \shiftFeas\varepsilon,\coefFeas{1},\ldots,\coefFeas{N}\right)$ is  a feasible solution to \eqref{restricted} with a probability of at least  $1-\delta$ and hence
	\begin{equation*}
		\left\lVert V^* - V(\hat\beta) \right\rVert_{1,\nu}   
		\le  \left\lVert V^*  -  \left(V(\coefFeasVec)-\shiftFeas\varepsilon\right) \right\rVert_{1,\nu}  \le
		\left\lVert V^*  -  \left(V(\coefFeasVec)-\shiftFeas\varepsilon\right) \right\rVert_\infty   \le \dfrac{2\varepsilon}{1-\gamma},
	\end{equation*}
	where we used the optimality of $\hat\beta$ to obtain the first inequality, the relationship between $(1,\nu)$- and $\infty$-norms to obtain the second inequality, and part (ii) of Lemma \ref{ec:lem:high-prob-feas-soln} for the last one. Since $N\ge N_\varepsilon$, the proof is complete if we choose 
	\[
	\varepsilon \le \dfrac{\tallNorm{\coefFELPVecBar/\rho}_{2,\rho}}{\underline{\rho}\sqrt{N}}\left(\contractionFactor+\sqrt{2\ln\left(\dfrac{1}{\delta}\right)}\right).
	\]
	\endproof
}

\section{Applying FALP Error Rate to Self-guided FALPs}\label{ec:sec:Analyzing an FALP-based Sampling Bound for FGLP}
In this section, we show that the direct application of the FALP VFA approximation error bound in Theorem \ref{prop:ALP} to self-guided FALPs
leads to an error bound that is weak and does not account for the quality of the self-guiding constraints in an insightful manner. 

To directly apply the analysis used for FALP to self-guided FALPs, we require that $V(\coefSGVecK{N})$ is $\kappa\vectorIndex{N}$ far from to $V^*$, that is, $\min_{s \in \sSpace}|V^*(s) - V(s;\coefSGVecK{N})| \geq \kappa\vectorIndex{N} > 0$. The positivity of $\kappa\vectorIndex{N}$ may not be true when $V^*( \hat s) = V( \hat s;\coefSGVecK{N})$ for a state $\hat s\in\sSpace$. This is thus a restrictive assumption. Proposition \ref{ec:prop:SG-ALP-coservative-bound} states a bound on the number of samples $M$ that follows directly from Lemma \ref{ec:lem:high-prob-feas-soln} and is analogous to the number of samples $N+H$ in \S\ref{sec:Self-guided Approximate Linear Programs}.
\begin{proposition}\label{ec:prop:SG-ALP-coservative-bound}
	Suppose we have an optimal solution $\coefSGVecK{N}$  to \FGLPprog{N} such that $\kappa\vectorIndex{N}>0$. Given $\varepsilon>0$ and $\delta\in(0,1]$, if 
	\[ M  \ge \left\lceil\dfrac{\min\{\varepsilon,\kappa\vectorIndex{N}\}^{-2}}{(1-\gamma)^2}\cdot \dfrac{4\tallNorm{\coefFELPVecBar/\rho}_{2,\rho}^2}{\underline{\rho}^2}\left(\contractionFactor+\sqrt{2\ln\left(\dfrac{1}{\delta}\right)}\right)^2 \right \rceil, \] 
	then any optimal solution $\coefSGVecK{M}$ to \FGLPprog{M} satisfies \[  \tallNorm{ V^* - V(\coefSGVecK{M})}_{1,\nu} \leq \min\{\varepsilon,\kappa\vectorIndex{N}\},\]
	with a probability of at least $1-\delta$. 
\end{proposition}
\proof{Proof.}
Let $\varepsilon^\prime = {(1-\gamma)\min\{\varepsilon,\kappa\vectorIndex{N}\}}/{2}$ and  $\coefFeasVec$ be the corresponding vector defined in Definition~\ref{ec:def:high-prob-feas} with $N$ replaced by $M$. Using Part (ii) of Lemma \ref{ec:lem:high-prob-feas-soln} with the choice of $\varepsilon$ set to $\varepsilon^\prime$, we have that for any
\[
M \ge  N_{\varepsilon^\prime}:=
\left\lceil \dfrac{\tallNorm{\coefFELPVecBar/\rho}_{2,\rho}^2}{\underline{\rho}^2(\varepsilon')^2}\left(\contractionFactor+\sqrt{2\ln\left(\dfrac{1}{\delta}\right)}\right)^2 \right \rceil
\]
vector $(\coefFeas{0} - \shiftFeas\varepsilon^\prime,\coefFeas{1},\dots,\coefFeas{M})$ is feasible to \FALPprog{N} and satisfies 
\[
\big\lVert{V^* -\big(V(\coefFeasVec)- \Gamma\varepsilon’\big)\big\rVert}_\infty \le \frac{2\varepsilon'}{(1-\gamma)}\leq \min\{\varepsilon,\kappa\vectorIndex{N}\}\leq\kappa\vectorIndex{N}
\] 
with a probability of at least $1-\delta$. Thus, for all $s\in \sSpace$, we obtain $V(s;\coefFeasVec)-\shiftFeas\varepsilon^\prime\ge V^*(s) - \kappa\vectorIndex{N}$, and from the definition of $\kappa\vectorIndex{N}$,  we have $V^*(s) -\kappa\vectorIndex{N} \ge V(s;\coefSGVecK{N})$. Hence, for all $s\in \sSpace$, it holds that
\begin{equation}\label{ec:eq:kappaFEasbile}
	V(s;\coefFeasVec)-\shiftFeas\varepsilon^\prime \ \ge \ V^*(s)  -   \kappa\vectorIndex{N} \ \ge \ V(s;\coefSGVecK{N}),
\end{equation}
with a probability of at least $1-\delta$. This shows that for $M \ge N_{\varepsilon^\prime}$, the vector $(\coefFeas{0} - \shiftFeas\varepsilon^\prime,\coefFeas{1},\dots,\coefFeas{M})$ is feasible to 
constraints \eqref{FALPConst2} and \eqref{FALPConst1} of \FGLPprog{M} and it satisfies $\big\lVert V^* -\big(V(\coefFeasVec)-\shiftFeas\varepsilon^\prime \big) \big\rVert_{\infty} \le \min\{\varepsilon,\kappa\vectorIndex{N}\}$, where these statements hold with probability at least $1-\delta$. Therefore, with the same probability, an optimal \FGLPprog{M} solution $\coefSGVecK{M}$ has a smaller $(1,\nu)$ difference with respect to $V^*$, that is, we have
\[
\tallNorm{ V^* - V(\coefSGVecK{M})}_{1,\nu} \le \tallNorm{ V^* - (V(\coefFeasVec)-\shiftFeas\varepsilon')}_{1,\nu}\le \tallNorm{ V^* - (V(\coefFeasVec)-\shiftFeas\varepsilon')}_{\infty}\le \min\{\kappa\vectorIndex{N},\varepsilon\}. 
\]
\hfill \Halmos
\endproof

The sampling lower bound in Proposition \ref{ec:prop:SG-ALP-coservative-bound}  is similar to the FALP bound but has two key differences: (i) it has an additional constant $1/(1-\gamma)^2$ and (ii) $\varepsilon$ is replaced by $\min\{\kappa\vectorIndex{N},\varepsilon\}$. The additional constant $1/(1-\gamma)^2$ stems from constructing in inequality \eqref{ec:eq:kappaFEasbile} a feasible solution to the self-guiding constraints. The intuition behind replacement of $\varepsilon$ by $\min\{\kappa\vectorIndex{N},\varepsilon\}$ is as follows. We assumed that $\min_{s \in \sSpace}|V^*(s) - V(s;\coefSGVecK{N})| \geq \kappa\vectorIndex{N}$, that is, $V(s;\coefSGVecK{N})$ is below $V^*$ by at least $\kappa_N$ at all states. Therefore, a conservative approach to satisfy the self-guiding constraints is to sample sufficiently many random basis functions such that $V(s;\coefSGVecK{M})$ is within $\min\{\kappa\vectorIndex{N},\varepsilon\}$ of $V^*(s)$ at all states. \looseness=-1


\section{A Heuristic Based on Constraint Violation Learning to Obtain a Lower Bound} \label{ec:sec:A Lower Bound Estimator for Constraint-sampled ALPs}
In this section, we elaborate on  how we estimate valid lower bounds given an arbitrary VFA. This material relies on a heuristic use of the exact constraint violation learning approach in \cite{lin2017ContViolLearning}, which is based on a primal-dual approach. The primal updates modify the VFA weights while the dual updates involve distributions that capture regions of high constraint violation. Our heuristic, fixes the VFA weights (i.e., no primal updates) and only employs the dual update and shows that a valid lower bound  on the optimal cost can be estimated. For any VFA $V(\coef)$ with $\coef\in\R^{N+1}$, we define the function
\[
y(s,a;\coef)\coloneqq \ \expt_{\chi}[V(\coef) ]  + \frac{1}{1-\gamma} \Big( c(s,a)  + \gamma \expt\big[V(s^\prime;\coef) \ | \ s,a\big] - V(s;\coef) \Big),
\]
that encodes the  violation of FALP constraints for a given $\coef$ at a state-action pair $(s,a)$.  Note that coefficient $\coef$ is not necessarily feasible to all FALP constraints, which is, $\expt_\nu[V(\beta)]\leq \expt_\nu[V^*]$ may not  hold. We observe that minimizing the function $y(s,a;\beta)$ over state-action pairs corresponds to finding the most violating constraint in the  \FALPprog{N} since term $\expt_{\chi}[V(\coef) ]$ is independent of the state and action and the term ${( c(s,a)  + \gamma \expt[V(s^\prime;\coef)  |  s,a] - V(s;\coef) )}/{(1-\gamma)}$ is the constraint slack. 
Thus, if the minimum value of function $y(s,a;\coef)$ over state-action pairs is strictly less than   $\expt_{\chi}[V(\coef)]$, then  $\coef$ violates a constraint of
\FALPprog{N}. Otherwise, $\coef$ is feasible to \FALPprog{N}. Under mild conditions, function $y$ is Lipschitz with constant $\mathrm{L}_y>0$. 

Lemma \ref{lem:Lin_et_al_lower_bound} is directly based on Lemma EC.3 in \cite{lin2017ContViolLearning} and provides a lower bound on the optimal cost. For a given VFA $V(\coef)$ and $\lambda\in(0,1]$, we define a density $Y$ on $\saSpaceS$ as $Y(s,a;\coef,\lambda) \coloneqq \exp\big({-y(s,a;\coef)}\big/{\lambda}\big)/\int_{\aSpaceS} \exp\big({-y(s,a;\coef)}\big/{\lambda}\big)\diff (s,a)$. Notice the theoretical analyses in \cite{lin2017ContViolLearning} are provided under an assumption that the action space $\aSpaceS$ is compact. However, Lemma EC.3 can still hold if  this assumption does not hold. 

\begin{lemma}[Lemma EC.3 in \citealt{lin2017ContViolLearning}]\label{lem:Lin_et_al_lower_bound}
	For all $\lambda \in (0,1]$ and $\coef$, we have
	$
	\mathrm{PC}(\pi^*)   \ge   	\expt_{Y}\left[y(s,a;\coef)\right]  + \lambda( \Lambda + \saDim \ln(\lambda))
	$
	where 
	\[
	\Lambda \coloneqq -\ln\left[ \bar\Gamma\bigg(1+\frac{\saDim}{2}\bigg) \ 
	\Big(R_{\saSpaceS} \sqrt{\uppi}\Big)^{-\saDim} \  \int_{\aSpaceS} \diff (s,a)\right]  - \mathrm{L}_y(R_{\saSpaceS}+\diamSSaspace),
	\]
	and $\saDim$ is the dimension of the space $\saSpaceS$. Function $\bar\Gamma$ is the standard gamma function, $\uppi$ is the Archimedes constant, $R_{\saSpaceS}>0$  is the radius of the largest ball contained in $\saSpace$,  and $\diamSSaspace$ is the diameter of $\saSpace$.
\end{lemma}

Given a solution $\coef$ and its VFA $V(\coef)$,  Lemma \ref{lem:Lin_et_al_lower_bound} suggests that  a valid lower bound on optimal cost $\mathrm{PC}(\pi^*)$ can be computed by estimating the expected value $\expt_{Y}\big[y(s,a;\coef)\big]$ and a constant term. For our numerical experiments in \S\ref{sec:Perishable Inventory Control}, we estimate $\expt_{Y}\big[f(\coef,s,a)\big]$ using the Metropolis-Hastings method with $4000$ samples by generating $8$ Markov Chains, each with length of $1500$, where we burn the first $1000$ samples and use the last $500$. Parameter $\Lambda$ can be easily evaluated for the instances studied in \S\ref{sec:Perishable Inventory Control}. For perishable inventory control application cost function is Lipschitz with constant $\mathrm{L}_c >0$, where $\mathrm{L}_c = 2(\gamma^Lc_o\bar{a} +c_h\bar{a} + c_b\underline{s} +c_d\bar{a}+c_l\bar{a})$. Hence, we have $\mathrm{L}_y = {(4\sNorm{\coef}_1+\mathrm{L}_c)}/({1-\gamma})$. We choose the other parameters defining $\Lambda$ as follows: $\saDim$ is given by the summation of the dimensions of MDP state and action spaces that  depends on each instance; $R_{\saSpaceS}$ is $\frac{\bar{a}}{2}$ and $\diamSSaspace=3\bar{a}^2 + (\underline{s}-\bar{a})^2$; and $\lambda$ is set to ${1}/{(\Lambda + \saDim) }$ but one can cross-validate this parameter to possibly obtain tighter bounds.

\bibliographystyle{informs2014}

\end{document}